\documentclass[english]{IEEEtran}
\usepackage[T1]{fontenc}
\usepackage[latin9]{inputenc}
\usepackage{array}
\usepackage{amsmath}
\usepackage{amsthm}
\usepackage{amssymb}
\usepackage{graphicx}
\usepackage{xargs}[2008/03/08]

\makeatletter

\providecommand{\tabularnewline}{\\}

\theoremstyle{plain}
\newtheorem{thm}{\protect\theoremname}
\theoremstyle{plain}
\newtheorem{cor}[thm]{\protect\corollaryname}

\usepackage{amsxtra,amssymb,amsmath}
\usepackage[english]{babel}
\usepackage{amsmath}
\usepackage{graphicx}
\usepackage{hyperref}
\usepackage{epstopdf}
\usepackage{framed}
\usepackage{amsthm}
\usepackage{url}
\usepackage{amssymb}
 \usepackage{color}
\usepackage{stmaryrd}
\usepackage{algorithm}
\usepackage{multirow}

\newcommand{\argmax}{\operatornamewithlimits{argmax\,}}
\newcommand{\argmin}{\operatornamewithlimits{argmin\,}}

\author{Viet~Hung~Tran~and~Wenwu~Wang
\IEEEcompsocitemizethanks{\IEEEcompsocthanksitem V.~H.~Tran and W.~Wang are with CVSSP, Univeristy of Surrey, GU27XH Surrey, U.K. Emails: \{v.tran, w.wang\}@surrey.ac.uk.}
\thanks{This work was supported by the Engineering and Physical Sciences Research Council (EPSRC) Grant number EP/K014307/2 and the MOD University Defence Research Collaboration in Signal Processing.}
}

\makeatother

\usepackage{babel}
\providecommand{\corollaryname}{Corollary}
\providecommand{\theoremname}{Theorem}

\begin{document}
\title{Bayesian~Inference~for~PCA~and~MUSIC~Algorithms\\
with Unknown Number of Sources}
\maketitle
\begin{abstract}
Principal component analysis (PCA) is a popular method for projecting
data onto uncorrelated components in lower dimension, although the
optimal number of components is not specified. Likewise, multiple
signal classification (MUSIC) algorithm is a popular PCA-based method
for estimating directions of arrival (DOAs) of sinusoidal sources,
yet it requires the number of sources to be known a priori. The accurate
estimation of the number of sources is hence a crucial issue for performance
of these algorithms. In this paper, we will show that both PCA and
MUSIC actually return the exact joint maximum-a-posteriori (MAP) estimate
for uncorrelated steering vectors, although they can only compute
this MAP estimate approximately in correlated case. We then use Bayesian
method to, for the first time, compute the MAP estimate for the number
of sources in PCA and MUSIC algorithms. Intuitively, this MAP estimate
corresponds to the highest probability that signal-plus-noise's variance
still dominates projected noise's variance on signal subspace. In
simulations of overlapping multi-tone sources for linear sensor array,
our exact MAP estimate is far superior to the asymptotic Akaike information
criterion (AIC), which is a popular method for estimating the number
of components in PCA and MUSIC algorithms. 
\end{abstract}



\global\long\def\argmin{\operatornamewithlimits{arg\,min}}%

\global\long\def\argmax{\operatornamewithlimits{arg\,max}}%


\global\long\def\TRIANGLEQ{\triangleq}%

\global\long\def\bigO{\mathcal{O}}%

\global\long\def\transpose{T}%

\global\long\def\Hermitian{*}%

\global\long\def\REAL{\mathbb{R}}%

\global\long\def\COMPLEX{\mathbb{C}}%

\global\long\def\EXPECTATION{\mathbb{E}}%

\global\long\def\Re{\text{Re}}%

\global\long\def\IDEN{\mathbf{I}}%

\global\long\def\UNIT{\mathbb{I}}%

\global\long\def\ONE{\mathbf{1}}%

\global\long\def\ZERO{\mathbf{0}}%

\global\long\def\Diag{\text{Diag}}%

\global\long\def\diag{\text{diag}}%

\global\long\def\Trace{\text{Tr}}%

\global\long\def\etr{\text{etr}}%

\global\long\def\Lp{\mathcal{L}}%

\global\long\def\Lzero{\Lp_{0}}%

\global\long\def\Lone{\Lp_{1}}%

\global\long\def\Ltwo{\Lp_{2}}%

\global\long\def\Linfty{\Lp_{\infty}}%

\global\long\def\Prob{\text{Pr}}%

\global\long\def\var{\text{var}}%

\global\long\def\calV{\mathcal{V}}%

\global\long\def\calY{\mathcal{Y}}%

\global\long\def\calS{\mathcal{S}}%

\global\long\def\calP{\mathcal{P}}%

\global\long\def\calQ{\mathcal{Q}}%

\global\long\def\calI{\mathcal{I}}%

\global\long\def\calC{\mathcal{C}}%

\global\long\def\calN{\mathcal{N}}%

\global\long\def\calCN{\mathcal{CN}}%

\global\long\def\calCMN{\mathcal{CMN}}%

\global\long\def\calVM{\mathcal{VM}}%

\global\long\def\calMu{Mu}%

\global\long\def\calG{\mathcal{G}}%

\global\long\def\caliG{i\mathcal{G}}%

\global\long\def\calrG{\mathcal{G}}%

\global\long\def\calGGL{\calG\calG}%

\global\long\def\calGGU{\calG\calG\mathcal{U}}%

\global\long\def\caliGGL{i\calG\calG}%

\global\long\def\caliGGU{i\calG\calG\mathcal{U}}%

\global\long\def\calEXP{\text{EXP}}%

\global\long\def\calNB{\mathcal{NB}}%

\global\long\def\calB{\mathcal{B}}%

\global\long\def\calNiW{\mathcal{N}i\mathcal{W}}%

\global\long\def\caliW{i\mathcal{W}}%

\global\long\def\SNR{\text{SNR}}%

\global\long\def\hSNR{\overline{\text{SNR}}}%


\global\long\def\ntime{N}%

\global\long\def\nstate{K}%

\global\long\def\Nstate{K_{max}}%

\global\long\def\nsource{M}%

\global\long\def\ndim{D}%

\global\long\def\ngrid{L}%

\global\long\def\npara{Q}%

\global\long\def\norder{R}%

\global\long\def\itime{t}%

\global\long\def\istate{k}%

\global\long\def\isource{m}%

\global\long\def\idim{d}%

\global\long\def\igrid{r}%

\global\long\def\iorder{r}%

\global\long\def\iton{\itime=1,\ldots,\ntime}%

\global\long\def\iinn{\itime\in\{1,\ldots,\ntime\}}%

\global\long\def\jtom{\istate=1,\ldots,\nstate}%

\global\long\def\kinK{\idim\in\{1,\ldots,\ndim\}}%

\global\long\def\ktom{\istate=1,\ldots,\nstate}%

\global\long\def\seti#1#2{#1\in\{1,2,\ldots,#2\}}%

\global\long\def\setk#1#2{#1{}_{1},#1{}_{2},\ldots,#1_{#2}}%

\global\long\def\setp#1#2{(#1{}_{1},#1{}_{2},\ldots,#1_{#2})}%

\global\long\def\setv#1#2{[#1{}_{1},#1{}_{2},\ldots,#1_{#2}]}%

\global\long\def\setd#1#2{\{#1{}_{1},#1{}_{2},\ldots,#1_{#2}\}}%

\global\long\def\sete#1#2#3{[#1{}_{1,#3},#1{}_{2,#3},\ldots,#1_{#2,#3}]}%

\global\long\def\NORMTWO#1{||#1||^{2}}%


\global\long\def\gULA{\boldsymbol{g}}%

\global\long\def\sFreq{\gamma}%

\global\long\def\sbFreq{\boldsymbol{\sFreq}}%

\global\long\def\sPhase{\phi}%

\global\long\def\sbPhase{\boldsymbol{\sPhase}}%

\global\long\def\data{x}%

\global\long\def\xbold{\boldsymbol{x}}%

\global\long\def\Data{\boldsymbol{X}}%

\global\long\def\ydata{y}%

\global\long\def\ysigma{\upsilon}%

\global\long\def\ybold{\boldsymbol{y}}%

\global\long\def\yData{\boldsymbol{Y}}%

\global\long\def\yVECeigen{\boldsymbol{Q}}%

\global\long\def\yVecEigen{\boldsymbol{q}}%

\global\long\def\yEIGEN{\boldsymbol{\Lambda}}%

\global\long\def\yEigen{\boldsymbol{\lambda}}%

\global\long\def\yeigen{\lambda}%

\global\long\def\radius{\eta}%

\global\long\def\Angle{\phi}%

\global\long\def\bAngle{\boldsymbol{\phi}}%

\global\long\def\angleDOA{\omega}%

\global\long\def\hangleDOA{\widehat{\angleDOA}}%

\global\long\def\oangleDOA{\overline{\angleDOA}}%

\global\long\def\bangleDOA{\boldsymbol{\angleDOA}}%

\renewcommandx\hangleDOA[1][usedefault, addprefix=\global, 1=]{\widehat{\boldsymbol{\angleDOA}}_{#1}}%

\global\long\def\initDOA{\omega_{0}}%

\global\long\def\gridDOA{\Delta\omega}%

\global\long\def\GridDOA{\Delta\omega_{0}}%

\global\long\def\stdDOA{\sigma_{\angleDOA}}%

\global\long\def\ULa{v}%

\global\long\def\bULa{\boldsymbol{v}}%

\global\long\def\bULA{\boldsymbol{V}}%

\global\long\def\ULA{\boldsymbol{V}_{\bangleDOA}}%

\global\long\def\UULA{\boldsymbol{\Phi}}%

\global\long\def\SULA{\boldsymbol{\Sigma}}%

\global\long\def\MeanULA{\overline{\boldsymbol{V}}}%

\global\long\def\VecEigenULA{\boldsymbol{U}}%

\global\long\def\EigenULA{\boldsymbol{\Lambda}}%

\global\long\def\signal{s}%

\global\long\def\bsignal{\boldsymbol{s}}%

\global\long\def\Signal{\boldsymbol{S}}%

\global\long\def\source{w}%

\global\long\def\sbold{\boldsymbol{w}}%

\global\long\def\Source{\boldsymbol{W}}%

\global\long\def\amplitude{a}%

\global\long\def\mamplitude{\overline{a}}%

\global\long\def\bamplitude{\boldsymbol{\amplitude}}%

\global\long\def\meanAmp{\overline{\bamplitude}}%

\global\long\def\Amplitude{\boldsymbol{A}}%

\global\long\def\AmpZero{\boldsymbol{A}_{0}}%

\global\long\def\MeanAmp{\overline{\boldsymbol{A}}}%

\global\long\def\stdZero{\sigma_{0}}%

\global\long\def\pseudoAmp{\boldsymbol{A}^{+}}%

\global\long\def\pseudoDOA{\bULA^{+}}%

\global\long\def\Height{\boldsymbol{H}}%

\global\long\def\HZero{\Height_{0}}%

\global\long\def\covULA{u}%

\global\long\def\bcovULA{\boldsymbol{\covULA}}%

\global\long\def\wavelength{\lambda}%


\global\long\def\NormHS#1{||#1||}%

\global\long\def\NormSquared#1{\NormHS{#1}^{2}}%


\global\long\def\para{\theta}%

\global\long\def\bpara{\boldsymbol{\theta}}%

\global\long\def\Para{\Theta}%

\global\long\def\HyperPara{\boldsymbol{\Upsilon}}%

\global\long\def\orderCond{\calI_{0}}%

\global\long\def\differCond{\calI_{1}}%


\global\long\def\noise{z}%

\global\long\def\Noise{\boldsymbol{E}}%

\global\long\def\bnoise{\boldsymbol{z}}%

\global\long\def\eNoise{\boldsymbol{Z}}%

\global\long\def\nsigma{\sigma}%

\global\long\def\vkappa{\kappa}%

\global\long\def\nalpha{\alpha}%

\global\long\def\nbeta{\beta}%

\global\long\def\mean{\mu}%

\global\long\def\std{\sigma}%

\global\long\def\stderror{\sigma}%

\global\long\def\Std{\Sigma}%

\global\long\def\stdamplitude{\std_{a}}%

\global\long\def\VarAmplitude{\boldsymbol{\Psi}}%

\global\long\def\varamp{\varrho}%

\global\long\def\varsignal{r}%

\global\long\def\varSignal{\varsignal_{\amplitude}}%

\global\long\def\varR{\boldsymbol{R}}%

\global\long\def\rationoise{\tau}%

\global\long\def\ratioNoise{\rationoise}%

\global\long\def\ratiosignal{\gamma}%

\global\long\def\ratioSignal{\ratiosignal_{a}}%

\global\long\def\hratioSignal{\widehat{\ratiosignal_{a}}}%

\global\long\def\hratioNoise{\widehat{\ratioNoise}}%

\global\long\def\zsigma{\zeta}%


\global\long\def\threshSNR{\tau_{0}}%

\global\long\def\threshMUSIC{\threshSNR}%

\global\long\def\threshV{\tau}%

\global\long\def\threshEnd{\tau_{\ndim}}%

\global\long\def\threshK{\nstate_{\max}}%


\global\long\def\StiefelRadius{R}%


\global\long\def\iGnx{\alpha}%

\global\long\def\iGsx{s}%

\global\long\def\iGny{\beta}%

\global\long\def\iGsy{t}%

\global\long\def\iGp{p}%

\global\long\def\iGq{q}%

\global\long\def\iGi{n}%


\global\long\def\GGx{X}%

\global\long\def\GGy{Y}%

\global\long\def\GGnx{n}%

\global\long\def\GGsx{s}%

\global\long\def\GGny{m}%

\global\long\def\GGsy{t}%

\global\long\def\GGu{\alpha}%

\global\long\def\GGv{\beta}%

\global\long\def\GGp{p}%

\global\long\def\GGq{q}%

\global\long\def\GGn{n}%

\global\long\def\GGi{k}%

\global\long\def\GGs{x}%

\global\long\def\GGt{t}%

\global\long\def\Regular{I}%

\section{Introduction}

In many systems of array signal processing, e.g. in radar, sonar and
antenna systems, linear sensor array is the most basic and universal
mathematical model. Because far distant sources with different directions
of arrival (DOAs) will oscillate the steering sensor array with different
angular frequencies, the array's output data is then a superposition
of sinusoidal signals \cite{DoA:AtomicNorm:16}. Hence, a common problem
of array systems is to detect the number of sources, as well as their
tone frequencies and DOAs, from noisy sinusoidal signals. 

In literature, most papers only consider the case of single-tone or
narrowband sources (i.e. sources with non-overlapping tones), for
which the line spectrum is a popular estimation method (e.g. in \cite{Atomic:Gongguo:13,Fleury:VBLineSpectra:17}).
When the number of sources is small, the DOA's line spectra are sparse
and can be estimated effectively via sparse techniques like atomic
norm (also known as total variation norm) \cite{DoA:AtomicNorm:16,Atomic:Gongguo:13},
LASSO \cite{LASSO:MichealJordan:12,DoA:SequenBayes:13} and Bayesian
compressed sensing \cite{Bayes:compress:17,Bayes:compress:EM:17}.
The near optimal bound for the atomic norm approach was also given
in \cite{Atomic:Gongguo:15}. 

In this paper, however, we are interested in a more general case with
arbitrary number of overlapping multi-tone~sources. In this case,
the most popular DOA's and frequency's estimation techniques are perhaps
discrete time Fourier transform (DTFT), MUSIC and ESPRIT algorithms,
originated from signal processing techniques \cite{Proakis:BOOK:DSP:06}.
Both MUSIC and ESPRIT belong to subspace methods, whose purpose is
to extract signal subspace and noise subspace from noisy data space
via eigen-decomposition. Although the computational complexity of
ESPRIT is lower, the MUSIC is, however, more popular in practice \cite{MUSIC:cite:12,MUSIC:cite:16}.
For example, in smart antenna models, the MUSIC algorithm for DOA
estimation was shown to be more stable and accurate than ESPRIT \cite{MUSIC:vs:ESPRIT:10,MUSIC:vs:ESPRIT:12}.
Also, the key disadvantage of ESPRIT is that it would require two
translational invariant sub-arrays in order to exploit the invariant
rotational subspace of angular frequencies \cite{Proakis:BOOK:DSP:06}. 

Hence, for this general case, we focus on DTFT and MUSIC algorithms
in this paper. Although these two methods can achieve high resolution
DOA's estimation for uncorrelated and weakly correlated sources, the
number $\nstate$ of sources must be known beforehand \cite{Proakis:BOOK:DSP:06,DoA:SORTE:13}.
In practice, an accurate estimation of unknown $\nstate$ is then
a critical issue for DOA's estimation in these methods. Our objective
is to provide the optimal maximum-a-posteriori (MAP) estimate for
this unknown $\nstate$ in this paper.

\subsection{Related works}

Since MUSIC algorithm is a variant of principal component analysis
(PCA), the most common approach for estimating $\nstate$ is to apply
the eigen-based methods in clustering literature (c.f. \cite{DoA:SORTE:13}),
of which the most widely-used methods are information criterions like
minimum description length (MDL) \cite{DoA:AIC_MDL:02} and Akaike
information criterion (AIC) for PCA \cite{DoA:AIC:VanTrees:02}. Nonetheless,
the information criterions like MDL and AIC are merely asymptotic
approximations of maximum likelihood (ML) estimate in the case of
infinite amount of data \cite{DoA:AIC_MDL:02,AIC:Akaike:74}. Likewise,
hard-threshold schemes for eigen-based methods are mostly heuristic
\cite{DoA:SORTE:13} and only optimal in asymptotic scenarios~\cite{PCA:AsymptoticMMSE:14}. 

Despite being invented in early years of twentieth century \cite{PCA:Pearson:01,PCA:Hotelling:33},
PCA is still one of the most popular methods for reducing data's dimension
\cite{PCA:review:16}. In \cite{PCA:Bishop:99}, PCA was shown to
be equivalent to maximum likelihood (ML) estimate of factor analysis
model with additive white Gaussian noise (AWGN). Then, apart from
heuristic eigen-based methods, the non-asymptotic probabilistic methods
for estimating $\nstate$ are mostly based on Bayesian theory. Nonetheless,
the posterior probability distributions of principal vectors and the
number $\nstate$ of components in PCA are very complicated in general
and do not belong to any known distribution family \cite{PCA:VB:AQuinn:07,PCA:tau:11}.
Hence, their closed-form is still an open problem in literature \cite{PCA:ExactDimension:17,PCA:Bayes:ICASSP17,PCA:nonparametric:17}.
For this reason, the MAP estimate of $\nstate$ could only be computed
via approximations like Laplace \cite{PCA:Minka:Laplace:00}, Variational
Bayes \cite{PCA:VB:AQuinn:07} and sampling-based Markov chain Monte
Carlo \cite{PCA:tau:11,PCA:Bayes:ICASSP17,PCA:nonparametric:17} in
literature.

To our knowledge, the most recent attempt to derive the exact MAP
estimate of $\nstate$ in PCA is perhaps the Theorem 5.1 in \cite{PCA:ExactDimension:17}.
Unfortunately, this theorem imposed a restricted form of standard
normal-gamma prior on signal's amplitudes and noise's variance and,
hence, still involved two unknown hyper-prior parameters for this
prior. These two unknown parameters were then estimated via heuristic
plug-in method in \cite{PCA:ExactDimension:17}, before estimating
$\nstate$.

\subsection{Paper's contributions}

In contrast to \cite{PCA:ExactDimension:17}, we will use non-informative
priors, without imposing any hyper-prior parameter in this paper.
For this purpose, we have derived two novel probability distributions,
namely double gamma and double inverse-gamma, in Appendix \ref{sec:Double-Gamma}.
These novel distributions will help us compute, for the first time,
the exact MAP estimate of $\nstate$ and posterior mean estimate of
signal's and noise's variance, without any prior knowledge of sources
in PCA and MUSIC models. Intuitively, our MAP estimate $\widehat{\nstate}$
is equivalent to picking the dimension $\nstate$ of signal subspace
such that the signal-plus-noise's variance is higher than the projected
noise's variance on that signal subspace with highest probability. 

Our novel distributions actually arise as a natural form for the joint
posterior distribution of signal's and noise's variance in the PCA
model. The motivation of these novel distributions is that, under
assumption of Gaussian noise, both variances of noise and signal-plus-noise
would follow inverse-gamma distributions a-posteriori, since inverse-gamma
distribution is conjugate to Gaussian distribution. Furthermore, since
signal-plus-noise's variance must dominate noise's variance a-posteriori,
the double inverse-gamma distribution arises naturally as the joint
distribution of these two inverse-gamma distributions under this domination
constraint. For this reason, these novel distributions are also useful
for joint estimation of unknown source's and noise's variance in generic
linear AWGN models. 

Owing to these novel distributions, we will show that PCA and MUSIC
actually return the joint MAP estimate of uncorrelated principal/steering
vectors for both cases of known and unknown noise's variance, although
these methods can only approximate this joint MAP estimate in the
case of correlated principal/steering vectors.

Since MAP estimate is the optimal estimate for averaged zero-one loss
(also known as averaged $\Linfty$-norm), as shown in Appendix \ref{sec:Bayesian-minimum-risk},
our MAP estimate $\widehat{\nstate}$ is superior to asymptotic AIC
method in our simulations. Also, since accurate estimation of $\nstate$
increases the performance of DTFT and MUSIC algorithms significantly,
our MAP estimate $\widehat{\nstate}$ subsequently leads to higher
accuracy for DOA's and amplitude's estimation in these two algorithms,
particularly for overlapping multi-tone sources. 

In literature of DOA, we recognize that very few papers consider the
case of multi-tone sources, even though such sources appear frequently
in practice. Indeed, both narrow band and overlapping band sources
are examples of this case. To our knowledge, this is the first paper
studying DOA's estimation for generic multi-tone sources in Bayesian
context. A much simpler version of this paper was recently published
in \cite{VH:ICASSP:18}, but merely for estimating binary on-off states
of multi-tone sources, with known noise's variance. 

\subsection{Paper's organization}

Firstly, in section \ref{sec:Sensor-array}, the linear sensor array
will be reinterpreted as a linear model in frequency domain, given
the AWGN assumption. The PCA, MUSIC and spectrum DTFT algorithms are
then reviewed in section \ref{sec:PCA}, under new perspective of
the Pythagorean theorem for Hilbert--Schmidt norm \cite{Hilbert_Schmidt:inequality:81}.
Full Bayesian analysis and MAP estimates for these three algorithms
are presented next in section \ref{sec:Bayesian-inference}. The simulations
in section \ref{sec:Simulations} will illustrate the superior performance
of exact MAP estimate to the asymptotic AIC method in linear sensor
array, particularly for overlapping multi-tone sources. The paper
is concluded in section \ref{sec:Conclusion}. 

\section{Sensor array's model\label{sec:Sensor-array}}

\begin{figure*}
\begin{equation}
\underset{\Data_{\ndim\times\ntime}}{\underbrace{\left[\begin{array}{ccc}
\overset{\xbold_{1}}{\overbrace{\ \data_{1,1}\ }} & \cdots & \overset{\xbold_{\ntime}}{\overbrace{\ \data_{1,\ntime}\ }}\\
\vdots &  & \vdots\\
\data_{\ndim,1} & \cdots & \data_{\ndim,\ntime}
\end{array}\right]}}=\underset{\bULA_{\ndim\times\nstate}}{\underbrace{\left[\begin{array}{ccc}
\overset{\bULa_{\angleDOA_{1}}}{\overbrace{\ e^{j\angleDOA_{1}}\ }} & \cdots & \overset{\bULa_{\angleDOA_{\nstate}}}{\overbrace{\ e^{j\angleDOA_{\nstate}}\ }}\\
\vdots &  & \vdots\\
e^{j\angleDOA_{1}\ndim} & \cdots & e^{j\angleDOA_{\nstate}\ndim}
\end{array}\right]}}\overset{\Signal_{\nstate\times\ntime}}{\overbrace{\underset{\Amplitude_{\nstate\times\nsource}}{\underbrace{\left[\begin{array}{ccc}
\bamplitude_{1}\left\{ \begin{array}{c}
\\
\\
\end{array}\amplitude_{1,1}\right. & \cdots & \amplitude_{1,\nsource}\\
\ \ \ \ \ \ \ \vdots &  & \vdots\\
\bamplitude_{\nstate}\left\{ \begin{array}{c}
\\
\\
\end{array}\amplitude_{\nstate,1}\right. & \cdots & \amplitude_{\nstate,\nsource}
\end{array}\right]}}\underset{\Source_{\nsource\times\ntime}}{\underbrace{\left[\begin{array}{ccc}
e^{j\sFreq_{1}} & \cdots & e^{j\sFreq_{1}\ntime}\\
\vdots &  & \vdots\\
e^{j\sFreq_{\nsource}} & \cdots & e^{j\sFreq_{\nsource}\ntime}
\end{array}\right]}}}}+\Noise_{\ndim\times\ntime}\label{eq:MATRIX_MODEL}
\end{equation}
\end{figure*}
In linear sensor array's model, as illustrated in equation (\ref{eq:MATRIX_MODEL})
at the top of next page, let $\Data\TRIANGLEQ\setv{\xbold}{\ntime}\in\COMPLEX^{\ndim\times\ntime}$
and $\Signal\TRIANGLEQ\setv{\bsignal}{\ntime}\in\COMPLEX^{\nstate\times\ntime}$
denote the complex matrix of $\ndim$ sensors' output and $\nstate$
multi-tone sources over $\ntime$ time points, respectively. Hence,
at time $\seti{\itime}{\ntime}$, the column vectors $\xbold_{\itime}\TRIANGLEQ[\data_{1,\itime},\ldots,\data_{\ndim,\itime}]^{\transpose}\in\COMPLEX^{\ndim\times1}$
and $\bsignal_{\itime}\TRIANGLEQ[\signal_{1,\itime},\ldots,\signal_{\nstate,\itime}]^{\transpose}\in\COMPLEX^{\nstate\times1}$
are complex-value snapshots of $\ndim$ sensors' output and unknown
$\nstate$ multi-tone sources, respectively, where $\transpose$ denotes
transpose operator.

Let $\ULA\TRIANGLEQ[\bULa(\angleDOA_{1}),\ldots,\bULa(\angleDOA_{\nstate})]\in\calV_{\ndim\times\nstate}\subset\COMPLEX^{\ndim\times\nstate}$
denote the steering array matrix, whose $\{\idim,\istate\}$-element
is $\ULa_{\idim,\istate}\TRIANGLEQ e^{j2\radius_{\idim}\angleDOA_{\istate}}=e^{j\angleDOA_{\istate}\idim}$,
with radius $\radius_{\idim}\TRIANGLEQ\frac{\varrho}{2}\idim\in\REAL$
denoting positions of $\ndim$ sensors spaced at half of the unit
wavelength $\varrho=1$ and DOA's spatial angular frequencies $\angleDOA_{\istate}\TRIANGLEQ\pi\cos\Angle_{\istate}\in[-\pi,\pi)$
corresponding to upper half-space arrival angle $\Angle_{\istate}\in[0,180^{0})$,
$\forall\seti{\idim}{\ndim}$, $\forall\seti{\istate}{\nstate}$. 

Let us also assume that each source is a superposition of at most
$\nsource$ potential tones over time. In matrix form, we write $\Signal=\Amplitude\Source$,
in which $\Amplitude\TRIANGLEQ\setv{\bamplitude}{\nstate}^{\transpose}\in\COMPLEX^{\nstate\times\nsource}$
is the matrix of $\nstate$ source's complex amplitudes $\bamplitude_{\istate}\TRIANGLEQ[\amplitude_{\istate,1},\ldots,\amplitude_{\istate,\nsource}]^{\transpose}$,
$\forall\seti{\istate}{\nstate}$, and $\Source\TRIANGLEQ\setv{\sbold}{\nsource}^{\transpose}\in\COMPLEX^{\nsource\times\ntime}$
is the matrix of each source's $\nsource$ potential tones $\sbold_{\isource}\TRIANGLEQ[e^{j\sFreq_{\isource}},\ldots,e^{j\sFreq_{\isource}\ntime}]^{\transpose}$,
$\forall\seti{\isource}{\nsource}$. In element form, we have $\signal_{\istate,\itime}=\sum_{\isource=1}^{\nsource}\amplitude_{\istate,\isource}\source_{\isource,\itime}$,
where $\amplitude_{\istate,\isource}\in\COMPLEX$ is complex amplitude
of $\isource$-th tone $\source_{\isource,\itime}\TRIANGLEQ e^{j\sFreq_{\isource}\itime}$
and $\sFreq_{\isource}\in[0,2\pi)$ is tone's angular frequency. 

In order to apply the fast Fourier transform (FFT), let us assume
that all tone's frequency falls into discrete Fourier transform (DFT)
bins $\frac{2\pi}{\ntime}$, i.e. $\sFreq_{\isource}\in\left\{ 0,\frac{2\pi}{\ntime}\ldots,(\ntime-1)\frac{2\pi}{\ntime}\right\} $. 

\subsection{Direction of arrival (DOA) model}

In time domain, the data model for the linear sensor array is then
written in matrix form in (\ref{eq:MATRIX_MODEL}), as follows: 
\begin{align}
\Data & =\ULA\Signal+\Noise,\ \text{with}\ \Signal=\Amplitude\Source,\label{eq:MATRIX_FORM}
\end{align}
where $\Noise\in\COMPLEX^{\ndim\times\ntime}$is matrix of complex
AWGN with power~$\varsignal^{2}$. 

\begin{figure}
\begin{centering}
\includegraphics[width=1\columnwidth]{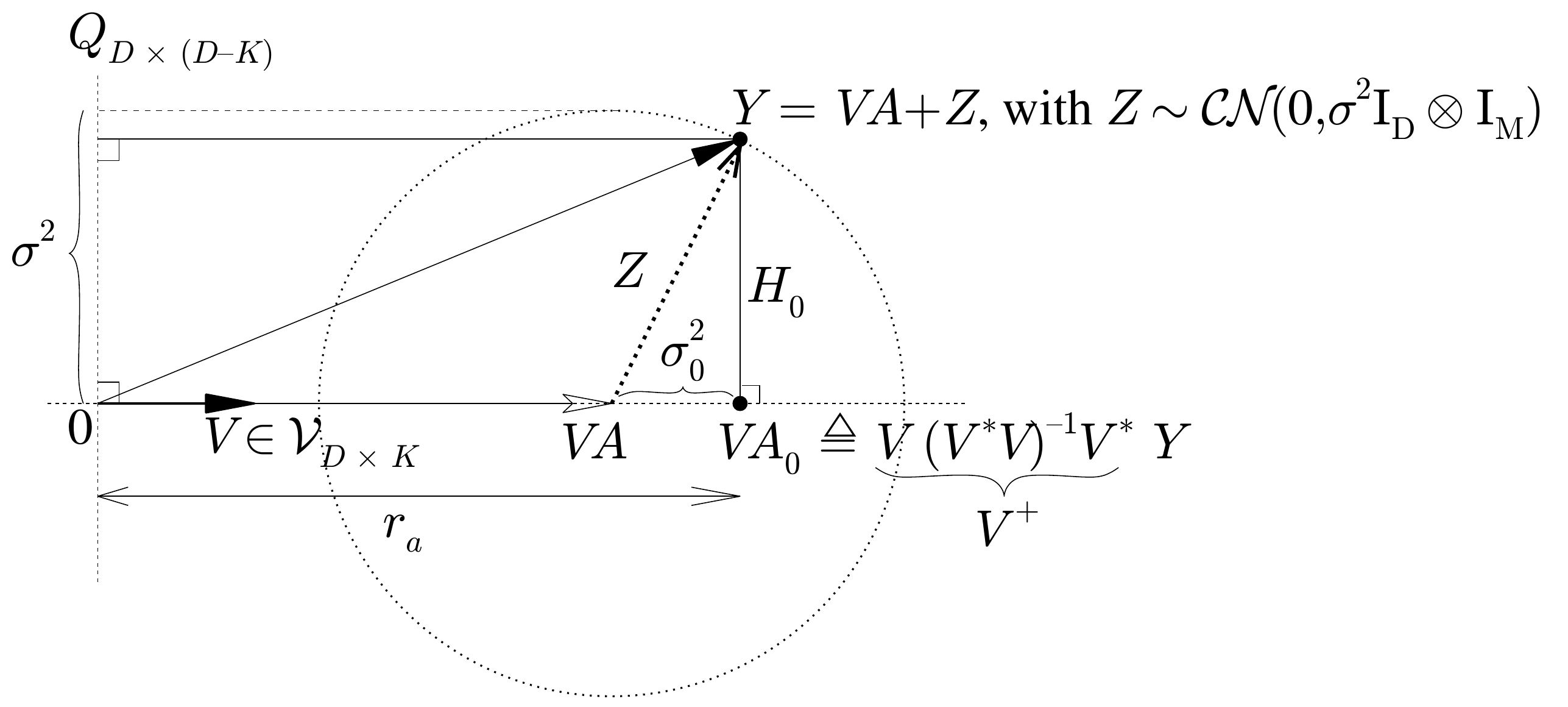}
\par\end{centering}
\caption{\label{fig:PCA_MUSIC}In likelihood model $\protect\yData\sim\protect\calCN(\protect\bULA\protect\Amplitude,\protect\stderror^{2}\protect\IDEN_{\protect\ndim}\otimes\protect\IDEN_{\protect\nsource})$,
the ground-truth $\protect\bULA\protect\Amplitude$ is regarded as
the reference point. Then $\protect\eNoise$ and $\protect\yData$
follow Gaussian distribution around $\protect\bULA\protect\Amplitude$,
with variance $\protect\stderror^{2}$. In contrast, Bayesian method
regards data $\protect\yData$ as the reference point and, given non-informative
priors, we can also say that the unknown quantities $\protect\eNoise$
and $\protect\bULA\protect\Amplitude$ follow Gaussian distribution
around $\protect\yData$, with the same variance $\protect\stderror^{2}$.
Similarly, both amplitude $\protect\Amplitude$ and its estimate $\protect\AmpZero\protect\TRIANGLEQ\protect\bULA^{+}\protect\yData$
follow Gaussian distribution around each other, with variance $\protect\stdZero^{2}$
in (\ref{eq:ra_PCA}, \ref{eq:condA}). If $\protect\nstate<\protect\ndim$
and $\protect\bULA^{\protect\Hermitian}\protect\bULA$ is diagonal,
the PCA in (\ref{eq:Vmax}, \ref{eq:V_PCA}) returns MAP estimate
$\widehat{\protect\bULA}\protect\TRIANGLEQ\protect\argmax_{\protect\bULA}\protect\NormHS{\protect\bULA\protect\AmpZero}=\protect\argmin_{\protect\bULA}\protect\NormHS{\protect\HZero}=\protect\yVECeigen_{1:\protect\nstate}$,
where $\protect\yVECeigen\protect\TRIANGLEQ[\protect\yVECeigen_{1:\protect\nstate},\protect\yVECeigen_{\protect\nstate+1:\protect\ndim}]$
are orthogonal eigenvectors of $\protect\yData\protect\yData^{\protect\Hermitian}\in\protect\COMPLEX^{\protect\ndim\times\protect\ndim}$
with $\protect\nstate$ highest and $\protect\ndim-\protect\nstate$
lowest eigenvalues, respectively. Intuitively, the MAP estimate of
dimension $\protect\nstate$ of signal subspace $\mathcal{V}_{\protect\ndim\times\protect\nstate}$
corresponds to the highest probability that signal-plus-noise variance
$\protect\varSignal$ still dominates projected noise's variance $\protect\stdZero^{2}$
on that subspace, as shown in (\ref{eq:Likeli=00005BV,K=00005Da},
\ref{eq:K_MAP_Stiefel}).}
\end{figure}

In frequency domain, we can multiply $\frac{\Source^{*}}{\ntime}$
from the right in (\ref{eq:MATRIX_MODEL}-\ref{eq:MATRIX_FORM}) and
rewrite our DOA's model as follows: 
\begin{equation}
\yData=\ULA\Amplitude+\eNoise,\label{eq:FFT_FORM}
\end{equation}
which is also a problem of factor analysis with unknown matrix product
$\ULA\Amplitude$, as illustrated in Fig. \ref{fig:PCA_MUSIC}. Since
all source's tones are evaluated at DFT bins in FFT method, the FFT
covariance matrix of tone components is diagonal, i.e. $\Source\Source^{\Hermitian}=\ntime\IDEN_{\nsource}$,
in which $\Hermitian$ denotes conjugate transpose and $\IDEN_{\nsource}$
is an $\nsource\times\nsource$ identity matrix. Then $\yData\TRIANGLEQ\frac{1}{\ntime}\Data\Source{}^{\Hermitian}=\setv{\ybold}{\nsource}\in\COMPLEX^{\ndim\times\nsource}$
is the normalized FFT output of the array data. The noise $\eNoise\TRIANGLEQ\Noise\frac{\Source^{*}}{\ntime}$
in (\ref{eq:FFT_FORM}) is also a complex AWGN with power $\stderror^{2}\TRIANGLEQ\frac{\varsignal^{2}}{\ntime}$. 

Given noisy data $\Data$ and its form $\yData$ in frequency domain
(\ref{eq:FFT_FORM}), our aim is then to estimate all unknown parameters
$\{\Amplitude,\bangleDOA,\nstate,\stderror\}$, where $\bangleDOA\TRIANGLEQ\setv{\angleDOA}{\nstate}^{\transpose}\in[-\pi,\pi)^{\nstate}$
are DOA's spatial frequencies of $\nstate$ sources.

\subsection{Uncorrelated condition for DOAs}

\begin{figure}
\begin{centering}
\includegraphics[width=0.55\columnwidth]{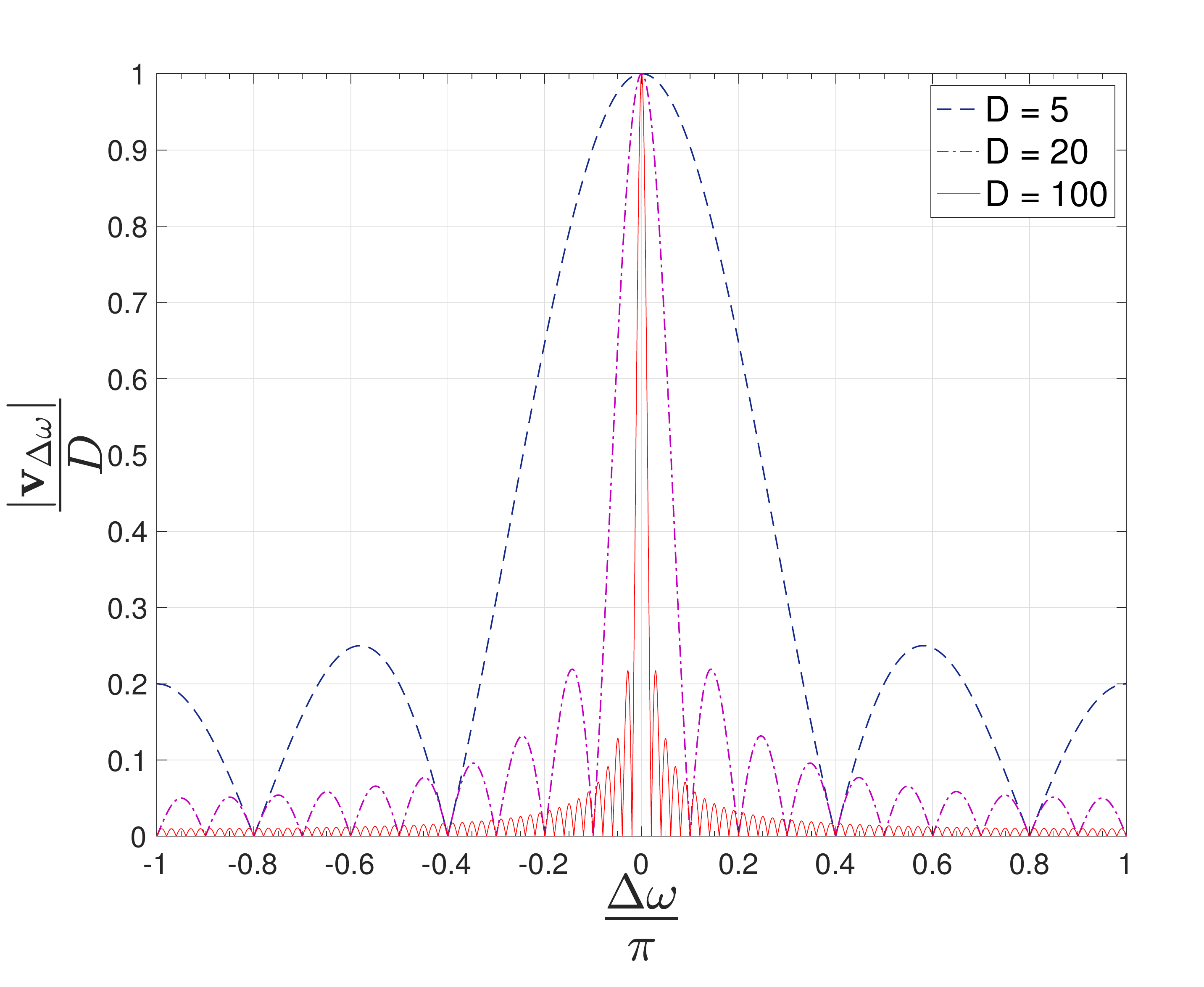}\includegraphics[width=0.45\columnwidth]{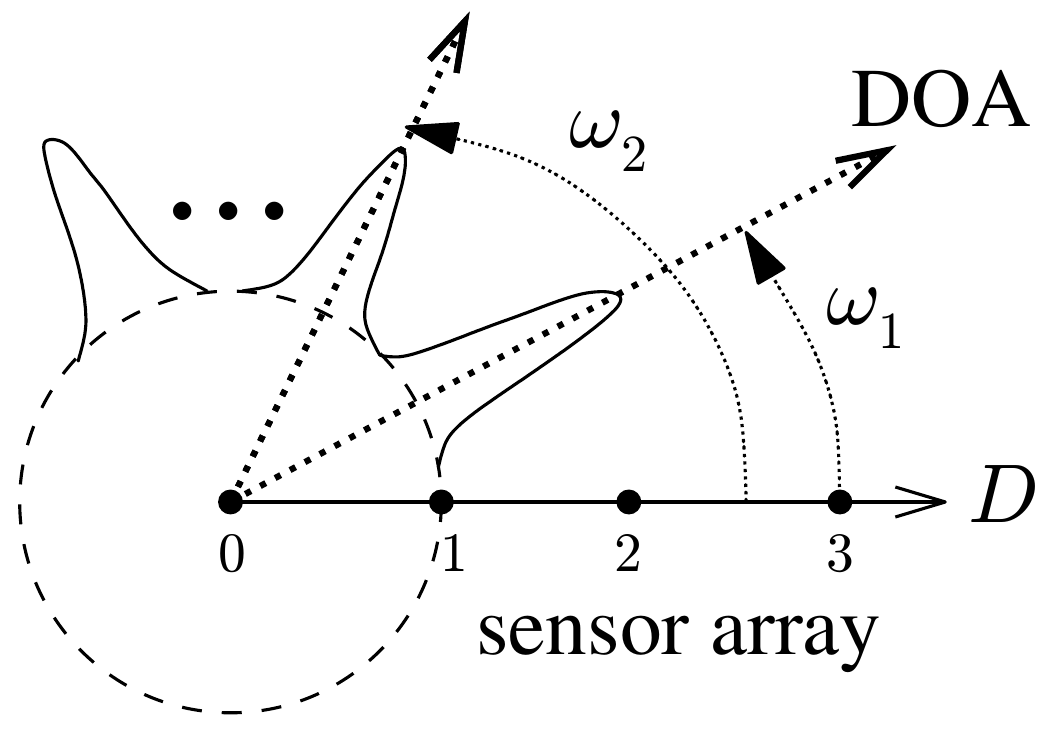}
\par\end{centering}
\caption{\label{fig:DOA} (Left) $\protect\bULa_{\Delta\protect\angleDOA}$
denotes inner product between two steering vectors in $\protect\ULA$,
as given in (\ref{eq:uncorrDOA}) . The absolute value $|\protect\bULa_{\Delta\protect\angleDOA}|$
is zero at multiple integers of $\protect\threshEnd\protect\TRIANGLEQ\frac{2\pi}{\protect\ndim}$,
where $\protect\ndim$ is the number of sensors. (Right) Illustration
of posterior distributions of uncorrelated DOAs $\protect\bangleDOA$
\cite{VH:ICASSP:18}.}
\end{figure}

For later use, the DOA's uncorrelated condition $\orderCond$ is defined
in this paper as follows: 
\begin{equation}
\orderCond:\frac{\ULA^{\Hermitian}\ULA}{\ndim}=\IDEN_{\nstate},\label{eq:uncorrV}
\end{equation}
which corresponds to orthogonality of steering vectors, i.e. $\forall\istate\neq\isource$:
\begin{equation}
\bULa_{\Delta\angleDOA_{\istate,\isource}}\TRIANGLEQ\bULa_{\angleDOA_{\istate}}^{*}\bULa_{\angleDOA_{\isource}}=\sum_{\idim=1}^{\ndim}e^{j\Delta\angleDOA_{\istate,\isource}\idim}=\frac{\sin\left(\frac{\Delta\angleDOA_{\istate,\isource}}{2}\ndim\right)}{\sin\left(\frac{\Delta\angleDOA_{\istate,\isource}}{2}\right)}=0,\label{eq:uncorrDOA}
\end{equation}
with $\Delta\angleDOA_{\istate,\isource}\TRIANGLEQ\angleDOA_{\isource}-\angleDOA_{\istate}$,
$\forall\seti{\istate,\isource}{\nstate}$. The sin function in (\ref{eq:uncorrDOA})
is owing to the fact that the value $\bULa_{\Delta\angleDOA_{\istate,\isource}}$
is equivalent to a discrete time Fourier transform (DTFT) of a unit
rectangular function over $[0,\ndim]$, since each steering vector
$\bULa_{\angleDOA_{\istate}}$ can be regarded as a discrete sequence
of complex sinusoidal values over $\ndim$ sensors, as defined in
(\ref{eq:MATRIX_MODEL}). Because $\bULa_{\Delta\angleDOA_{\istate,\isource}}$
in (\ref{eq:uncorrDOA}) is only zero at multiple integers of $\threshEnd\TRIANGLEQ\frac{2\pi}{\ndim}$,
we can see that there is the so-called power leakage if $\Delta\angleDOA_{\istate,\isource}$
is not multiple integers of $\frac{2\pi}{\ndim}$, as shown in Fig.~\ref{fig:DOA}. 

For later use, let us also relax the uncorrelated condition (\ref{eq:uncorrV})
in weaker form, as follows: 
\begin{align}
\differCond:\frac{\ULA^{\Hermitian}\ULA}{\ndim}\approx\IDEN_{\nstate} & \Leftrightarrow0\leq\frac{|\bULa_{\Delta\angleDOA_{\istate,\isource}}|}{\ndim}\ll1,\forall\istate\neq\isource,\label{eq:uncorrApprox}
\end{align}
This weaker condition will be used in Bayesian analysis of the MUSIC
algorithm below. 


\section{PCA and MUSIC algorithms\label{sec:PCA}}

Given uncorrelated condition (\ref{eq:uncorrV}), the DOA's model
(\ref{eq:FFT_FORM}) is essentially a special case of traditional
PCA method. Indeed, given an observation matrix $\yData\in\COMPLEX^{\ndim\times\nsource}$
and AWGN $\eNoise$ in (\ref{eq:FFT_FORM}) , the PCA's purpose is
to estimate its orthogonal principal vectors $\bULA\in\COMPLEX^{\ndim\times\nstate}$
and the complex amplitudes $\Amplitude\in\COMPLEX^{\nstate\times\nsource}$,
as follows:
\begin{equation}
\yData=\bULA\Amplitude+\eNoise,\ \text{with}\ \begin{cases}
\eNoise\sim\calCN(\ZERO,\stderror^{2}\IDEN_{\ndim}\otimes\IDEN_{\nsource})\\
\bULA^{\Hermitian}\bULA=\ndim\IDEN_{\nstate}\ \text{and}\ \nstate<\ndim
\end{cases},\label{eq:PCA}
\end{equation}
where $\ZERO$ denotes zero matrix with appropriate dimensions, as
illustrated in Fig. \ref{fig:PCA_MUSIC}. In this section, let us
briefly review the traditional PCA and MUSIC algorithms for estimating
$\bULA$ and $\ULA$ in (\ref{eq:PCA}) and (\ref{eq:FFT_FORM}),
respectively. 

\subsection{Euclidean formulation\label{subsec:Probabilistic-model}}

Since the noise $\eNoise$ is Gaussian, let us interpret the PCA model
(\ref{eq:PCA}) via Euclidean distance first, as shown in Fig. \ref{fig:PCA_MUSIC},
before applying Bayesian method in next section. From (\ref{eq:PCA}),
we have: 
\begin{align}
\Amplitude & =\bULA^{+}(\yData-\eNoise),\ \text{with}\ \bULA^{+}\TRIANGLEQ(\bULA^{\Hermitian}\bULA)^{-1}\bULA^{\Hermitian},\label{eq:pseudo_FORM}
\end{align}
where $\bULA^{+}$ is the Moore-Penrose pseudo-inverse of $\bULA$,
i.e. $\bULA^{+}\bULA=\IDEN_{\nstate}$. Then, from (\ref{eq:pseudo_FORM}),
we can see that $\AmpZero\TRIANGLEQ\bULA^{+}\yData$ is the conditional
mean estimate of $\Amplitude$, given $\bULA$, since $\EXPECTATION(\eNoise)=\ZERO$. 

As illustrated in Fig. \ref{fig:PCA_MUSIC}, we note that\footnote{The Pythagorean equality in (\ref{eq:Pythagore}) can also be verified
directly, as follows: $(\yData-\bULA\Amplitude)^{\Hermitian}(\yData-\bULA\Amplitude)=\Amplitude^{*}(\bULA^{*}\bULA)\Amplitude-\Amplitude^{*}\bULA^{\Hermitian}\yData-\yData^{\Hermitian}\bULA\Amplitude+\yData^{*}\yData=(\Amplitude-\pseudoDOA\yData)^{*}(\bULA^{*}\bULA)(\Amplitude-\pseudoDOA\yData)+\yData^{*}\yData-\yData^{*}(\bULA\pseudoDOA)\yData$. }: 
\begin{equation}
\NormSquared{\eNoise}=\NormSquared{\yData-\bULA\Amplitude}=\NormSquared{\bULA(\Amplitude-\AmpZero)}+\NormSquared{\HZero},\label{eq:Pythagore}
\end{equation}
in which $\NormHS{\cdot}$ denotes the length (i.e. Hilbert--Schmidt
norm \cite{Hilbert_Schmidt:inequality:81}) operator: $\NormHS{\yData}^{2}\TRIANGLEQ\Trace(\yData^{\Hermitian}\yData)$,
with $\Trace(\cdot)$ denoting Trace operator. The term $\HZero\TRIANGLEQ\yData-\bULA\AmpZero$
represents the height between data $\yData$ and its projection on
vector space $\bULA\in\calV_{\ndim\times\nstate}$, as follows:
\begin{equation}
\NormSquared{\HZero}=\NormSquared{\yData-\bULA\AmpZero}=\NormSquared{\yData}-\NormSquared{\bULA\AmpZero},\label{eq:Height}
\end{equation}
where: 
\begin{align}
\NormSquared{\bULA\AmpZero} & =\Trace(\yData^{*}(\bULA\pseudoDOA)\yData)=\Trace(\bULA\pseudoDOA(\yData\yData^{*}))\nonumber \\
 & =\Trace(\pseudoDOA(\yData\yData^{\Hermitian})\bULA),\label{eq:VA}
\end{align}
since the trace is invariant under cyclic permutations. These Pythagorean
forms (\ref{eq:Pythagore}, \ref{eq:Height}) will simplify the derivation
in equations (\ref{eq:condA}, \ref{eq:postDOAsigma}) below.

\subsection{Principal component analysis (PCA)}

Let us now substitute the diagonal condition $\bULA^{\Hermitian}\bULA=\ndim\IDEN_{\nstate}$
in (\ref{eq:PCA}) to (\ref{eq:pseudo_FORM}-\ref{eq:VA}):
\begin{equation}
\bULA^{+}=\frac{\bULA^{\Hermitian}}{\ndim},\AmpZero=\bULA^{+}\yData=\frac{\bULA^{\Hermitian}\yData}{\ndim},\NormSquared{\bULA\AmpZero}=\ndim\NORMTWO{\AmpZero}.\label{eq:VV_forms}
\end{equation}
The estimate $\widehat{\bULA}$ closest to data $\yData$, as illustrated
in Fig. \ref{fig:PCA_MUSIC}, can be computed from (\ref{eq:Height},
\ref{eq:VA}), as follows:
\begin{align}
\widehat{\bULA} & \TRIANGLEQ\argmax_{\bULA\in\calS_{\ndim}^{\nstate}}\NormSquared{\bULA\AmpZero}=\argmin_{\bULA\in\calS_{\ndim}^{\nstate}}\NORMTWO{\HZero}\nonumber \\
 & =\argmax_{\bULA\in\calS_{\ndim}^{\nstate}}\Trace(\bULA^{\Hermitian}(\yData\yData^{\Hermitian})\bULA)=\argmax_{\bULA\in\calS_{\ndim}^{\nstate}}\sum_{\istate=1}^{\nstate}\bULa_{\istate}^{\Hermitian}(\yData\yData^{\Hermitian})\bULa_{\istate}\nonumber \\
 & =\argmax_{\bULA\in\calS_{\ndim}^{\nstate}}\sum_{\istate=1}^{\nstate}\sum_{\idim=1}^{\ndim}\yeigen_{\idim}\NORMTWO{\yVecEigen_{\idim}^{\Hermitian}\bULa_{\istate}}.\label{eq:Vmax}
\end{align}
in which $\calS_{\ndim}^{\nstate}\TRIANGLEQ\{\bULA\in\COMPLEX^{\ndim\times\nstate}:\ \bULA^{\Hermitian}\bULA=\StiefelRadius^{2}\IDEN_{\nstate}\}$
denotes the Stiefel manifold with radius $\StiefelRadius\TRIANGLEQ\sqrt{\ndim}$
and $\yVecEigen_{\idim}\in\COMPLEX^{\ndim\times1}$ are orthogonal
eigenvectors of positive semi-definite covariance matrix $\yData\yData^{\Hermitian}\in\COMPLEX^{\ndim\times\ndim}$
with $\idim$-th highest eigenvalues $\yeigen_{\idim}$. Since the
amplitudes of both component vectors $\bULa_{\istate}$ and eigenvectors
are constant, i.e. $\bULa_{\istate}^{\Hermitian}\bULa_{\istate}=\ndim$
and $\yVecEigen_{\idim}^{\Hermitian}\yVecEigen_{\idim}=1$, $\forall\idim,\istate$,
the inner product $\NORMTWO{\yVecEigen_{\idim}^{\Hermitian}\bULa_{\istate}}$
in (\ref{eq:Vmax}) is maximized when $\bULa_{\istate}=\yVecEigen_{\idim}$.
Then, if $\nstate<\ndim$, solving (\ref{eq:Vmax}) for $\bULA$ yields:
\begin{equation}
\widehat{\bULA}=\argmin_{\bULA\in\calS_{\ndim}^{\nstate}}\NORMTWO{\yVECeigen_{\nstate+1:\ndim}^{\Hermitian}\bULA}=\yVECeigen_{1:\nstate}\TRIANGLEQ[\yVecEigen_{1},\ldots,\yVecEigen_{\nstate}].\label{eq:V_PCA}
\end{equation}
The $\nstate$ eigenvectors $\yVECeigen_{1:\nstate}$ are essentially
the output of traditional PCA algorithm and $\yVECeigen_{\nstate+1:\ndim}\TRIANGLEQ[\yVecEigen_{\nstate+1},\ldots,\yVecEigen_{\ndim}]\in Q_{\ndim\times(\ndim-\nsource)}$
is called residue eigen-subspace, as illustrated in Fig. \ref{fig:PCA_MUSIC}. 

\subsection{MUSIC algorithm}

Similar to PCA, the aim of the MUSIC algorithm is to find the estimate
$\widehat{\bangleDOA}$ of DOAs $\bangleDOA$, such that $\widehat{\bangleDOA}\TRIANGLEQ\argmax_{\bangleDOA}\NormSquared{\ULA\AmpZero}$.
Since the pseudo-inverse form $\ULA^{+}$ in (\ref{eq:VA}) is complicated,
MUSIC assumes the weakly uncorrelated form (\ref{eq:uncorrApprox}),
i.e. $\ULA^{+}\approx\frac{\bULA^{\Hermitian}}{\ndim}$. Then, similar
to (\ref{eq:Vmax}, \ref{eq:V_PCA}), we have: 
\begin{align}
\widehat{\bangleDOA} & =\argmax_{\bangleDOA\in[-\pi,\pi)^{\nstate}}\NormSquared{\ULA\AmpZero}\approx\argmax_{\bangleDOA\in[-\pi,\pi)^{\nstate}}\frac{1}{\NORMTWO{\yVECeigen_{\nstate+1:\ndim}^{\Hermitian}\ULA}}\nonumber \\
 & =\argmax_{\bangleDOA\in[-\pi,\pi)^{\nstate}}\frac{1}{\sum_{\istate=1}^{\nstate}\sum_{\idim=\nstate+1}^{\ndim}\NORMTWO{\yVecEigen_{\idim}^{\Hermitian}\bULa_{\istate}(\angleDOA_{\istate})}}.\label{eq:MUSIC}
\end{align}
Since the steering matrix $\ULA$ has a restricted form over DOAs
$\bangleDOA$, as defined in section \ref{sec:Sensor-array}, the
optimal matrix $\bULA_{\widehat{\bangleDOA}}$ in (\ref{eq:MUSIC})
is not equal to eigenvectors $\yVECeigen_{1:\nstate}$ like the PCA
method (\ref{eq:V_PCA}) and, hence, the denominator $\NORMTWO{\yVECeigen_{\nstate+1:\ndim}^{\Hermitian}\bULA_{\widehat{\bangleDOA}}}$
in (\ref{eq:MUSIC}) is not equal to zero in general. Nonetheless,
since steering vectors $\bULa_{\istate}(\angleDOA)=\bULa(\angleDOA)$
have the same functional form for any $\istate$, the optimal DOAs
$\widehat{\bangleDOA}=\setv{\widehat{\angleDOA}}{\nstate}^{\transpose}$
in (\ref{eq:MUSIC}) correspond to the $\nstate$ highest peaks $\widehat{\angleDOA}$
of the so-called pseudo-spectrum, defined as follows: 
\begin{equation}
\widehat{\angleDOA}\TRIANGLEQ\argmax_{\angleDOA\in[-\pi,\pi)}\frac{1}{\sum_{\idim=\nstate+1}^{\ndim}\NORMTWO{\yVecEigen_{\idim}^{\Hermitian}\bULa(\angleDOA)}}.\label{eq:pseudo_spectrum}
\end{equation}
If $\nstate$ is unknown, the true value $\nstate$ in (\ref{eq:pseudo_spectrum})
is replaced by a threshold $\threshK$, with $\nstate<\threshK\leq\ndim$,
in practice. 

\subsection{DTFT spectrum method}

If we assume the strictly uncorrelated condition $\ULA^{\Hermitian}\ULA=\ndim\IDEN_{\nstate}$,
i.e. $\NormSquared{\ULA\AmpZero}=\ndim\NormSquared{\AmpZero}$, in
(\ref{eq:uncorrV}, \ref{eq:VV_forms}), the DOA's estimate in (\ref{eq:MUSIC})
can be computed via DTFT method (\ref{eq:uncorrDOA}), as follows:
\begin{align}
\widehat{\bangleDOA}_{0} & \TRIANGLEQ\argmax_{\bangleDOA\in[-\pi,\pi)^{\nstate}}\NormSquared{\AmpZero}=\argmax_{\bangleDOA\in[-\pi,\pi)^{\nstate}}\NORMTWO{\ULA^{\Hermitian}\yData}\nonumber \\
 & =\argmax_{\bangleDOA\in[-\pi,\pi)^{\nstate}}\sum_{\istate=1}^{\nstate}\NormSquared{\overline{\bamplitude}_{\istate}}=\argmax_{\bangleDOA\in[-\pi,\pi)^{\nstate}}\sum_{\istate=1}^{\nstate}\NORMTWO{\bULa_{\angleDOA_{\istate}}^{\Hermitian}\yData},\label{eq:DOA_DTFT}
\end{align}
where $\overline{\bamplitude}_{\istate}$ is the $\istate$-th row
of $\AmpZero$. Then $\widehat{\bangleDOA}_{0}$ corresponds to the
$\nstate$ highest peaks of power spectrum $\NORMTWO{\bULa_{\angleDOA}^{\Hermitian}\yData}$
of $\yData$, $\forall\angleDOA\in[-\pi,\pi)$. Hence, we can regard
the spectrum method as a special case of PCA (\ref{eq:VV_forms},
\ref{eq:Vmax}) and MUSIC algorithm (\ref{eq:MUSIC}) for strictly
uncorrelated DOAs. 

\section{Bayesian inference of number of components\label{sec:Bayesian-inference}}

In this section, we will compute the Bayesian posterior estimate for
all unknown quantities in the PCA model (\ref{eq:PCA}). Owing to
our novel double inverse-gamma distribution in Appendix~\ref{sec:Double-Gamma},
we will be able to marginalize out the unknown noise's variance and
derive, for the first time, the closed-form solution for MAP estimate
of the number of components in PCA and MUSIC algorithms at the end
of this section. 

For this purpose, let us rewrite the PCA model (\ref{eq:PCA}) in
normalized form, as follows: 
\begin{equation}
\frac{\yData}{\varSignal}=\frac{\bULA\Amplitude}{\stdamplitude^{2}}(1-\ratioNoise)+\ratioNoise\frac{\eNoise}{\stdZero^{2}},\ \begin{cases}
\ \ratioNoise\TRIANGLEQ\frac{\stdZero^{2}}{\varSignal}=\frac{\stdZero^{2}}{\stdamplitude^{2}+\stdZero^{2}}\\
\stdZero^{2}\TRIANGLEQ\frac{\stderror^{2}}{\NormHS{\bULA}}=\frac{\stderror^{2}}{\ndim}
\end{cases},\label{eq:ra_PCA}
\end{equation}
in which $\varSignal\TRIANGLEQ\stdamplitude^{2}+\stdZero^{2}$ is
called signal-plus-noise variance, $\stdamplitude^{2}\TRIANGLEQ\sum_{\istate=1}^{\nstate}\frac{\sigma_{\istate}^{2}}{\nstate}=\frac{\NormHS{\Amplitude}^{2}}{\nstate\nsource}$
is the empirical amplitude's variance of all sources, $\sigma_{\istate}^{2}\TRIANGLEQ\frac{\NormHS{\bamplitude_{\istate}}^{2}}{\nsource}$
is the empirical amplitude's variance of the $\istate$-th source
in (\ref{eq:MATRIX_MODEL}), $\stdZero^{2}\TRIANGLEQ\frac{\stderror^{2}}{\NormHS{\bULA}}=\frac{\stderror^{2}}{\ndim}$
is the projected noise's variance on vector space $\bULA$ and $\ratioNoise\TRIANGLEQ\frac{\stdZero^{2}}{\varSignal}\in(0,1)$
is called noise-to-signal percentage in this paper. 

Note that, our definition of $\ratioNoise=(1+\stdamplitude^{2}/\stdZero^{2})^{-1}$
is consistent with definition $\ratioNoise=(1+\SNR)^{-1}$ for the
PCA model, as shown in \cite{PCA:tau:11}. If the signal-to-noise
ratio (SNR) is high, i.e. $\stdamplitude^{2}\gg\stdZero^{2}$, we
have $\ratioNoise\rightarrow0$ and, hence, data $\yData$ leans toward
the signal. In contrast, if SNR is low, i.e. $\stdamplitude^{2}\ll\stdZero^{2}$
and $\ratioNoise\rightarrow1$, the noise dominates signal and, hence,
data $\yData$ leans toward the noise~$\eNoise$. If $\nstate=0$,
we set $\ratioNoise=1$ (i.e. $\stdamplitude^{2}=0$, since $\NormHS{\bULA\Amplitude}=0$)
and, hence, data $\yData$ consists of noise $\eNoise$ only.

\subsection{Likelihood model}

The likelihood in (\ref{eq:PCA}, \ref{eq:ra_PCA}) is a complex Gaussian
matrix-variate distribution, as follows:
\begin{align}
f(\yData|\Amplitude,\bULA,\nstate,\stderror^{2}) & =\begin{cases}
\calCN_{\yData}\left(\bULA\Amplitude,\stderror^{2}\IDEN_{\ndim}\otimes\IDEN_{\nsource}\right), & \text{if}\ \nstate>0\\
\calCN_{\yData}\left(\ZERO,\stderror^{2}\IDEN_{\ndim}\otimes\IDEN_{\nsource}\right), & \text{if}\ \nstate=0
\end{cases}\label{eq:Likeli_Gauss}
\end{align}
in which $\otimes$ denotes Kronecker product and $\calCN_{\yData}\left(\bULA\Amplitude,\stderror^{2}\IDEN_{\ndim}\otimes\IDEN_{\nsource}\right)\TRIANGLEQ\frac{\etr(-\IDEN_{\nsource}^{-1}(\yData-\bULA\Amplitude)^{\Hermitian}\IDEN_{\ndim}^{-1}(\yData-\bULA\Amplitude)/\stderror^{2})}{(\pi\stderror^{2})^{\ndim\nsource}}$
(c.f. \cite{ComplexGauss:BOOK:95}), with $\etr(\cdot)\TRIANGLEQ\exp(\Trace(\cdot))$
denoting the exponential trace operator and, hence, $f(\yData|\nstate=0,\stderror^{2})=\frac{\exp(-\NormSquared{\yData}/\stderror^{2})}{(\pi\stderror^{2})^{\ndim\nsource}}$.

\subsection{Non-informative prior for amplitudes}

Let us consider the non-informative Jeffreys' prior for $\Amplitude$
first, i.e. $f(\Amplitude|\nstate)\propto\frac{1}{\zeta^{\nstate\nsource}}$
with sufficiently large normalizing constant $\zeta$ (ideally $\zeta\rightarrow\infty$),
$\forall\nstate>0$, and $f(\Amplitude|\nstate=0)=\delta(\Amplitude)$,
with Dirac-delta function $\delta(\cdot)$. The posterior for $\Amplitude$
can be derived from (\ref{eq:Likeli_Gauss}), as follows: 
\begin{align}
f(\yData,\Amplitude & |\bULA,\nstate,\stderror^{2})=f(\yData|\Amplitude,\bULA,\nstate,\stderror^{2})f(\Amplitude|\nstate)\label{eq:JointModel}\\
= & \begin{cases}
f(\Amplitude|\yData,\bULA,\nstate,\stderror^{2})f(\yData|\bULA,\nstate,\stderror^{2}), & \text{if}\ \nstate>0\\
f(\yData|\nstate=0,\stderror^{2})\delta(\Amplitude), & \text{if}\ \nstate=0
\end{cases},\nonumber 
\end{align}
and: 
\begin{align}
f(\Amplitude|\yData,\bULA,\nstate,\stderror^{2})= & \calCN_{\Amplitude}(\AmpZero,\UULA^{-1}\otimes\IDEN_{\nsource})\label{eq:condA}\\
= & \frac{1}{\left(\pi\stdZero^{2}\right)^{\nstate\nsource}}\exp\left(-\NormSquared{\bULA(\Amplitude-\AmpZero)}/\stderror^{2}\right),\nonumber 
\end{align}
\begin{align}
f(\yData|\bULA,\nstate,\stderror^{2}) & =\frac{(\pi\stdZero^{2}/\zeta)^{\nstate\nsource}}{(\pi\stderror^{2})^{\ndim\nsource}}\exp\left(-\NormSquared{\HZero}/\stderror^{2}\right)\label{eq:postDOAsigma}\\
 & =\frac{f(\yData|\nstate=0,\stderror^{2})}{\left(\frac{\zeta}{\pi\stdZero^{2}}\right)^{\nstate\nsource}}\exp\left(\NormSquared{\bULA\AmpZero}/\stderror^{2}\right),\nonumber 
\end{align}
in which we have applied the Pythagorean forms (\ref{eq:Pythagore},
\ref{eq:Height}) to (\ref{eq:Likeli_Gauss}, \ref{eq:JointModel}),
with $\UULA\TRIANGLEQ\frac{\bULA^{\Hermitian}\bULA}{\nsigma^{2}}=\frac{\IDEN_{\nstate}}{\stdZero^{2}}$
and $\stdZero^{2}=\sqrt[\nstate]{\det(\UULA^{-1})}=\frac{\nsigma^{2}}{\sqrt[\nstate]{\det(\bULA^{\Hermitian}\bULA)}}=\frac{\stderror^{2}}{\ndim}$.
Note that, if $\nstate=0$ in (\ref{eq:postDOAsigma}), we have: $f(\yData|\bULA,\nstate,\stderror^{2})=f(\yData|\nstate=0,\stderror^{2})$,
since $\NormSquared{\HZero}=\NormSquared{\yData}$ in (\ref{eq:Height})
in this case, owing to convention $\NormSquared{\bULA\AmpZero}=0$
in (\ref{eq:Likeli_Gauss}). 

From (\ref{eq:postDOAsigma}), we can see that, given $\bULA$, the
posterior mean of $\Amplitude$ is $\AmpZero\TRIANGLEQ\pseudoDOA\yData$,
as illustrated in Fig. \ref{fig:PCA_MUSIC}. Hence the conditional
distribution (\ref{eq:condA}) is similar to the pseudo-inverse form
in (\ref{eq:pseudo_FORM}), except that it is now given explicitly
in the form of Gaussian distribution. 

For later use, let us compute the likelihood $f(\yData|\nstate=0)$
by multiplying the non-informative Jeffreys' prior $f(\stderror^{2})\propto\frac{1}{\stderror^{2}}$
with likelihood $f(\yData|\nstate=0,\stderror^{2})$ in (\ref{eq:postDOAsigma}),
as follows:
\begin{align}
f(\yData|\nstate=0) & =\int_{0}^{\infty}f(\yData|\nstate=0,\stderror^{2})f(\stderror^{2})d\stderror^{2}\label{eq:f(Y|K=00003D0)}\\
 & =\frac{1}{\zeta_{\yData}}\TRIANGLEQ\frac{1}{\pi{}^{\ndim\nsource}}\frac{\Gamma(\ndim\nsource)}{(\NORMTWO{\yData})^{\ndim\nsource}}.\nonumber 
\end{align}

\subsection{Conjugate prior for amplitudes}

Note that, the uniform prior $f(\Amplitude|\nstate)\propto\frac{1}{\zeta^{\nstate\nsource}}$
is improper since it implies that the averaged signal's power $\stdamplitude^{2}$
in (\ref{eq:ra_PCA}) is infinite, which is not the case in practice.
For this reason, let us consider a conjugate prior with finite averaged
variance $\stdamplitude^{2}$ for $\Amplitude$, as follows: $f(\Amplitude|\nstate,\stdamplitude^{2})=\calCN_{\Amplitude}(\ZERO,\stdamplitude^{2}\IDEN_{\nstate}\otimes\IDEN_{\nsource})=\frac{1}{(\pi\stdamplitude^{2})^{\nstate\nsource}}\exp\left(\frac{-\NormHS{\Amplitude}^{2}}{\stdamplitude^{2}}\right)$,
which is conjugate to Gaussian model (\ref{eq:condA}), $\forall\nstate>0$.
This prior variance $\stdamplitude^{2}$ represents the unknown value
of averaged signal's power in the PCA's model (\ref{eq:ra_PCA}),
which can be estimated from data $\yData$ a-posteriori. Note that,
if we set $\stdamplitude^{2}\rightarrow\infty$, we have $f(\Amplitude|\nstate)\rightarrow\frac{1}{\zeta^{\nstate\nsource}}=\frac{1}{(\pi\stdamplitude^{2})^{\nstate\nsource}}$
and this conjugate case will return to the case of uniform prior above. 

\subsubsection{Posterior distribution of amplitudes}

Replacing $f(\Amplitude|\nstate)$ in (\ref{eq:JointModel}) with
$f(\Amplitude|\nstate,\stdamplitude^{2})$ yields\footnote{Here we use: $\NormSquared{\bULA(\Amplitude-\AmpZero)}/\stderror^{2}+\NormHS{\Amplitude}^{2}/\stdamplitude^{2}=\NormSquared{\bULA\AmpZero}/\stderror^{2}+\Trace(\Std(\Amplitude-\SULA^{-1}\UULA\AmpZero)^{\Hermitian}(\Amplitude-\SULA^{-1}\UULA\AmpZero))-\Trace(\SULA(\SULA^{-1}\UULA\AmpZero)^{\Hermitian}\SULA^{-1}\UULA\AmpZero)$.}:
\begin{align}
f(\Amplitude|\yData,\bULA,\nstate,\stderror^{2},\stdamplitude^{2}) & =\calCN_{\Amplitude}(\SULA^{-1}\UULA\AmpZero,\SULA^{-1}\otimes\IDEN_{\nsource}),\nonumber \\
f(\yData|\bULA,\nstate,\stderror^{2},\stdamplitude^{2}) & =\frac{\rationoise^{\nstate\nsource}}{(\pi\stderror^{2})^{\ndim\nsource}}\exp\left(-\frac{\NormSquared{\Height_{\SULA}}}{\stderror^{2}}\right),\label{eq:postAmatrix}
\end{align}
where $\SULA\TRIANGLEQ\UULA+\frac{\IDEN_{\nstate}}{\stdamplitude^{2}}$
and:
\begin{align}
\NormSquared{\Height_{\SULA}}\TRIANGLEQ & \NormSquared{\HZero}+\NormSquared{\bULA\AmpZero}\label{eq:H_Sigma}\\
 & -\stderror^{2}\Trace(\SULA\AmpZero(\UULA^{\Hermitian}(\SULA^{\Hermitian}\SULA)^{-1})\UULA\AmpZero).\nonumber 
\end{align}

\subsubsection{Diagonal condition}

If $\bULA^{\Hermitian}\bULA$ is not diagonal, it is not feasible
to factorize the likelihood form $f(\yData|\bULA,\nstate,\stderror^{2},\stdamplitude^{2})$
in (\ref{eq:postAmatrix}). Hence, from (\ref{eq:VV_forms}), substituting
the diagonal forms $\UULA=\frac{\bULA^{\Hermitian}\bULA}{\nsigma^{2}}=\frac{\IDEN_{\nstate}}{\stdZero^{2}}$
and $\SULA=\left(\frac{1}{\stdZero^{2}}+\frac{1}{\stdamplitude^{2}}\right)\IDEN_{\nstate}$
into (\ref{eq:H_Sigma}), we can factorize the likelihood form in
(\ref{eq:postAmatrix}) feasibly, as follows\footnote{We have: $\Trace(\SULA\AmpZero(\UULA^{\Hermitian}(\SULA^{\Hermitian}\SULA)^{-1}\UULA)\AmpZero)=(\SULA^{-1}\UULA^{\Hermitian})\Trace(\AmpZero\UULA\AmpZero)=(1-\rationoise)\NormSquared{\bULA\AmpZero}/\nsigma^{2}$
in this case.}: 
\begin{align}
\NormSquared{\Height_{\SULA}}=\NormSquared{\Height_{\rationoise}} & \TRIANGLEQ\NormSquared{\HZero}+\rationoise\NormSquared{\bULA\AmpZero}\label{eq:ra_Height}\\
 & =\NormSquared{\yData}-(1-\rationoise)\NormSquared{\bULA\AmpZero},\nonumber 
\end{align}
i.e. we have $\NormSquared{\HZero}<\NormSquared{\Height_{\rationoise}}<\NormSquared{\yData},\ \forall\rationoise=\frac{\stdZero^{2}}{\varSignal}\in(0,1).$
Substituting (\ref{eq:ra_Height}) back to (\ref{eq:postAmatrix}),
we obtain: 
\begin{align}
f(\Amplitude|\yData,\bULA,\nstate,\stderror^{2}, & \stdamplitude^{2})=\calCN_{\Amplitude}((1-\rationoise)\AmpZero,\rationoise\stdamplitude^{2}\IDEN_{\nstate}\otimes\IDEN_{\nsource})\nonumber \\
f(\yData|\bULA,\nstate,\stderror^{2},\varSignal) & =\frac{\rationoise^{\nstate\nsource}}{(\pi\stderror^{2})^{\ndim\nsource}}\exp\left(-\frac{\NormSquared{\Height_{\rationoise}}}{\stderror^{2}}\right)\label{eq:ra_Diagonal}\\
 & \propto\calCN_{\NormSquared{\HZero}}\left(0,\stderror^{2}\right)\calCN_{\frac{\NormSquared{\bULA\MeanAmp_{0}}}{\ndim}}\left(0,\varSignal\right),\nonumber 
\end{align}
or, equivalently:
\begin{equation}
f(\yData|\bULA,\nstate,\stderror^{2},\varSignal)=\frac{\caliG_{\varSignal}\left(\iGnx,\frac{\iGsx}{\ndim}\right)\caliG_{\stderror^{2}}\left(\iGny,\iGsy\right)}{\zeta_{\iGp}}\stderror^{2}\varSignal,\label{eq:uncorr_posterior}
\end{equation}
in which $\zeta_{\iGp}$ consists of normalizing constants of inverse-gamma
distributions $\caliG$ and: 
\begin{align}
\iGnx & \TRIANGLEQ\nstate\nsource,\ \iGny\TRIANGLEQ(\ndim-\nstate)\nsource,\ \iGnx+\iGny=\ndim\nsource,\ \iGp\TRIANGLEQ\frac{\iGsx}{\iGsx+\iGsy},\nonumber \\
\iGsx & \TRIANGLEQ\NormSquared{\bULA\AmpZero},\ \iGsy\TRIANGLEQ\NormSquared{\HZero},\ \iGsx+\iGsy=\NORMTWO{\yData},\ \iGq\TRIANGLEQ\frac{\iGsy}{\iGsx+\iGsy},\nonumber \\
\zeta_{\iGp} & \TRIANGLEQ\frac{\pi{}^{\ndim\nsource}\iGsx^{\iGnx}\iGsy^{\iGny}}{\Gamma(\iGnx)\Gamma(\iGny)}=\iGp\iGq\calB_{\iGp}(\iGnx,\iGny)\zeta_{\yData},\ \zeta_{\yData}=\frac{\pi{}^{\ndim\nsource}(\iGsx+\iGsy)^{\iGnx+\iGny}}{\Gamma(\iGnx+\iGny)},\label{eq:ra_LIST1}
\end{align}
with $\zeta_{\yData}$ defined in (\ref{eq:f(Y|K=00003D0)}). $\calB_{\iGp}(\iGnx,\iGny)\TRIANGLEQ\frac{\iGp^{\iGnx-1}\iGq^{\iGny-1}}{B(\iGnx,\iGny)}$
is the beta distribution, $B(\iGnx,\iGny)\TRIANGLEQ\frac{\Gamma(\iGnx)\Gamma(\iGny)}{\Gamma(\iGnx+\iGny)}$
and $\Gamma(\GGn)=(\GGn-1)!$ are beta  and gamma functions for natural
number\footnote{We can use Stirling's approximation: $\log\Gamma(x)\approx(x-1/2)\log(x)-x+\log(2\pi)/2$,
for large $x$.}, respectively. 

Comparing (\ref{eq:ra_Height}, \ref{eq:ra_Diagonal}) with (\ref{eq:ra_PCA}),
we can see that the noise-to-signal percentage $\rationoise=\frac{\stdZero^{2}}{\varSignal}\in(0,1)$
is a calibrated factor for amplitude's estimation, as follows: 
\begin{equation}
\overline{\Amplitude}\TRIANGLEQ(1-\rationoise)\AmpZero=\left(1-\frac{\stdZero^{2}}{\varSignal}\right)\AmpZero,\label{eq:A_tau}
\end{equation}
in which $\overline{\Amplitude}$ is both conditional mean and MAP
estimate for $\Amplitude$ in (\ref{eq:ra_Diagonal}) and, hence,
$\overline{\Amplitude}$ is closer to the true value $\Amplitude$
than $\AmpZero=\bULA^{+}\yData$ in average, as illustrated in Fig.
\ref{fig:PCA_MUSIC}. When SNR is high (i.e. $\stdamplitude^{2}\gg\stdZero^{2}$
and $\rationoise\rightarrow0$), $\overline{\Amplitude}$ is almost
the same as $\AmpZero$. In contrast, when SNR is low (i.e. $\stdamplitude^{2}\ll\stdZero^{2}$
and $\rationoise\rightarrow1$), $\overline{\Amplitude}$ is closer
to zero and yields lower mean squared error, as illustrated in Fig.
\ref{fig:Overlapping} in the simulation section. 

Also, intuitively, the PCA's likelihood model in (\ref{eq:ra_Diagonal})
is proportional to a product of two Gaussian distributions, one for
observed signal $\NormSquared{\bULA\AmpZero}$ with signal-plus-noise
variance $\varSignal$ in signal subspace and one for the height $\NormSquared{\HZero}$
with noise's variance $\stderror^{2}$ in noise subspace, as illustrated
in Fig. \ref{fig:PCA_MUSIC}. Since both terms $\NormSquared{\bULA\AmpZero}$
and $\NormSquared{\HZero}$ can be computed from observed data $\NormSquared{\yData}$,
the unknown variances $\varSignal$ and $\stderror^{2}$ can also
be estimated from $\NormSquared{\bULA\AmpZero}$ and $\NormSquared{\HZero}$,
respectively, via inverse-gamma distributions in (\ref{eq:uncorr_posterior}),
as shown below.

\subsubsection{\label{subsec:Posterior-variance}Posterior distribution of noise's
variance}

Multiplying the non-informative Jeffreys' priors $f(\stderror^{2})\propto\frac{1}{\stderror^{2}}$
and $f(\varSignal)\propto\frac{1}{\varSignal}$ of positive values
$\stderror^{2}$ and $\varSignal$ with (\ref{eq:uncorr_posterior}),
respectively, we can write down their posterior distributions via
chain rule of probability: 
\begin{align}
f(\yData,\varSignal,\stderror^{2}|\bULA,\nstate) & =f(\yData|\bULA,\nstate,\stderror^{2},\varSignal)f(\stderror^{2})f(\varSignal)\label{eq:ra_joint}\\
=f(\varSignal & |\yData,\bULA,\nstate,\stderror^{2})f(\stderror^{2}|\yData,\bULA,\nstate)f(\yData|\bULA,\nstate).\nonumber 
\end{align}
Since $\{\varSignal,\stdZero\}$ are two random variables (r.v.) of
inverse-gamma distributions and $\varSignal\geq\stdZero$ in (\ref{eq:uncorr_posterior},
\ref{eq:ra_joint}), let us apply the double inverse-gamma distribution
in Appendix \ref{sec:Double-Gamma} to the posteriors in (\ref{eq:ra_joint}),
as follows:
\begin{align}
f(\varSignal|\yData,\bULA,\nstate,\stderror^{2} & )=\caliG_{\varSignal\geq\stdZero^{2}}\left(\iGnx,\frac{\iGsx}{\ndim}\right)\TRIANGLEQ\frac{\caliG_{\varSignal}\left(\iGnx,\frac{\iGsx}{\ndim}\right)}{\frac{\gamma\left(\iGnx,\frac{\iGsx}{\stderror^{2}}\right)}{\Gamma(\iGnx)}},\nonumber \\
f(\stderror^{2}|\yData,\bULA,\nstate)= & \caliGGL_{\stderror^{2}}(\iGnx,\iGny,\iGsx,\iGsy)\TRIANGLEQ\frac{\gamma\left(\iGnx,\frac{\iGsx}{\stderror^{2}}\right)}{\Gamma(\iGnx)}\frac{\caliG_{\stderror^{2}}\left(\iGny,\iGsy\right)}{\Regular_{p}(\iGnx,\iGny)},\label{eq:ra_posterior}
\end{align}
and: 
\begin{align}
f(\yData|\bULA,\nstate) & =\frac{\Pr[\UNIT_{\varSignal\geq\stdZero^{2}}]}{\zeta_{\iGp}}=\frac{\Regular_{p}(\iGnx,\iGny)}{\zeta_{\iGp}}=\frac{\sum_{\iGi=\iGnx}^{\infty}\calNB_{\iGi}(\iGny,\iGp)}{\zeta_{\iGp}}\nonumber \\
 & =f(\yData|\nstate=0)\left(\frac{\Regular_{p}(\iGnx,\iGny)}{\calB_{\iGp}(\iGnx,\iGny)\iGp\iGq}\right)\label{eq:Likeli=00005BV,K=00005Da}\\
 & =f(\yData|\nstate=0)\underset{\ \ \ \ \ \ \ \ \ \ \ \ \mathcal{P}(\bULA,\nstate)=1,\ \text{if}\ \nstate=0.}{\underbrace{\left(\sum_{\iGi=\iGnx}^{\infty}\frac{\Gamma(\iGnx)\Gamma(\iGi+\iGny)}{\iGi!\Gamma(\iGnx+\iGny)}\iGp^{\iGi-\iGnx}\right)}},\nonumber 
\end{align}
or, equivalently:
\begin{align}
f(\yData|\bULA,\nstate) & =\frac{1-\Regular_{\iGq}(\iGny,\iGnx)}{\zeta_{\iGp}}=\frac{\sum_{\iGi=0}^{\iGny-1}\calNB_{\iGi}(\iGnx,\iGq)}{\zeta_{\iGp}}\label{eq:Likeli=00005BV,K=00005Db}\\
 & =f(\yData|\nstate=0)\underset{\ \ \ \ \ \ \ \ \ \ \ \ \mathcal{Q}(\bULA,\nstate)=\mathcal{P}(\bULA,\nstate)}{\underbrace{\left(\sum_{\iGi=0}^{\iGny-1}\frac{\Gamma(\iGny)\Gamma(\iGnx+\iGi)}{\iGi!\Gamma(\iGnx+\iGny)}\frac{1}{\iGq^{\iGny-\iGi}}\right)}}\nonumber 
\end{align}
where $f(\yData|\nstate=0)$ is given in (\ref{eq:f(Y|K=00003D0)}),
$\gamma\left(\iGnx,x\right)$ is the lower incomplete gamma function,
$\Regular_{p}(\iGnx,\iGny)$ is the regularized incomplete beta function
and $\calNB_{\iGi}(\iGnx,\iGq)$ is the negative binomial distribution,
as given in (\ref{eq:Ip}). 

Note that, in order to derive the likelihood $f(\yData|\bULA,\nstate)$
in (\ref{eq:Likeli=00005BV,K=00005Da}), we have marginalized out
all possible values of signal's and noise's variance in (\ref{eq:ra_joint}).
Since signal-plus-noise variance $\varSignal$ must be higher than
noise's variance $\stderror^{2}$, we recognize that the PCA's likelihood
$f(\yData|\bULA,\nstate)$ in (\ref{eq:Likeli=00005BV,K=00005Da})
is actually proportional to probability $\Pr[\UNIT_{\varSignal\geq\stdZero^{2}}]$
of the event $\varSignal\geq\stderror^{2}$. Intuitively, the negative
binomial form in (\ref{eq:Likeli=00005BV,K=00005Da}, \ref{eq:Likeli=00005BV,K=00005Db})
implies that the likelihood probability $f(\yData|\bULA,\nstate)$
would take into account all binomial combination of all possible values
of signal's dimension $\iGnx=\nstate\nsource$ and noise's dimension
$\iGny=(\ndim-\nstate)\nsource$ over $\nstate$.

If $\nstate=0$, we have $\NormSquared{\bULA\AmpZero}=0$ and, hence,
$\iGnx=\iGp=0$ by convention in (\ref{eq:Likeli_Gauss}). We then
have $\mathcal{P}(\bULA,\nstate=0)=1$ in (\ref{eq:Likeli=00005BV,K=00005Da}),
owing to convention $\iGnx!=0!=\iGp^{0}=0^{0}=1$. Note that, although
the infinite sums $\mathcal{P}(\bULA,\nstate)$ in (\ref{eq:Likeli=00005BV,K=00005Da})
is guaranteed to converge with $\iGp\in(0,1)$, the form of finite
sums $\mathcal{Q}(\bULA,\nstate)$ in (\ref{eq:Likeli=00005BV,K=00005Db})
is more suitable for efficient computation in practice. 

From (\ref{eq:inverse-moment}) in Appendix \ref{sec:Double-Gamma},
the conditional mean estimates of $\varSignal$ and $\stderror^{2}$
can be computed as follows: 
\begin{align}
\overline{\varSignal} & \TRIANGLEQ\EXPECTATION_{f(\varSignal|\yData,\bULA,\nstate)}(\varSignal)=\frac{\frac{\iGsx}{\ndim}}{\iGnx-1}\frac{\Regular_{p}(\iGnx-1,\iGny)}{\Regular_{p}(\iGnx,\iGny)}\approx\frac{\frac{1}{\ndim}\NormSquared{\bULA\AmpZero}}{\nstate\nsource},\nonumber \\
\overline{\stderror^{2}} & \TRIANGLEQ\EXPECTATION_{f(\stderror^{2}|\yData,\bULA,\nstate)}(\stderror^{2})=\frac{\GGsy}{\iGny-1}\frac{\Regular_{p}(\iGnx,\iGny-1)}{\Regular_{p}(\iGnx,\iGny)}\approx\frac{\NormSquared{\HZero}}{(\ndim-\nstate)\nsource},\label{eq:noise_MSE}
\end{align}
in which the approximations are accurate if $\iGnx,\iGny\gg1$. The
plug-in estimate for the noise-to-signal percentage $\rationoise\in(0,1)$
is then\footnote{Note that, substituting (\ref{eq:ratio_noise_MSE}) into (\ref{eq:ra_Height})
yields: $\NormSquared{\Height_{\overline{\rationoise}}}=\frac{\ndim}{\ndim-\nstate}\NormSquared{\HZero}<\NormSquared{\yData}=\NormSquared{\bULA\AmpZero}+\NormSquared{\HZero}\Leftrightarrow\frac{\nstate}{\ndim-\nstate}\NormSquared{\HZero}<\NormSquared{\bULA\AmpZero}\Leftrightarrow\overline{\rationoise}<1.$}: 
\begin{equation}
\overline{\rationoise}\TRIANGLEQ\frac{\overline{\stdZero^{2}}}{\overline{\varSignal}}\approx\frac{\frac{1}{\ndim-\nstate}\NormSquared{\HZero}}{\frac{1}{\nstate}\NormSquared{\bULA\AmpZero}},\ \text{with}\ \overline{\stdZero^{2}}\TRIANGLEQ\frac{\overline{\std^{2}}}{\ndim}.\label{eq:ratio_noise_MSE}
\end{equation}

\subsubsection{MAP estimate of principal vectors}

Since our principal vectors $\bULA$ belong to the Stiefel manifold
$\calS_{\ndim}^{\nstate}$ with radius $\StiefelRadius=\sqrt{\ndim}$,
as defined in (\ref{eq:Vmax}), its non-informative prior can be defined
uniformly over the volume $\text{vol}(\calS_{\ndim}^{\nstate})$,
i.e. $f(\bULA)=\frac{1}{\text{vol}(\calS_{\ndim}^{\nstate})}$, $\forall\bULA\in\calS_{\ndim}^{\nstate}$,
where $\text{vol}(\calS_{\ndim}^{\nstate})=\prod_{\istate=\ndim-\nstate+1}^{\ndim}\frac{2(\pi\StiefelRadius^{2})^{\istate}}{\Gamma(\istate)\StiefelRadius}$,
as shown in \cite{Stiefel:complex:05}. 

Nonetheless, it is not feasible to derive a closed-form for posterior
distribution $f(\bULA|\yData,\nstate)\propto f(\yData|\bULA,\nstate)f(\bULA)$
in (\ref{eq:Likeli=00005BV,K=00005Da}, \ref{eq:Likeli=00005BV,K=00005Db}).
For this reason, let us compute the MAP estimate $\widehat{\bULA}\TRIANGLEQ\argmax_{\bULA\in\calS_{\ndim}^{\nstate}}f(\bULA|\yData,\nstate)$,
as follows: 
\begin{align}
\widehat{\bULA} & =\argmax_{\bULA\in\calS_{\ndim}^{\nstate}}f(\yData|\bULA,\nstate)f(\bULA)\label{eq:MAP_V_Stiefel}\\
 & =\argmax_{\bULA\in\calS_{\ndim}^{\nstate}}(\iGp)=\argmin_{\bULA\in\calS_{\ndim}^{\nstate}}(\iGq)\nonumber \\
 & =\argmax_{\bULA\in\calS_{\ndim}^{\nstate}}\NormSquared{\bULA\AmpZero}=\argmin_{\bULA\in\calS_{\ndim}^{\nstate}}\NormSquared{\HZero},\nonumber 
\end{align}
which can be computed via the PCA method (\ref{eq:Vmax}). Hence,
given uniform prior $f(\bULA)$, PCA actually returns the same MAP
estimate $\widehat{\bULA}$ for both cases of known and unknown noise's
variance in (\ref{eq:postDOAsigma}, \ref{eq:ra_Diagonal}) and (\ref{eq:MAP_V_Stiefel}),
respectively.

\subsubsection{MAP estimate of the number of components}

Multiplying the uniform prior $f(\nstate)=\frac{1}{1+\threshK}$,
$\forall\nstate\in\{0,\ldots,\threshK\}$ with the likelihood $f(\yData,\widehat{\bULA}|\nstate)=f(\yData|\widehat{\bULA},\nstate)f(\widehat{\bULA})$
in (\ref{eq:MAP_V_Stiefel}), we can find the MAP estimate $\widehat{\nstate}\TRIANGLEQ\argmax_{\nstate}f(\yData,\widehat{\bULA},\nstate)$
for $\nstate$, as follows: 
\begin{align}
\widehat{\nstate} & =\argmax_{\nstate\geq0}\log\left(f(\yData,\widehat{\bULA}|\nstate)f(\nstate)\right)\label{eq:K_MAP_Stiefel}\\
 & =\argmax_{\nstate\geq0}\left(\log\mathcal{Q}(\widehat{\bULA},\nstate)+\log\frac{1}{\text{vol}(\calS_{\ndim}^{\nstate})}\right).\nonumber 
\end{align}
Note that, although we can feasibly compute the likelihood $f(\yData,\widehat{\bULA}|\nstate)$
in (\ref{eq:K_MAP_Stiefel}) via standard beta and binomial form $\frac{\Regular_{p}(\iGnx,\iGny)}{\calB_{\iGp}(\iGnx,\iGny)\iGp\iGq}$
in (\ref{eq:Likeli=00005BV,K=00005Da}), the computation of these
standard functions is often overflown when $\iGnx$ and $\iGny$ are
high in practice. For this reason, we prefer the direct logarithm
form in (\ref{eq:K_MAP_Stiefel}) via finite sums $\mathcal{Q}(\bULA,\nstate)$
in (\ref{eq:Likeli=00005BV,K=00005Db}).

\subsection{MAP estimates for MUSIC algorithm}

As shown in (\ref{eq:MUSIC}), the MUSIC algorithm is similar to the
PCA method, except that the uniform prior $f(\ULA)=f(\bangleDOA)=\frac{1}{(2\pi)^{\nstate}}$,
$\forall\bangleDOA\in[-\pi,\pi)^{\nstate}$, for steering vectors
is defined over space of DOAs in this case. Hence, the pseudo-spectrum
(\ref{eq:MUSIC}) in the MUSIC algorithm also returns the MAP estimate
$\bULA_{\bangleDOA=\widehat{\bangleDOA}}$ of $\ULA$ via (\ref{eq:MAP_V_Stiefel}).
The number of sources is then estimated via (\ref{eq:K_MAP_Stiefel}),
as follows: 
\begin{align}
\widehat{\nstate} & =\argmax_{\nstate\geq0}\log\left(f(\yData,\bULA_{\bangleDOA=\widehat{\bangleDOA}}|\nstate)f(\nstate)\right)\label{eq:K_MAP_MUSIC}\\
 & =\argmax_{\nstate\geq0}\left(\log\mathcal{Q}(\bULA_{\bangleDOA=\widehat{\bangleDOA}},\nstate)+\log\frac{1}{(2\pi)^{\nstate}}\right).\nonumber 
\end{align}
Note that, the DTFT spectrum method (\ref{eq:DOA_DTFT}) is also the
MAP estimate of $\ULA$ via (\ref{eq:MAP_V_Stiefel}) under the condition
of strictly uncorrelated DOAs (\ref{eq:uncorrV}). Hence, we can also
compute $\widehat{\bULA}$ in (\ref{eq:K_MAP_MUSIC}) via the DTFT
spectrum method (\ref{eq:DOA_DTFT}), although this method is only
accurate if all DOAs lie at zero points of DTFT spectrum in Fig. \ref{fig:DOA}. 

\section{Simulations\label{sec:Simulations}}

In this section, let us compare the performance of MAP estimate with
that of Akaike information criterion (AIC) \cite{DoA:AIC:VanTrees:02}
for the DTFT and MUSIC algorithms. For the sake of comparison, the
case of known ground-truth $\nstate$ is also given. 

\subsection{Uncorrelated multi-tone sources}

For uncorrelated condition (\ref{eq:uncorrApprox}), we need to set
$\nstate\leq\threshK\ll\ndim$ and $\Delta\angleDOA\gg\frac{2\pi}{\ndim}$,
as illustrated in Fig. \ref{fig:DOA}. Let us consider this case first,
with default setting below. The simulation of this case is given in
Fig. \ref{fig:Overlapping}.

\subsubsection{Default setting\label{subsec:Default}}

Throughout simulations, our default parameters are $\ndim=100$ sensors,
$\nstate=5$ sources and $\nsource=\ntime=2^{12}$ FFT-bins. The preset
number of sources in DTFT and MUSIC algorithms is $\threshK=10$.
Also, for high resolution, we discretize the range $[0^{0},180^{0})$
of DOA angles into very small steps of $0.1^{0}$ in the DTFT and
MUSIC algorithms. The number of Monte Carlo runs for all cases is
$10^{3}$. The signal-to-noise ratio (SNR) is defined as follows:
\begin{equation}
\SNR\TRIANGLEQ10\log_{10}\frac{\max_{\istate=1,\ldots,\nstate}\sigma_{\istate}^{2}}{\stdZero^{2}}\ \text{(dB)},\ \text{with}\ \sigma_{\istate}^{2}=\frac{\NormHS{\bamplitude_{\istate}}^{2}}{\nsource},\label{eq:SNR}
\end{equation}
which corresponds to the ratio between maximum averaged source's power
per tone, as defined in (\ref{eq:MATRIX_MODEL}, \ref{eq:ra_PCA}),
and projected noise's variance $\stdZero^{2}=\frac{\std^{2}}{\ndim}$
on signal subspace, as illustrated in Fig. \ref{fig:PCA_MUSIC}. 

The true amplitudes are  $\amplitude_{\istate,\isource}=\UNIT_{\isource\in[\isource_{\istate},\isource_{\istate}+\text{BW}]}$,
in which $\isource_{\istate}\TRIANGLEQ1+(\istate-1)\left\lceil (1-\vartheta)\text{BW}\right\rceil $
and the bandwidth of each source is $\text{BW}=\left\lfloor \frac{\nsource}{\nstate}\right\rfloor $,
with $\left\lceil \cdot\right\rceil $ and $\left\lfloor \cdot\right\rfloor $
denoting the upper- and lower-rounded integer operator, respectively,
$\forall\seti{\istate}{\nstate}$. The overlapping ratio is $\vartheta\in[0,1]$,
as illustrated in Fig. \ref{fig:Overlapping_PCA}. Note that, the
value $\vartheta=99.9\%$ would yield $\left\lceil (1-\vartheta)\text{BW}\right\rceil =1$
FFT-bin, which is the smallest number of non-overlapping tones between
two consecutive sources in this setting.

The true DOA angles $\Angle_{\istate}=10^{0}+(\istate-1)\Delta\Angle$
are separated equally over the range $[10^{0},180^{0}]$, where DOA's
difference is $\Delta\Angle\TRIANGLEQ\left\lfloor \frac{170^{0}}{\nstate}\right\rfloor $,
$\forall\seti{\istate}{\nstate}$. Since DOAs are continuous values
and their accuracy also depends on the accuracy of the estimated number
of sources, there is no unique way to evaluate the DOA's estimate
error in the DOA's literature. For this reason, we use a method similar
to purity (i.e. successful rate of correct classification) in clustering
literature \cite{purity:cite:06}. Let us arrange the true DOA angles
$\bAngle\TRIANGLEQ\setv{\Angle}{\nstate}^{\transpose}$ and their
estimates $\widehat{\bAngle}\TRIANGLEQ\setv{\widehat{\Angle}}{\widehat{\nstate}}^{\transpose}$
in non-decreasing order $\Angle_{\istate}\leq\Angle_{\istate+1}$
and $\widehat{\Angle}_{\istate}\leq\widehat{\Angle}_{\istate+1}$,
respectively, in which $\widehat{\nstate}$ is our estimate of the
number $\nstate$ of sources. The estimate's error-rate $\text{ERR}(\widehat{\bAngle},\bAngle)\in[0,1]$
is then defined as follows: 
\begin{equation}
\text{ERR}(\widehat{\bAngle},\bAngle)\TRIANGLEQ\UNIT_{\widehat{\nstate}=0}+\UNIT_{\widehat{\nstate}>0}\frac{1}{\widehat{\nstate}}\frac{\sum_{j=1}^{\widehat{\nstate}}\min_{\seti{\istate}{\nstate}}|\widehat{\Angle_{j}}-\Angle_{\istate}|}{180^{0}},\label{eq:ERR}
\end{equation}
and, hence, the error-rate $\text{ERR}$ is $100\%$ if $\widehat{\nstate}=0$. 

For estimate's error of amplitudes, we use a method similar to Kolmogorov--Smirnov
distance for cumulative density function (c.d.f) \cite{Kolmogorov_distance,VH:PhDthesis:14}.
Let $\widehat{\amplitude}_{\istate,\isource}$ denote the $\{\istate,\isource\}$-element
of matrix $\widehat{\Amplitude}$, which is our estimate of the true
amplitude matrix $\Amplitude$. Since $\amplitude_{\istate,\isource}$
and $\widehat{\amplitude}_{\istate,\isource}$ are associated with
true DOAs $\Angle_{\istate}$ and estimated DOAs $\widehat{\Angle}_{\istate}$,
respectively, let us define the cumulative power spectrums as follows:
$F_{\isource}(\Angle)\TRIANGLEQ\int_{0}^{\Angle}\sum_{\istate=1}^{\nstate}\NORMTWO{\amplitude_{\istate,\isource}}\delta(\Angle-\Angle_{\istate})d\Angle$,
and $\widehat{F}_{\isource}(\Angle)\TRIANGLEQ\UNIT_{\widehat{\nstate}>0}\int_{0}^{\Angle}\sum_{\istate=1}^{\widehat{\nstate}}\NORMTWO{\widehat{\amplitude}_{\istate,\isource}}\delta(\Angle-\widehat{\Angle}_{\istate})d\Angle$,
in which we set $\widehat{\amplitude}_{\istate,\isource}=0$, $\forall\istate,\isource$,
if $\widehat{\nstate}=0$. The empirical root mean squared error $\text{RMSE}(\widehat{\Amplitude},\Amplitude)$
is defined as follows: 
\begin{equation}
\text{RMSE}(\widehat{\Amplitude},\Amplitude)=\sqrt{\frac{1}{\nsource}\sum_{\isource=1}^{\nsource}\int_{0}^{180}(F_{\isource}(\Angle)-\widehat{F}_{\isource}(\Angle))^{2}d\Angle}.\label{eq:MSE_A}
\end{equation}
We then consider two choices of $\widehat{\Amplitude}$: the maximum
likelihood (ML) estimate $\AmpZero$ in (\ref{eq:condA}) and conditional
MAP estimate $\overline{\Amplitude}$ in (\ref{eq:A_tau}).

\subsubsection{Illustration of overlapping sources via PCA model}

Since PCA and MUSIC have the same form of factor analysis in (\ref{eq:PCA}),
the illustration of their similarity for overlapping multi-tone sources
is given in Fig. \ref{fig:Overlapping_PCA}. 

When $\vartheta\rightarrow0\%$, there are very few overlapping tones
between sources and, hence, there is little correlation between PCA's
components. Note that, the observed data are strongly correlated along
each principal vector in this case and, hence, these principal vectors
are feasible to detect. 

When $\vartheta\rightarrow100\%$, all sources are overlapped with
each other and, hence, there is full correlation between PCA's components.
Since the observed data are now uncorrelated for any choice of the
principal vectors, the principal vectors become ambiguous and difficult
to detect in this case.

\begin{figure*}
\begin{centering}
\includegraphics[width=0.33\textwidth]{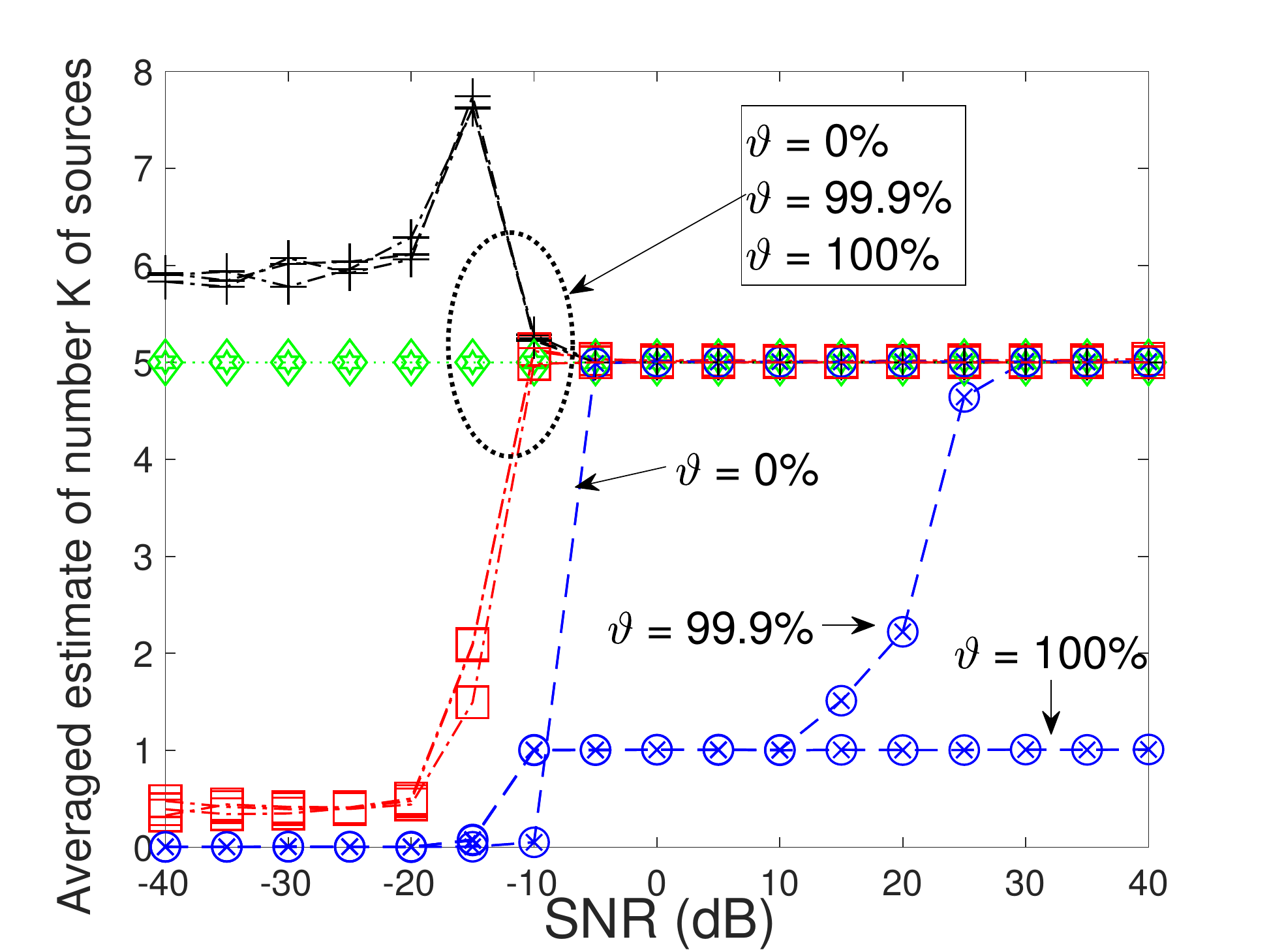}\includegraphics[width=0.33\textwidth]{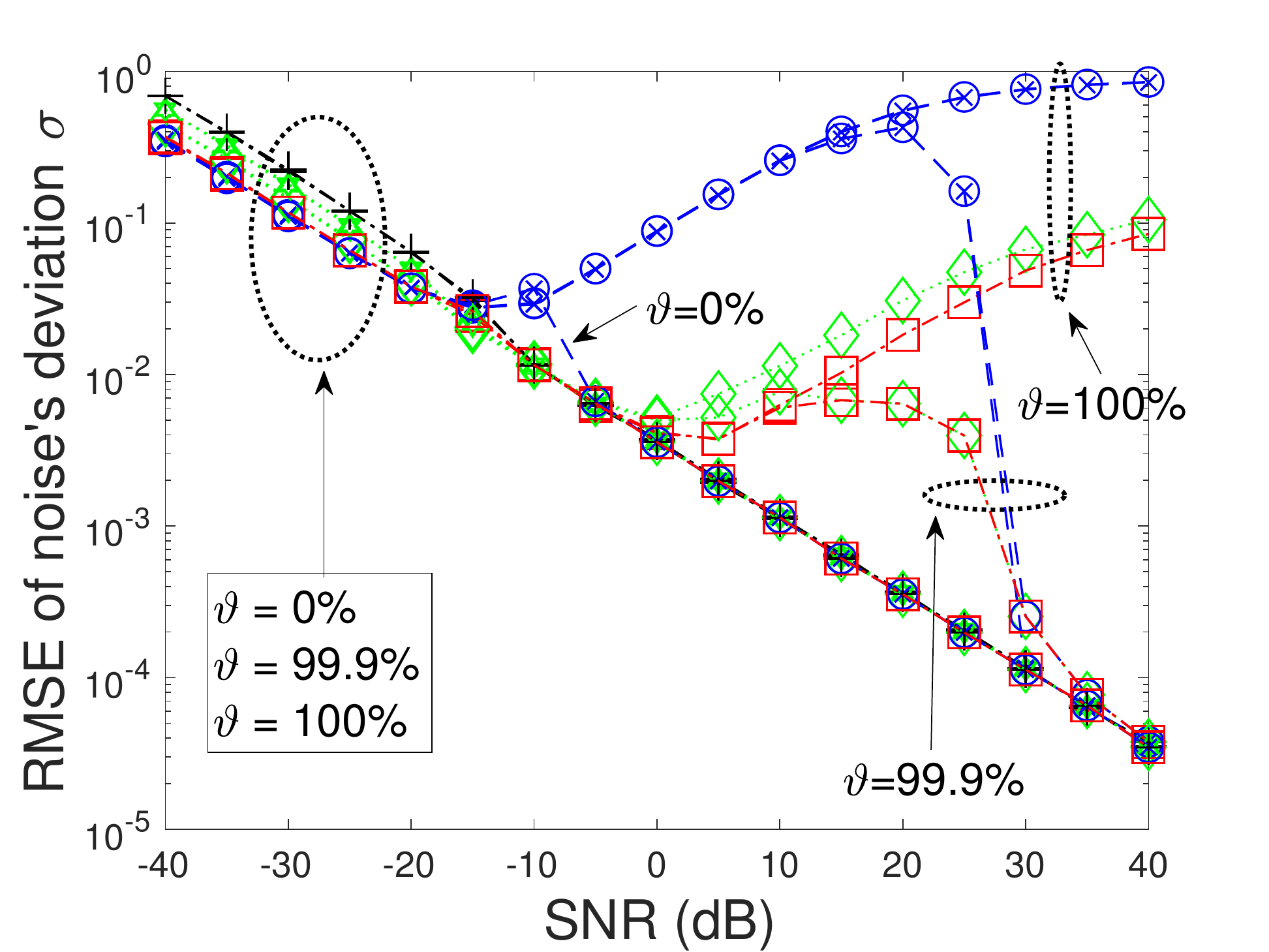}\includegraphics[width=0.33\textwidth]{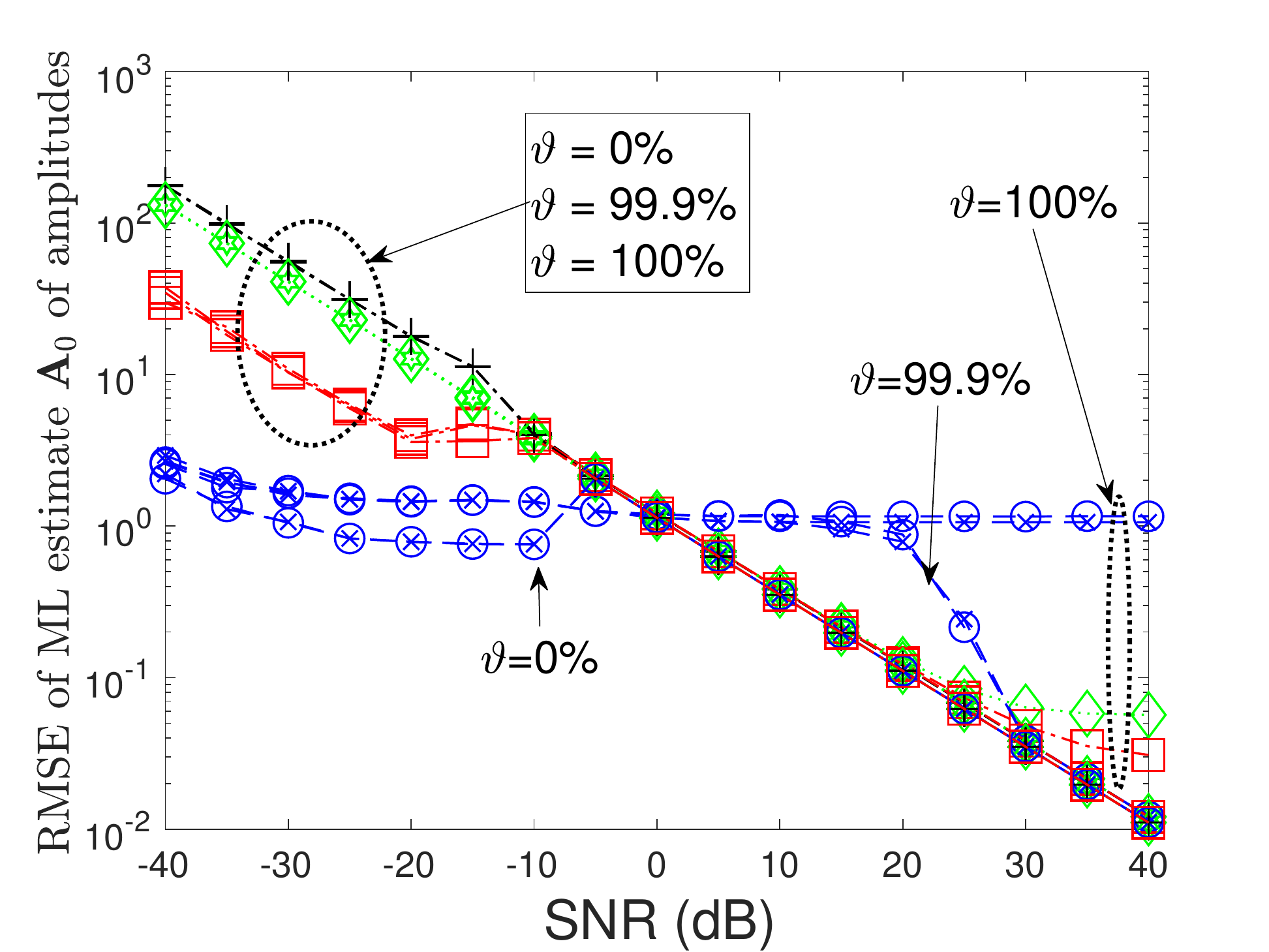}
\par\end{centering}
\begin{centering}
\begin{tabular}{>{\centering}p{0.33\textwidth}>{\centering}p{0.25\textwidth}>{\centering}p{0.33\textwidth}}
{\footnotesize{}(a)} & {\footnotesize{}(c)} & {\footnotesize{}(e)}\tabularnewline
\end{tabular}
\par\end{centering}
\begin{centering}
\includegraphics[width=0.33\textwidth]{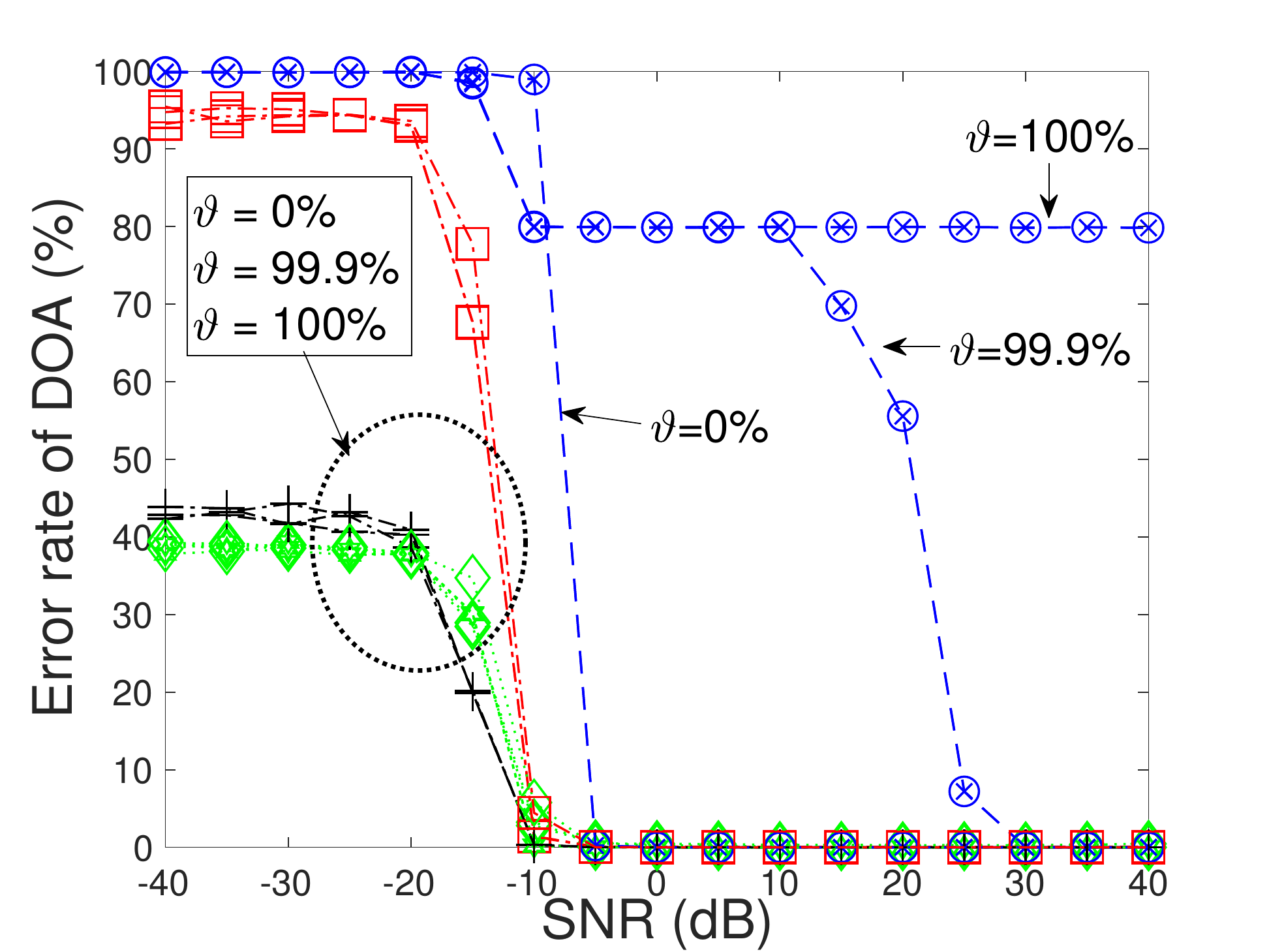}\includegraphics[width=0.33\textwidth]{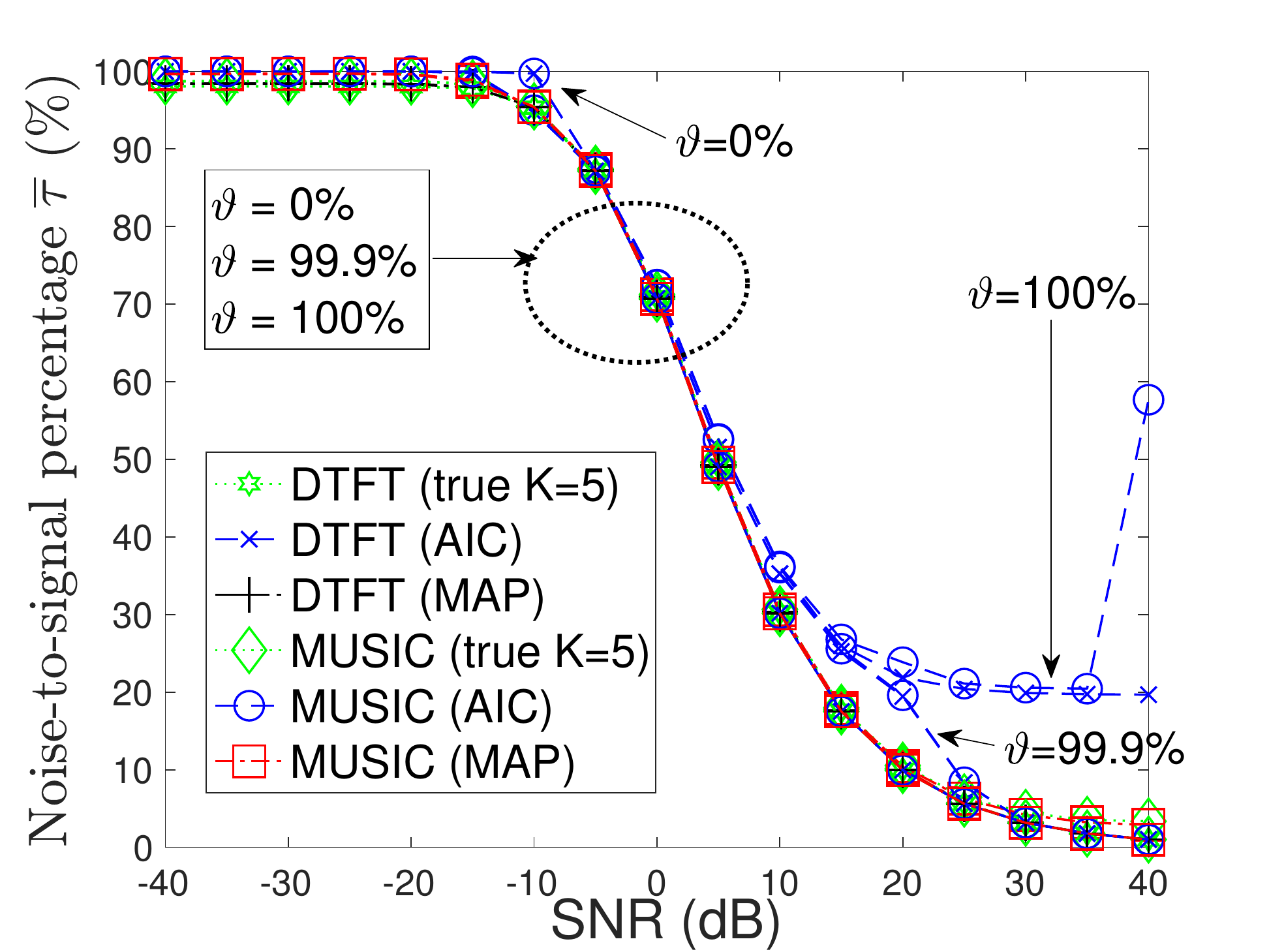}\includegraphics[width=0.33\textwidth]{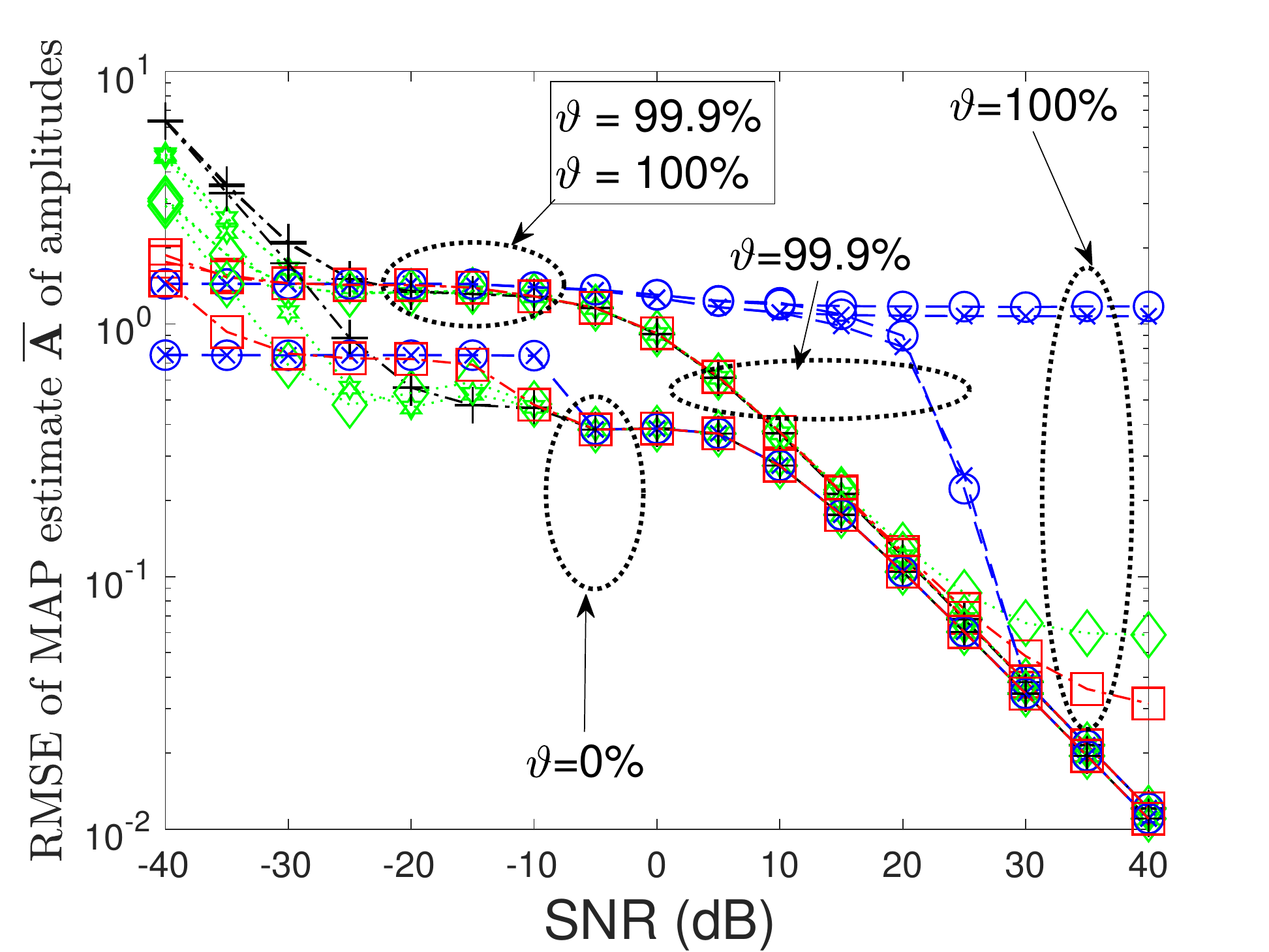}
\par\end{centering}
\begin{centering}
\begin{tabular}{>{\centering}p{0.33\textwidth}>{\centering}p{0.25\textwidth}>{\centering}p{0.33\textwidth}}
{\footnotesize{}(b)} & {\footnotesize{}(d)} & {\footnotesize{}(f)}\tabularnewline
\end{tabular}
\par\end{centering}
\caption{\label{fig:Overlapping}Three cases of uncorrelated DOAs: non-overlapping
($\vartheta=0\%$), almost overlapping ($\vartheta=99.9\%$) and completely
overlapping ($\vartheta=100\%$) for multi-tone sources, with default
setting in section \ref{subsec:Default}. The legend is the same for
all figures. Some curves are almost identical with different values
$\vartheta$ and, hence, indicated by dotted ellipses.}
\end{figure*}

\begin{figure}[h]
\begin{centering}
\includegraphics[width=0.65\columnwidth]{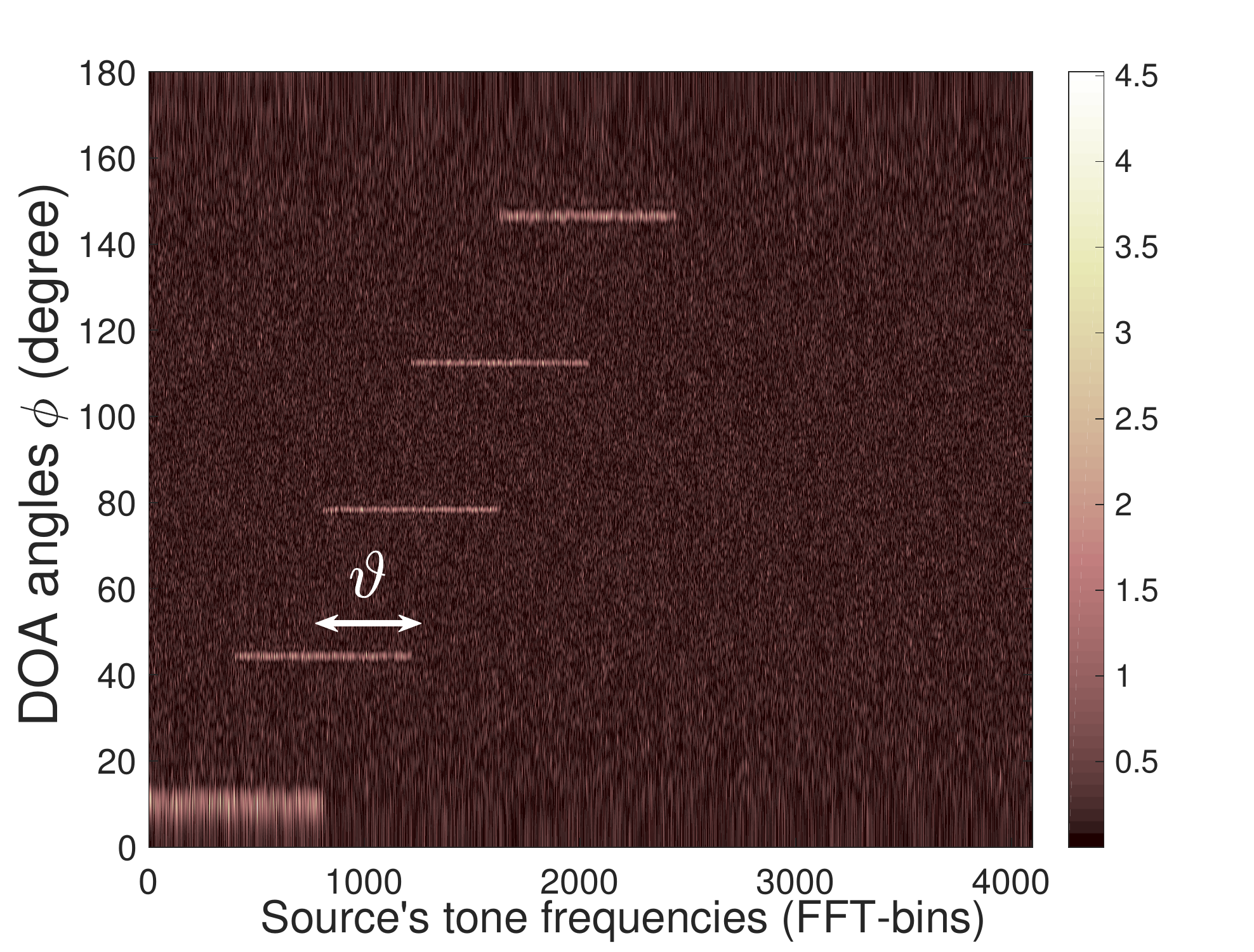}\includegraphics[width=0.35\columnwidth]{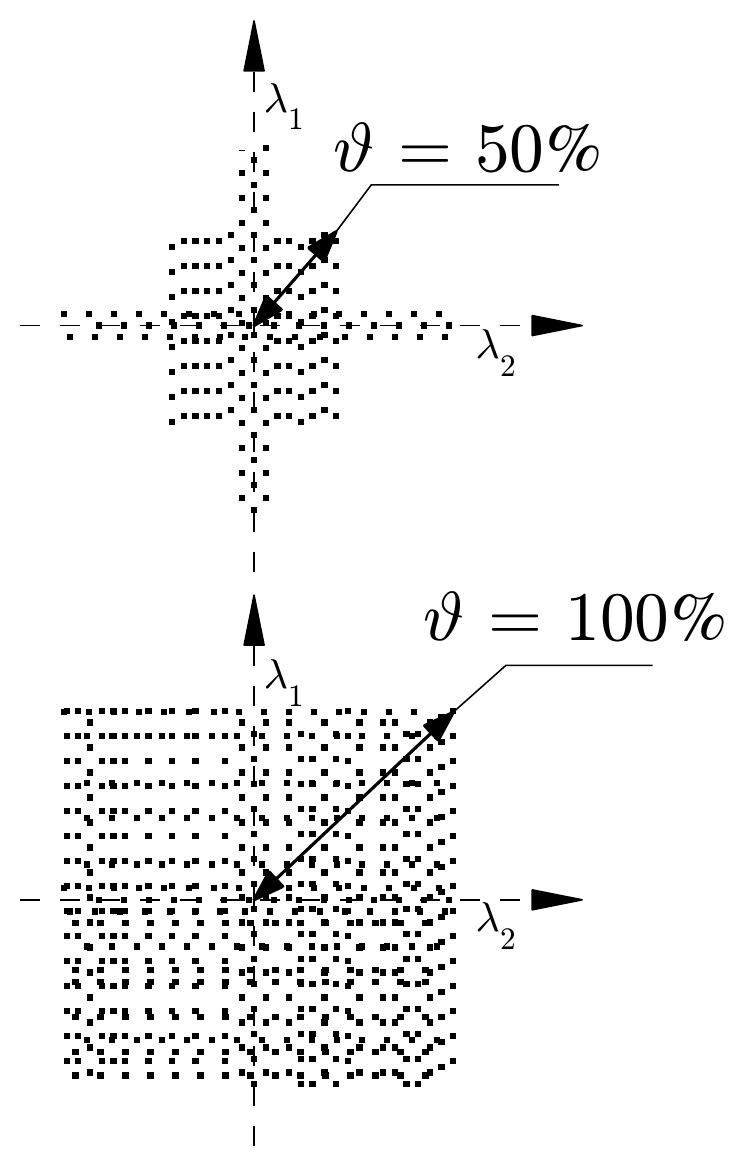}
\par\end{centering}
\caption{\label{fig:Overlapping_PCA}Illustration of similarity between amplitudes
$\protect\Amplitude$ of overlapping multi-tone sources and projected
components $\protect\Amplitude$ on principal vectors in PCA. (Left)
DTFT power spectrum over FFT-bins, with SNR = 0 dB and $\vartheta=50\%$
in Fig. \ref{fig:Overlapping}. (Right) The dots represent the observed
data in PCA, with two cases of overlapping ratio $\vartheta$. }
\end{figure}

\begin{figure}
\begin{centering}
\includegraphics[width=0.25\textwidth]{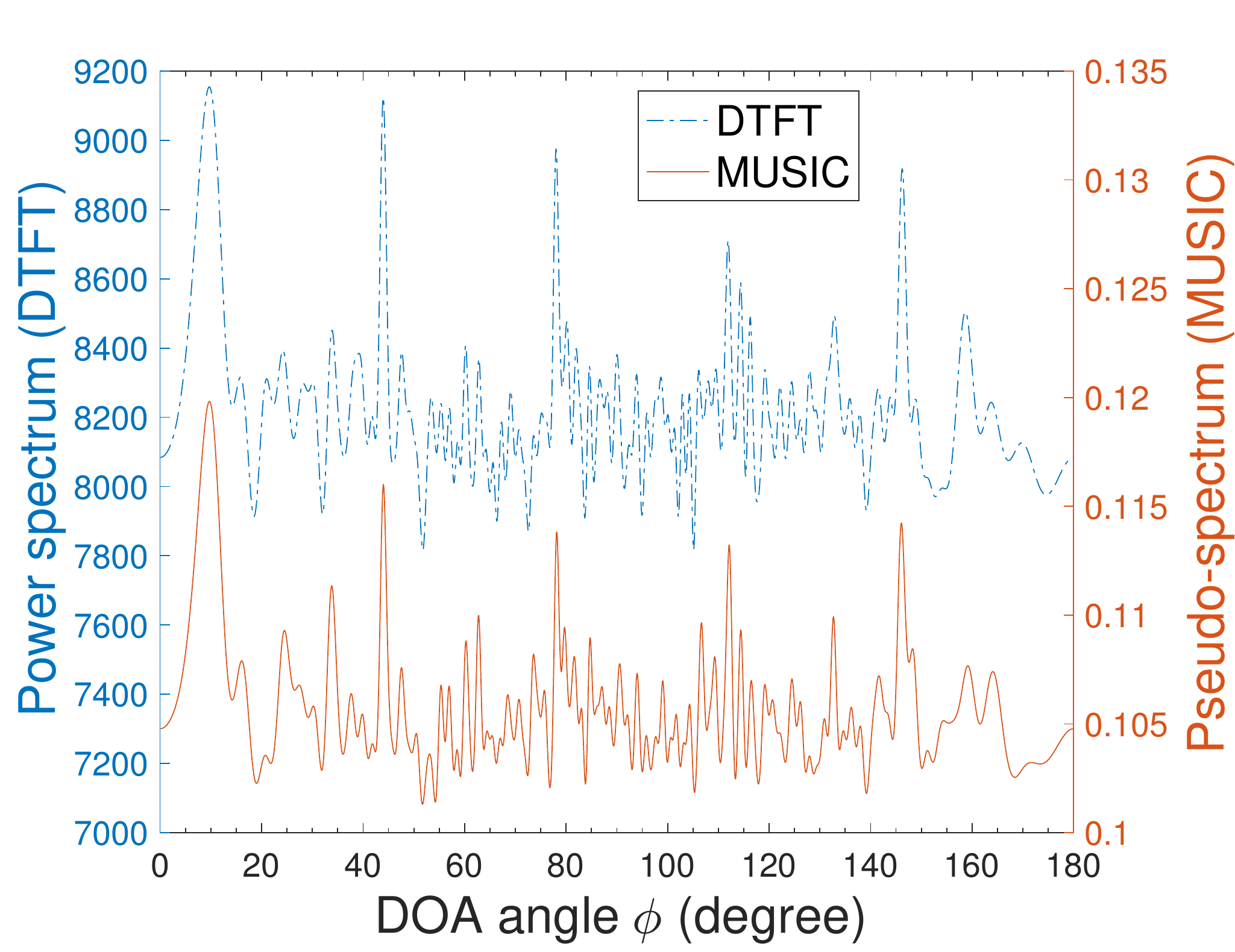}\includegraphics[width=0.25\textwidth]{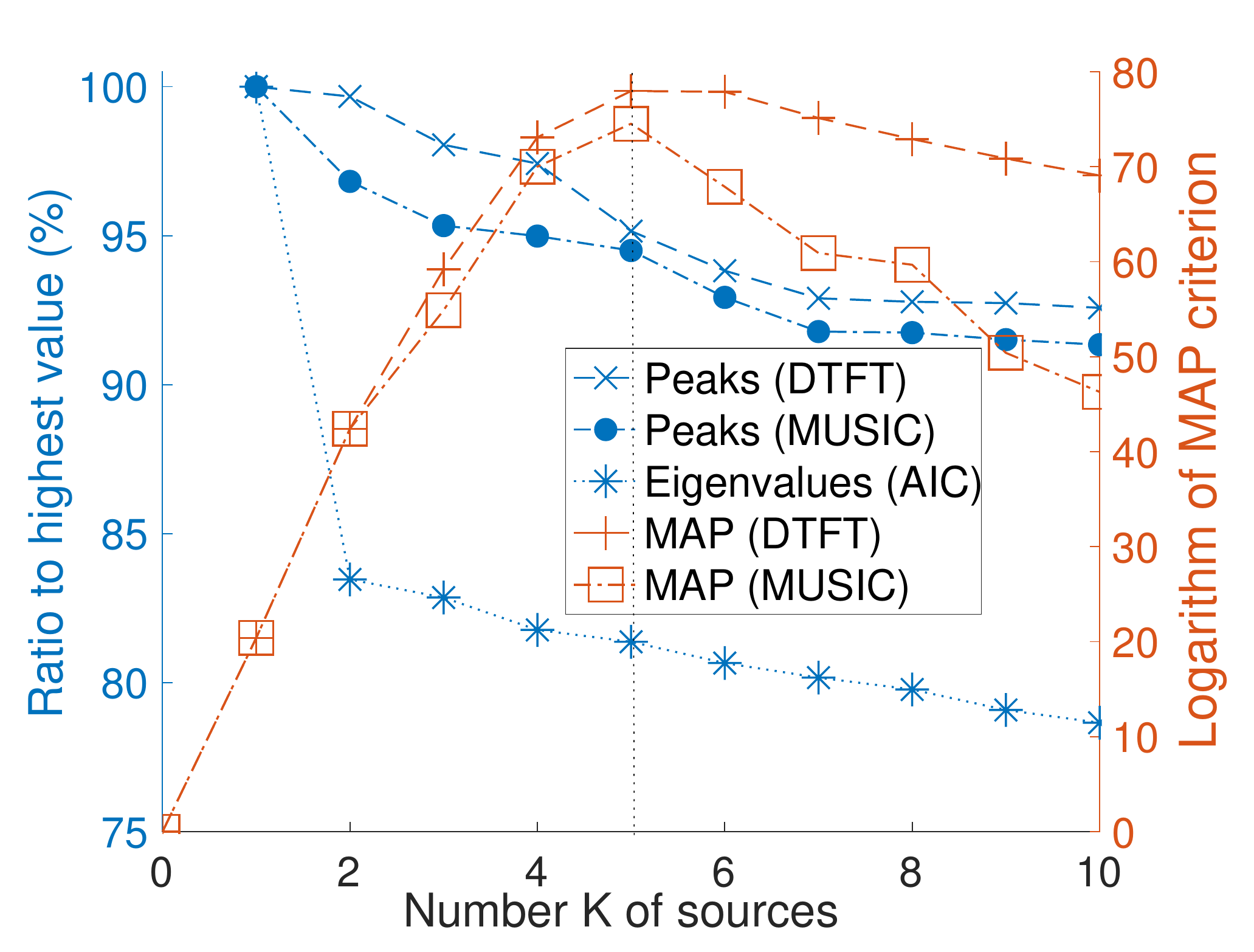}
\par\end{centering}
\caption{\label{fig:OverlapSpectrum}Eigenvalues (AIC), peaks of power spectrum
(DTFT) and peaks of pseudo-spectrum (MUSIC) in descending order, with
one Monte Carlo run, $\protect\SNR=-10$ (dB), $\vartheta=99.9\%$
and the same setting in Fig. \ref{fig:Overlapping}.}
\end{figure}

\begin{figure}
\begin{centering}
\includegraphics[width=0.5\columnwidth]{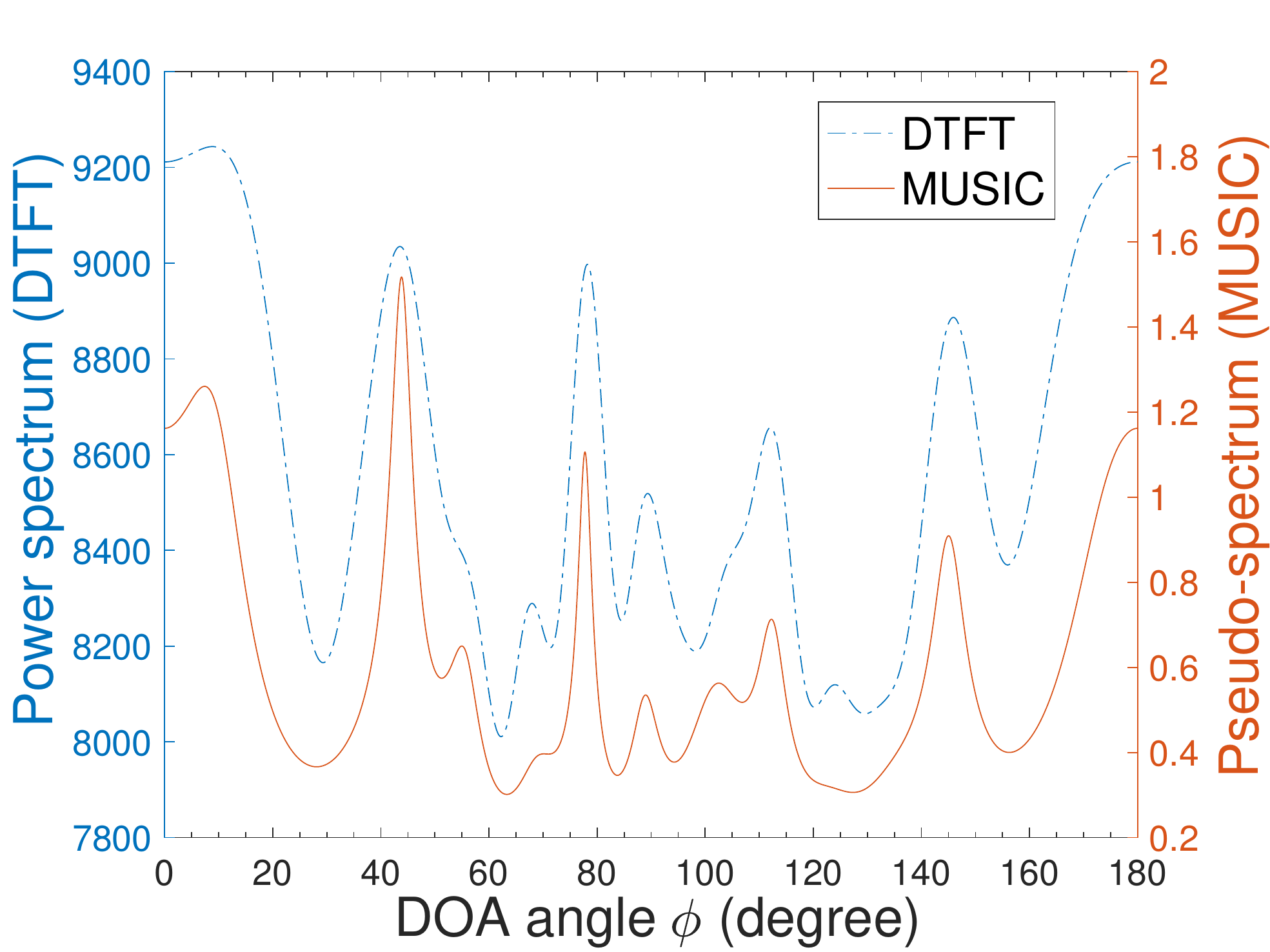}\includegraphics[width=0.5\columnwidth]{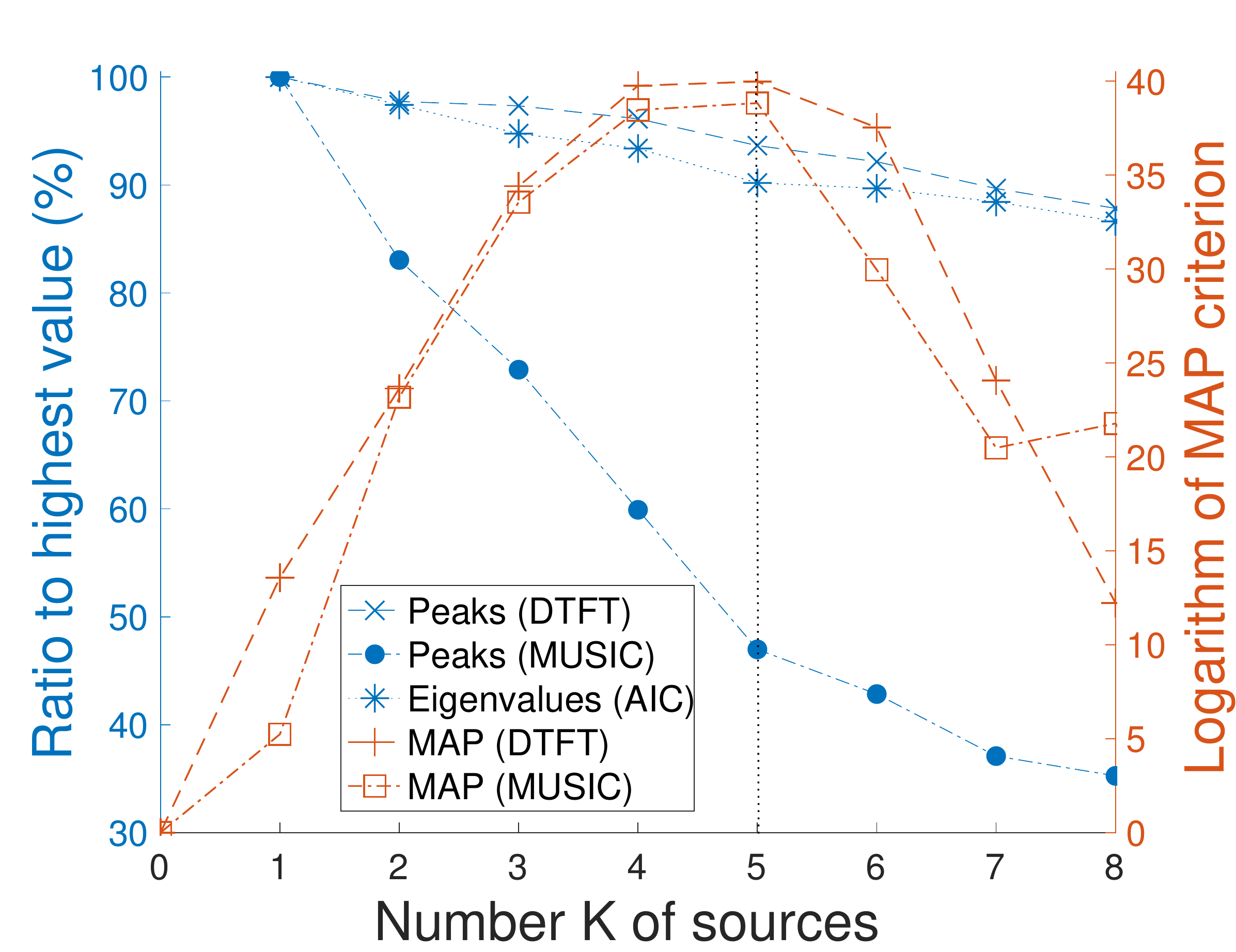}
\par\end{centering}
\begin{centering}
\includegraphics[width=0.5\columnwidth]{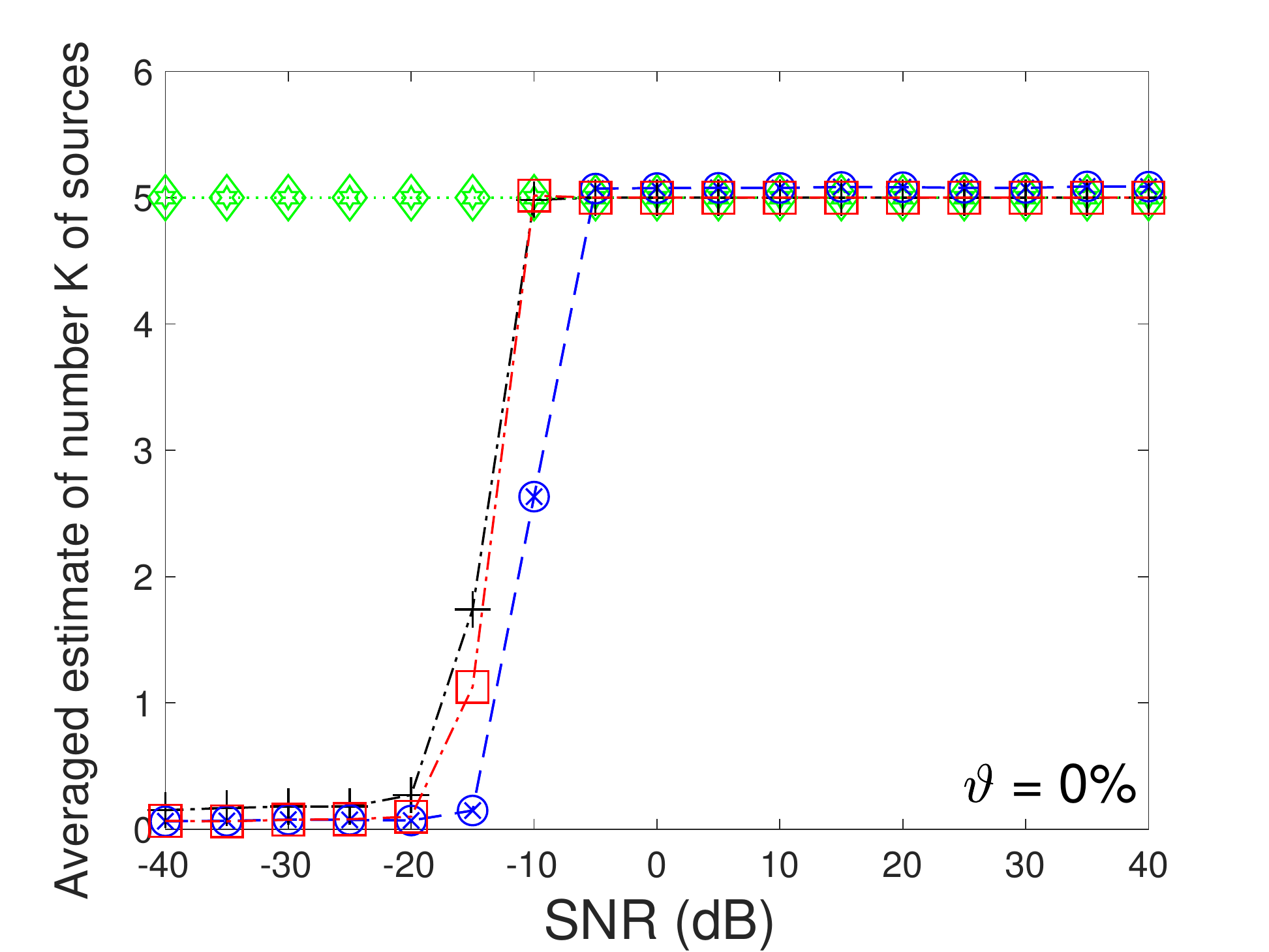}\includegraphics[width=0.5\columnwidth]{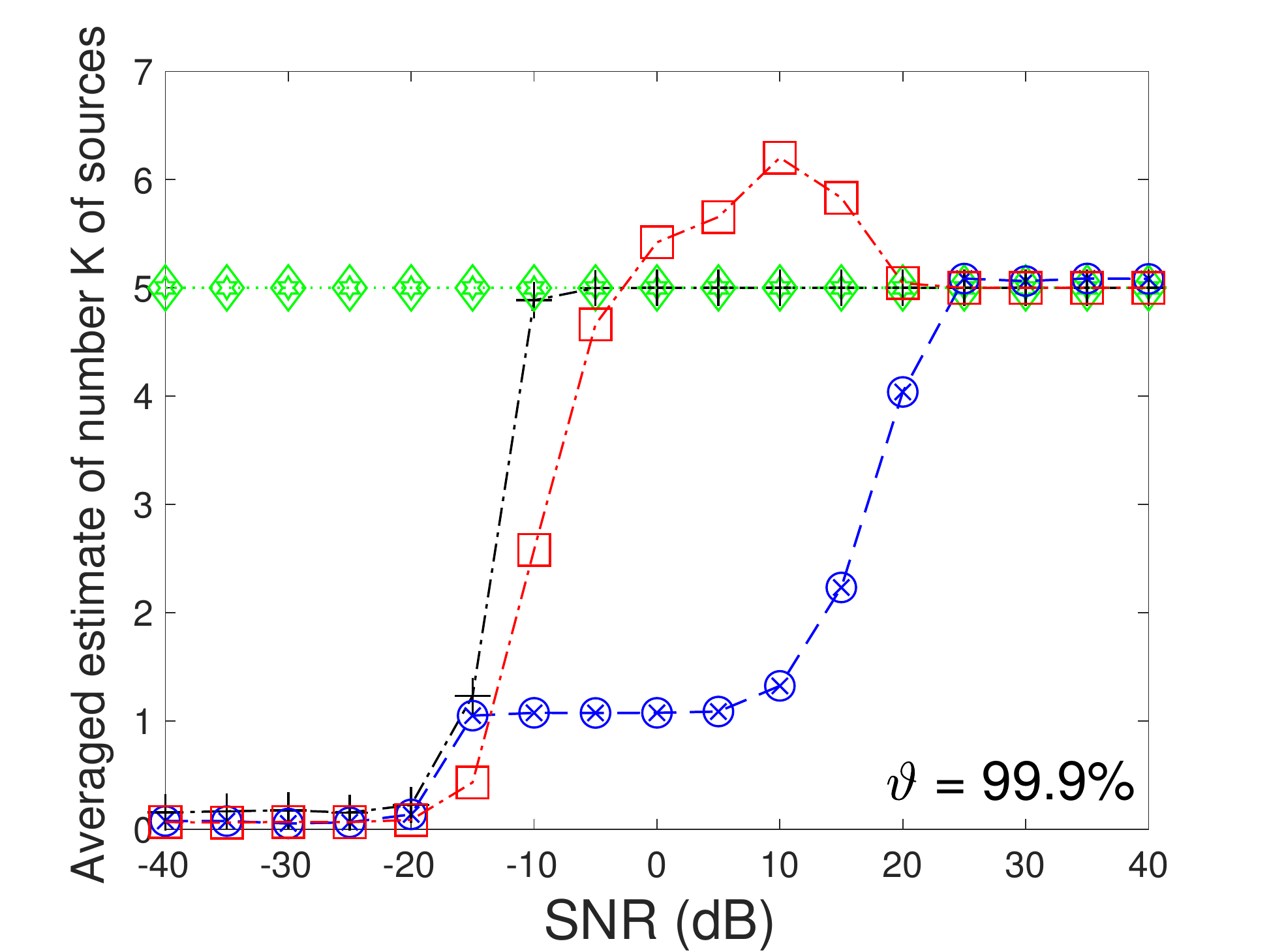}
\par\end{centering}
\begin{centering}
\includegraphics[width=0.5\columnwidth]{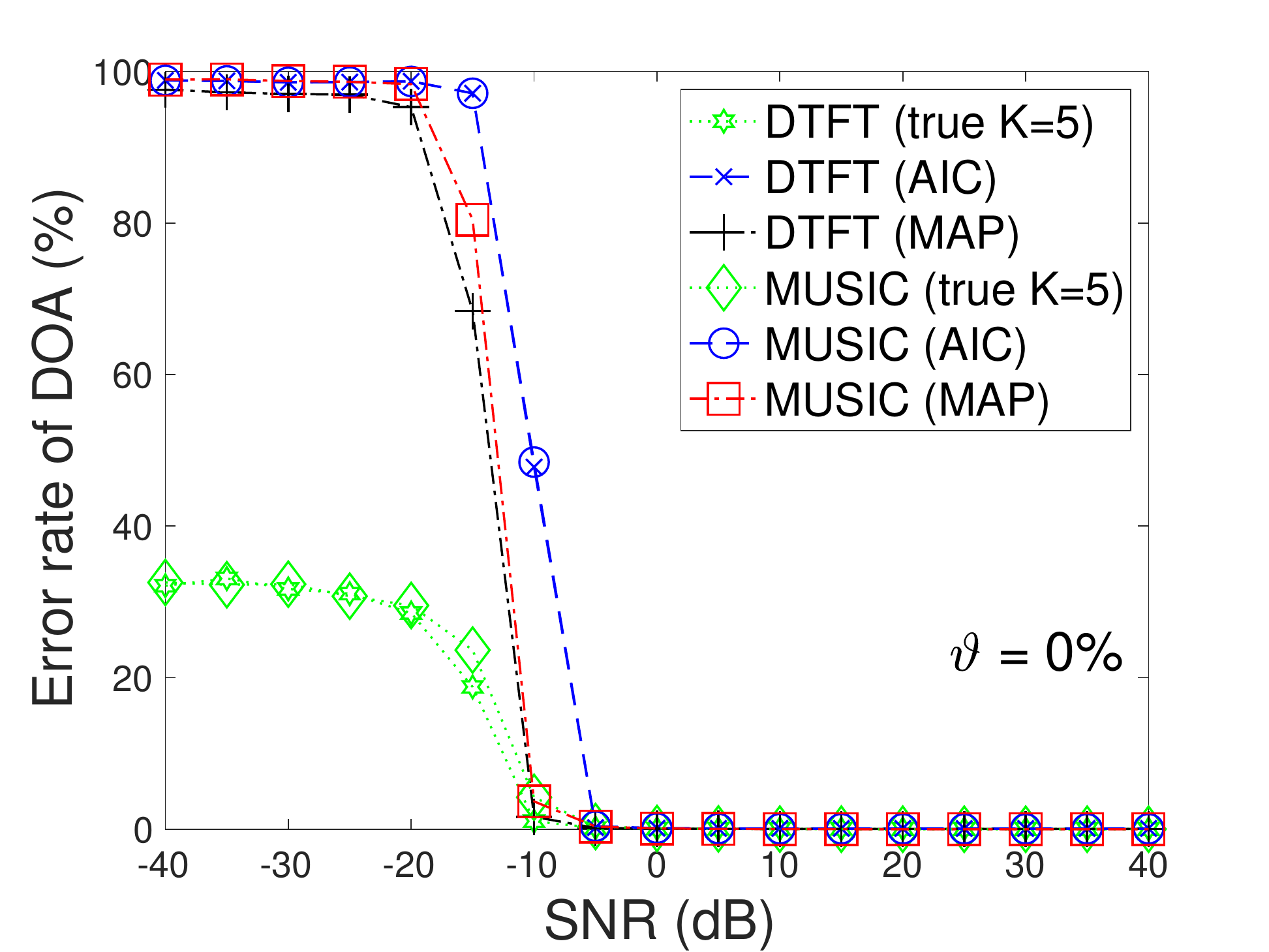}\includegraphics[width=0.5\columnwidth]{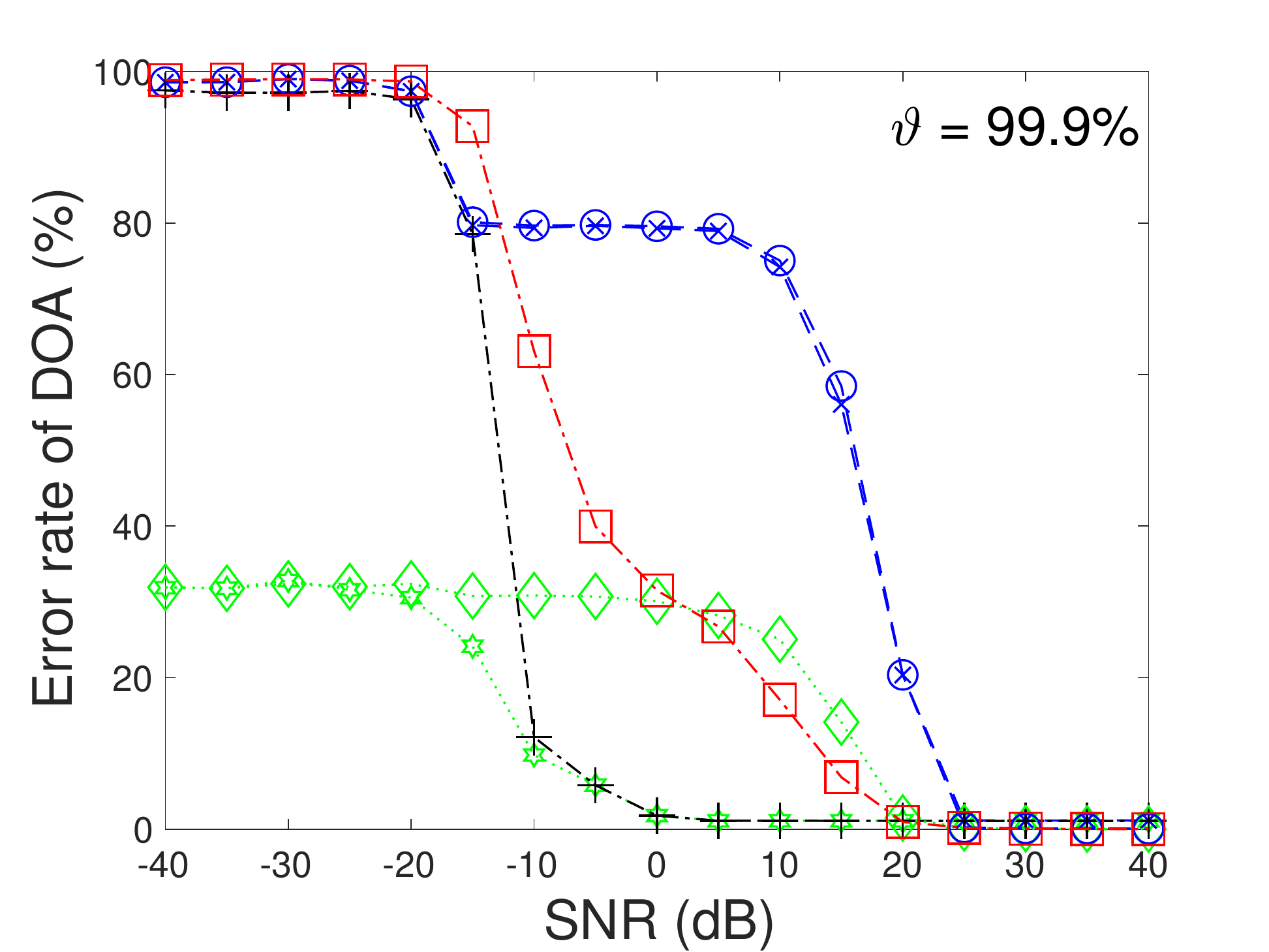}
\par\end{centering}
\begin{centering}
\includegraphics[width=0.5\columnwidth]{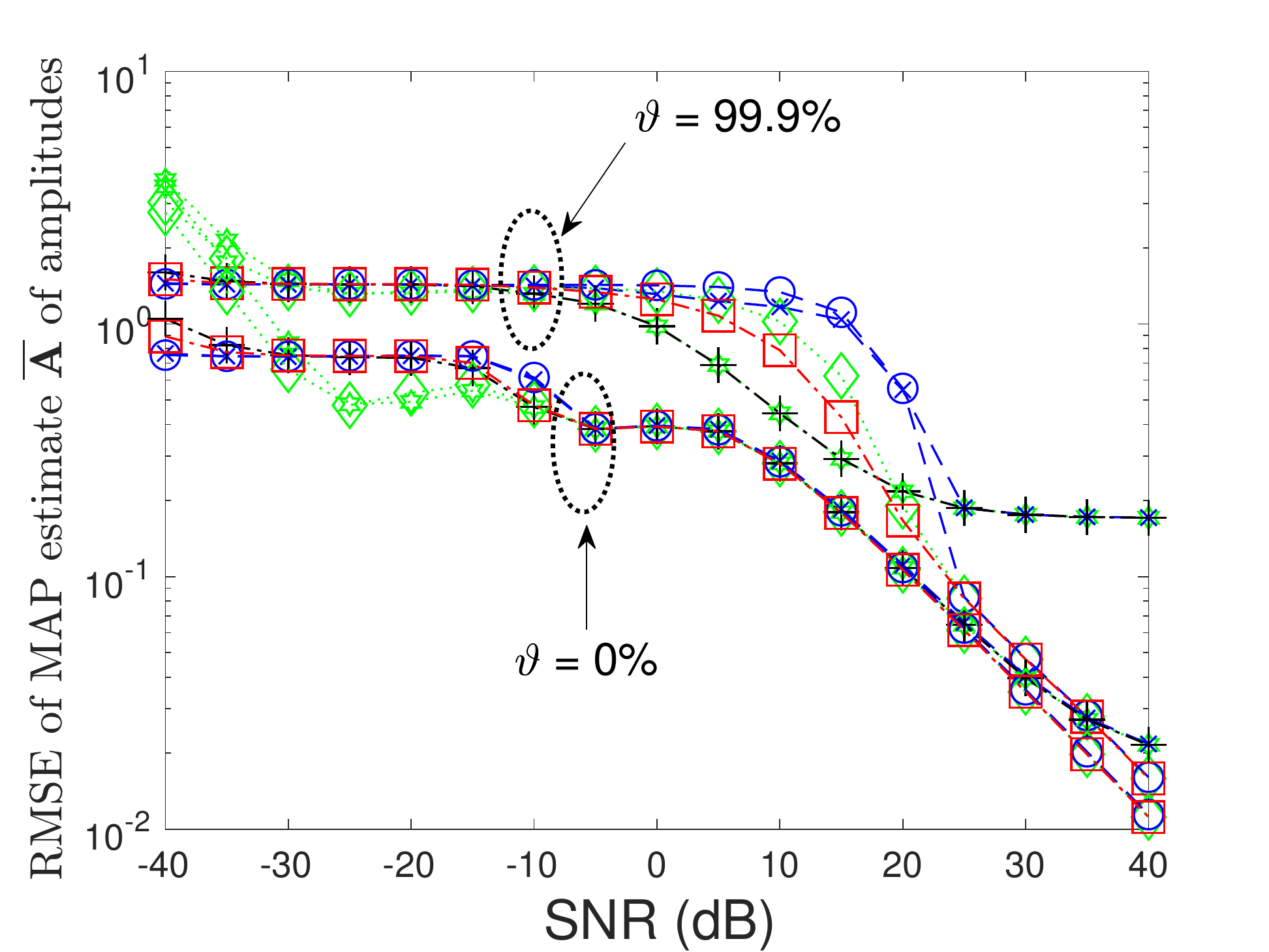}\includegraphics[width=0.5\columnwidth]{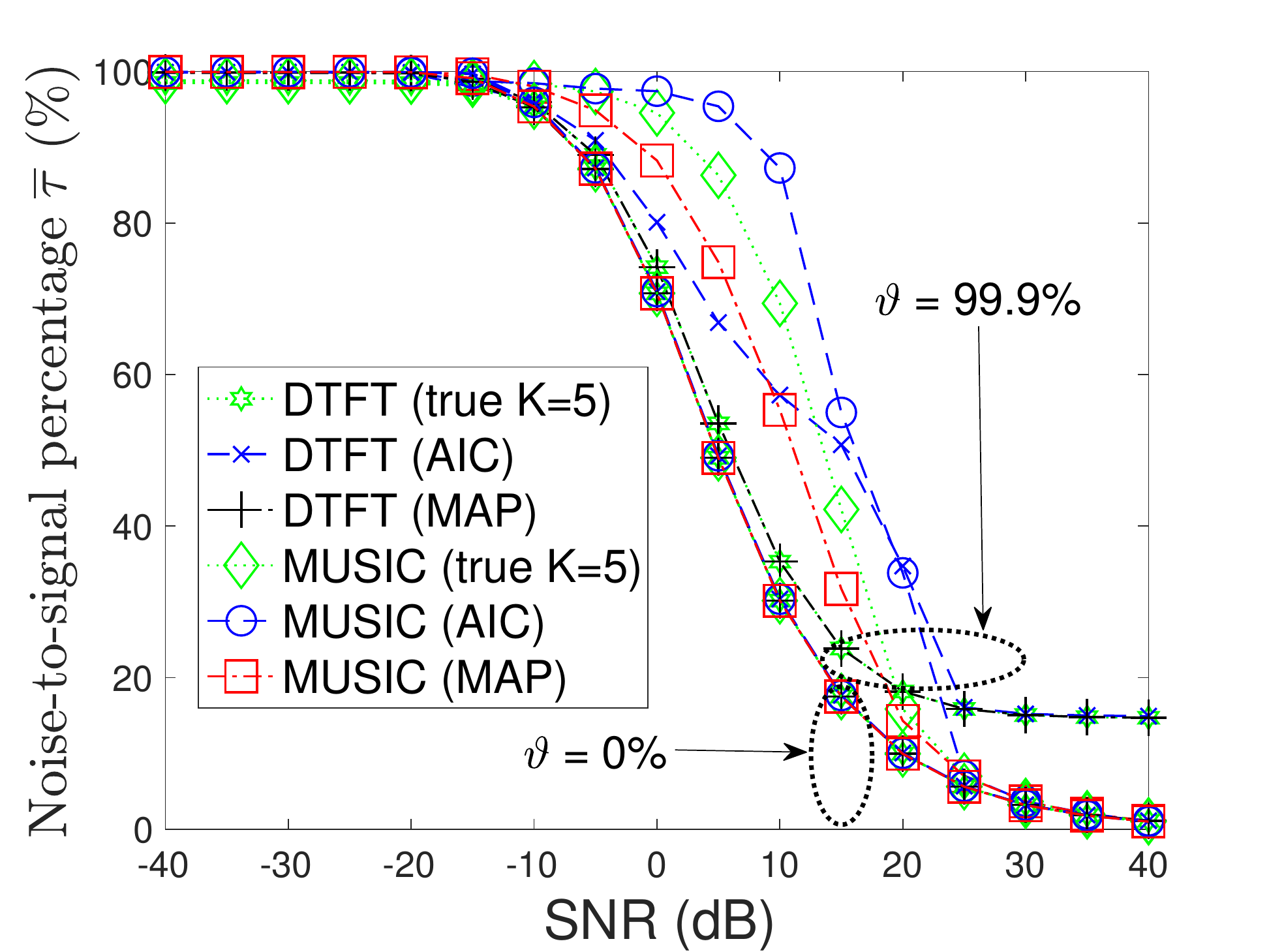}
\par\end{centering}
\caption{\label{fig:correlatedD}Simulations for the case of moderately correlated
DOAs ($\protect\ndim=15$ sensors), with the same default setting
in Fig. \ref{fig:Overlapping}. The first row is DTFT and MUSIC spectrums
for the case $\protect\SNR=-10$ (dB) and non-overlapping ($\vartheta=0\%$),
with the same convention in Fig. \ref{fig:OverlapSpectrum}. The legend
is the same for all figures in other rows.}
\end{figure}

\begin{figure}
\begin{centering}
\includegraphics[width=0.5\columnwidth]{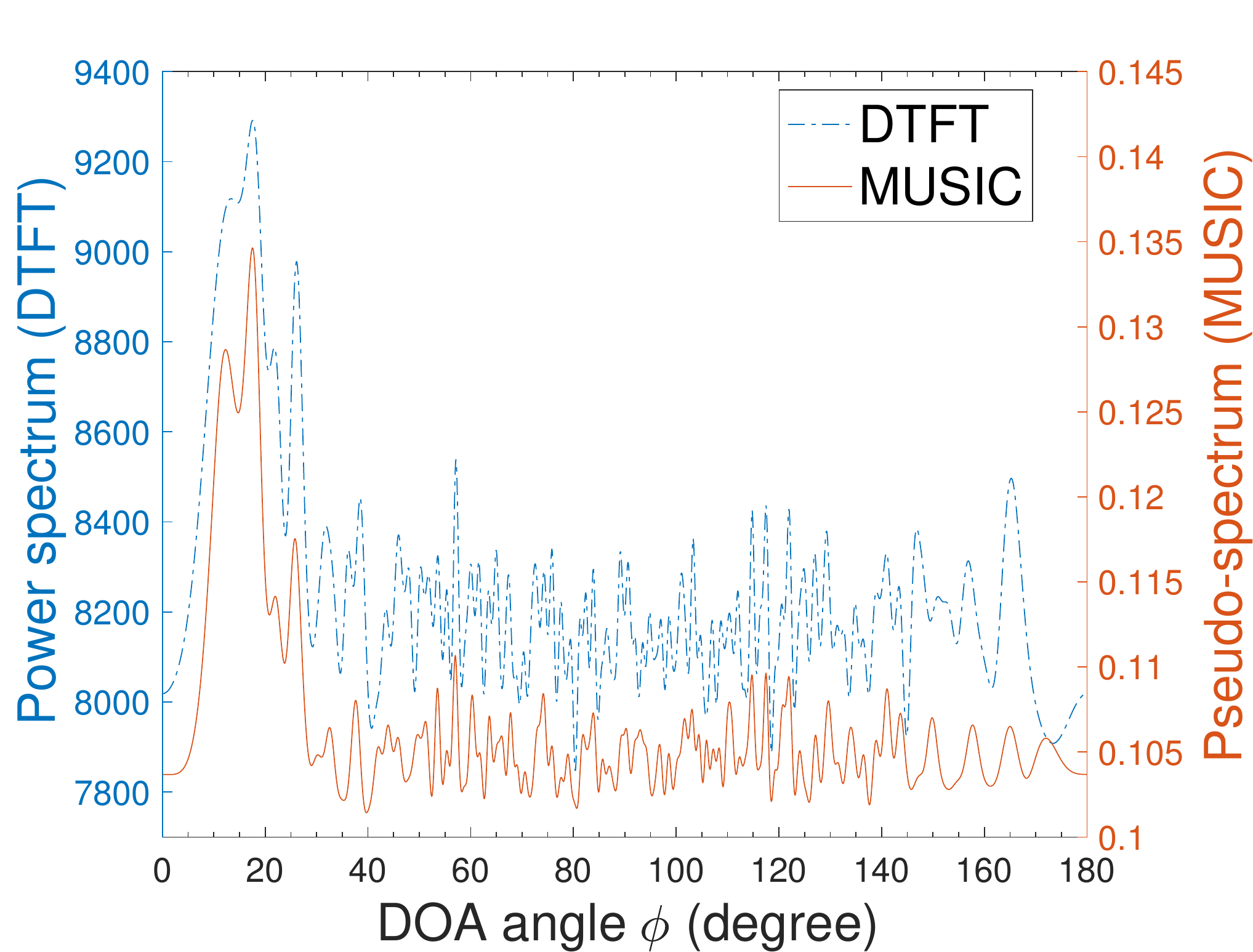}\includegraphics[width=0.5\columnwidth]{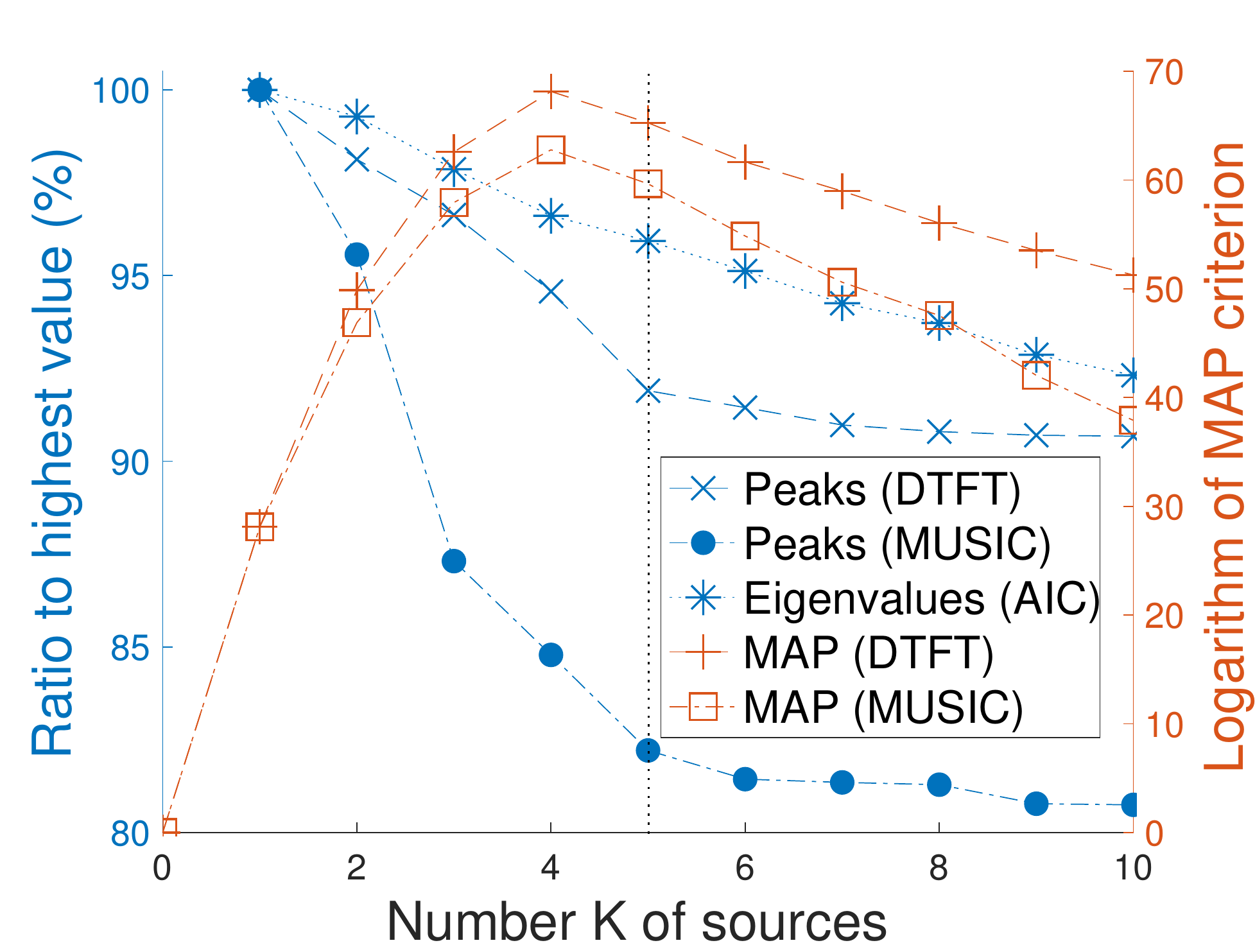}
\par\end{centering}
\begin{centering}
\includegraphics[width=0.5\columnwidth]{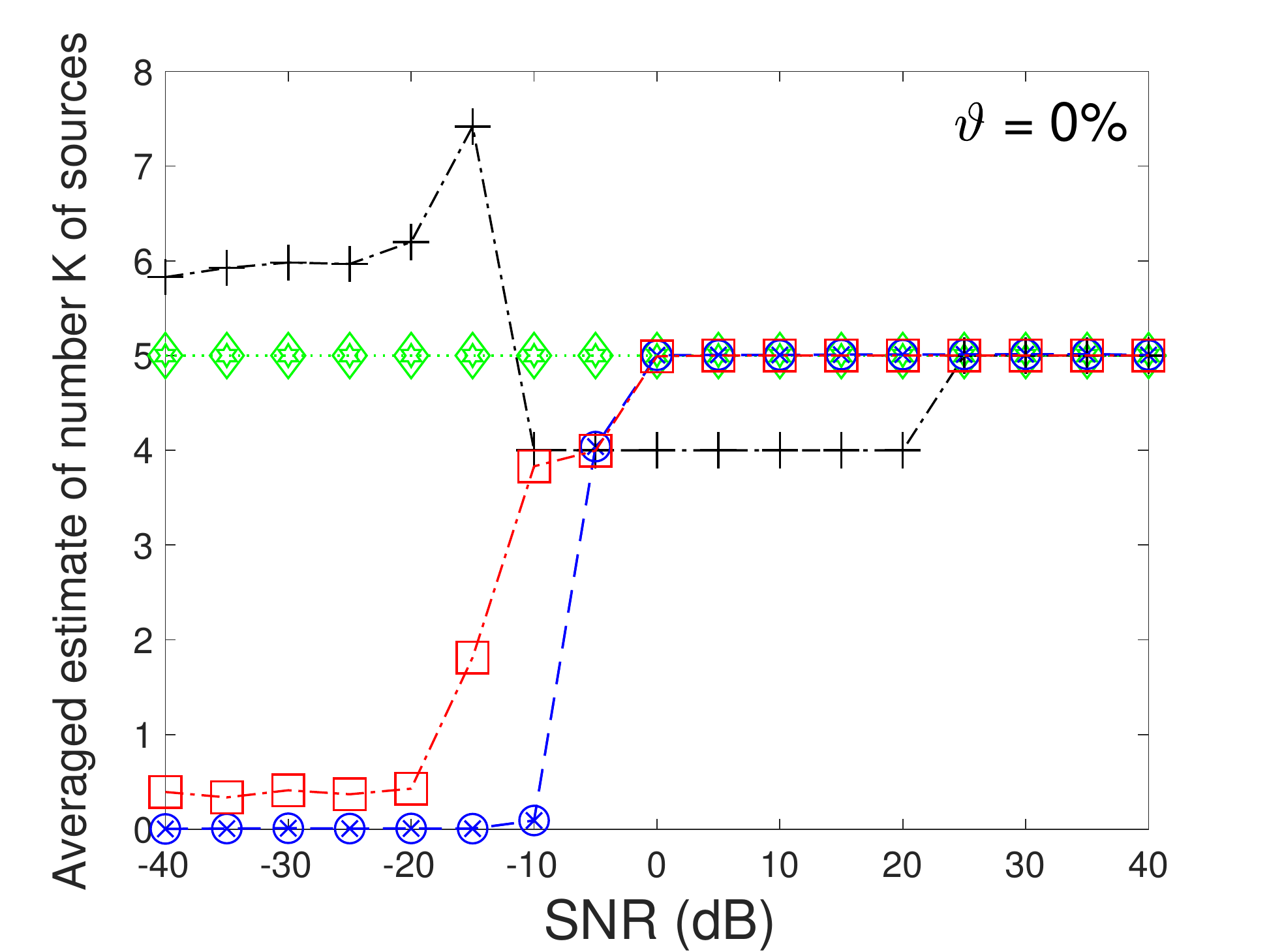}\includegraphics[width=0.5\columnwidth]{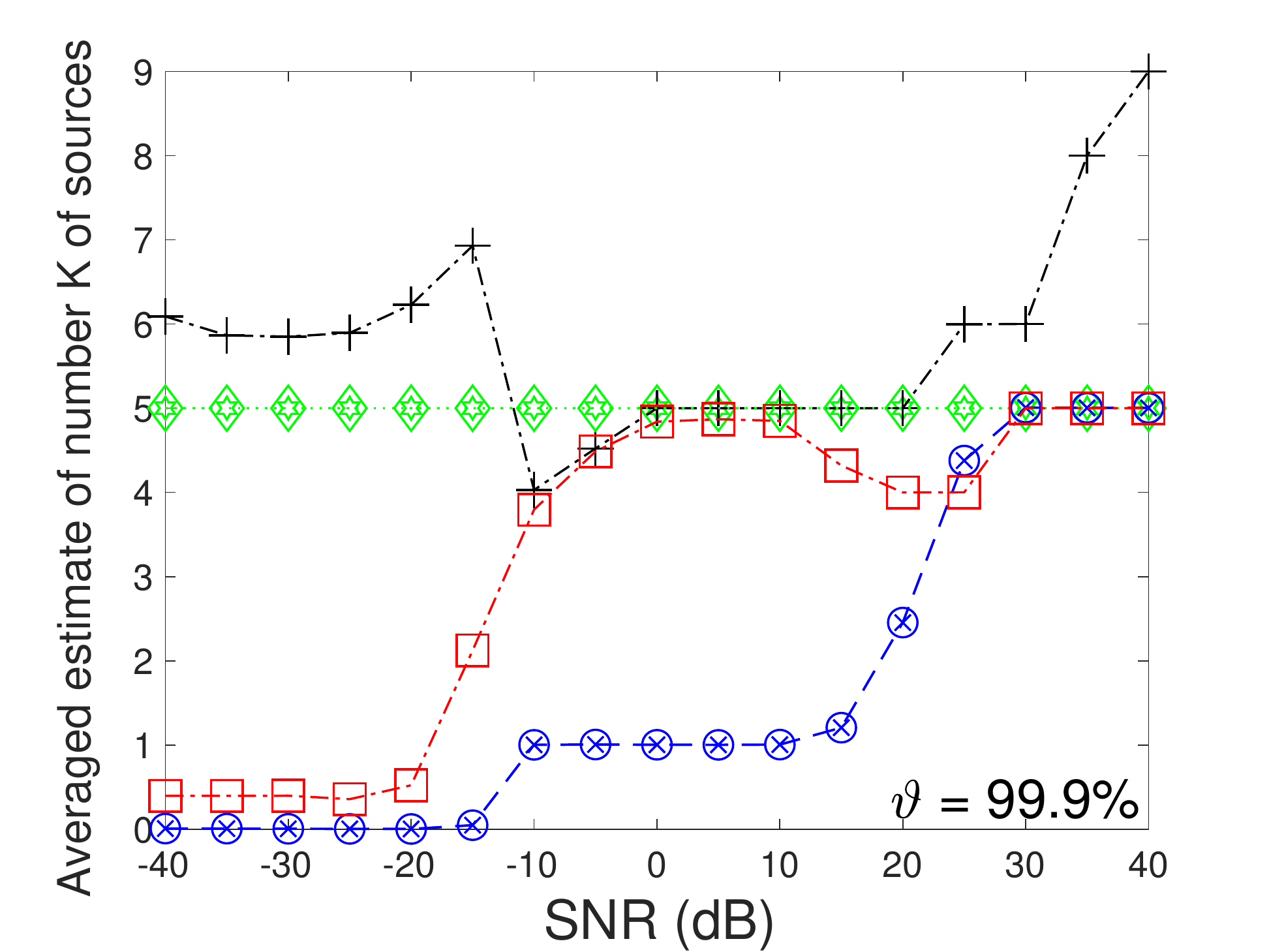}
\par\end{centering}
\begin{centering}
\includegraphics[width=0.5\columnwidth]{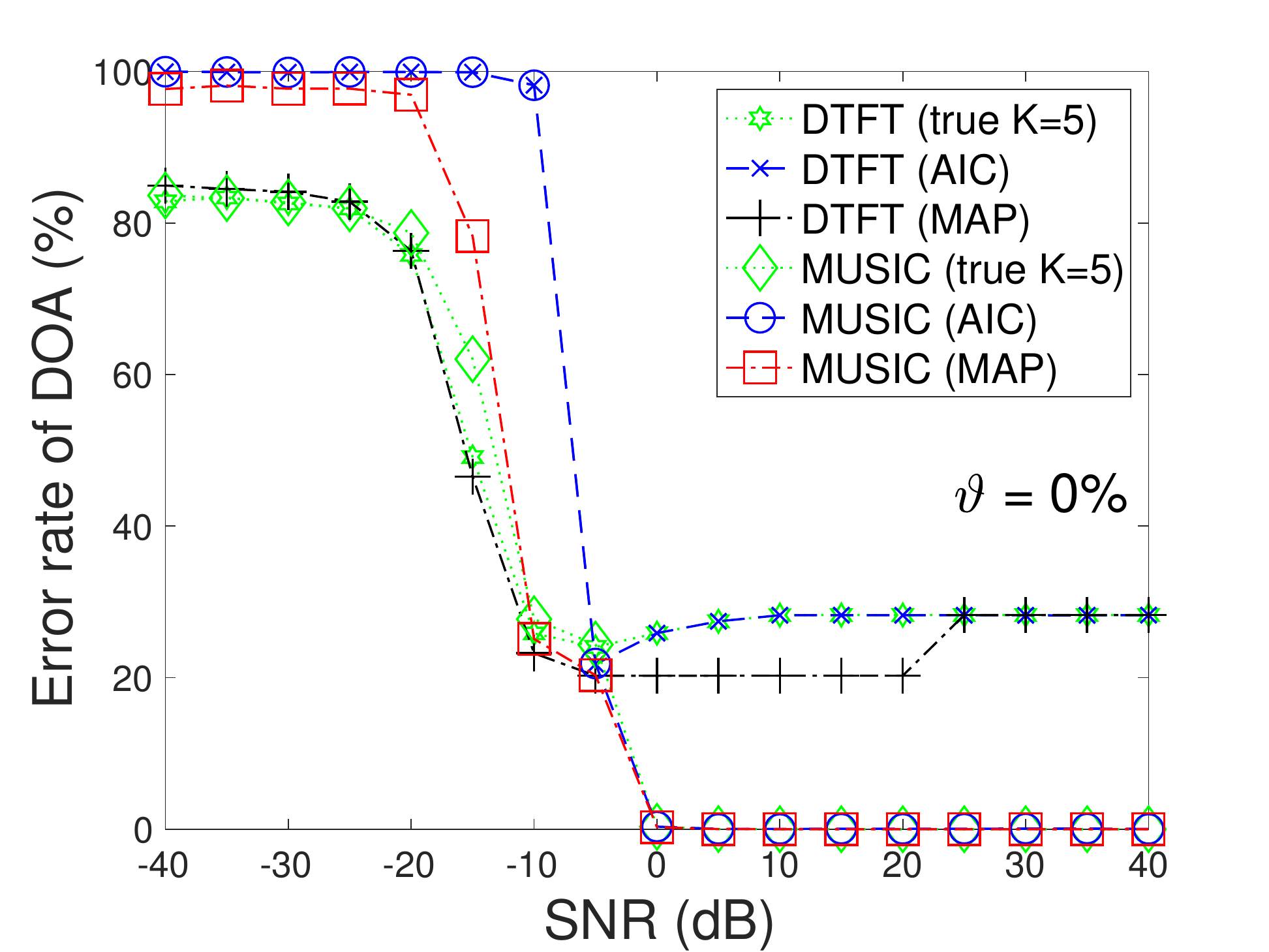}\includegraphics[width=0.5\columnwidth]{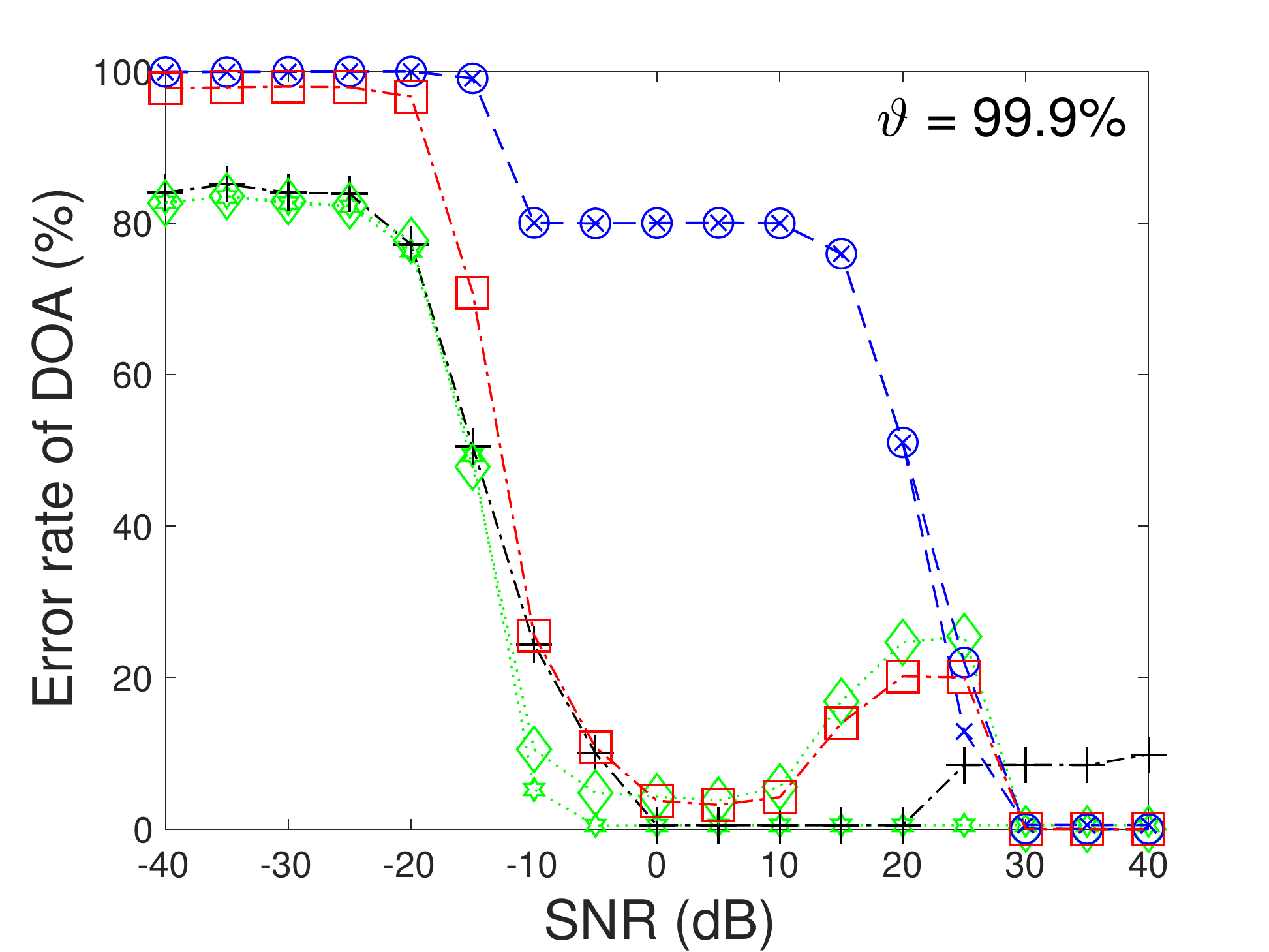}
\par\end{centering}
\begin{centering}
\includegraphics[width=0.5\columnwidth]{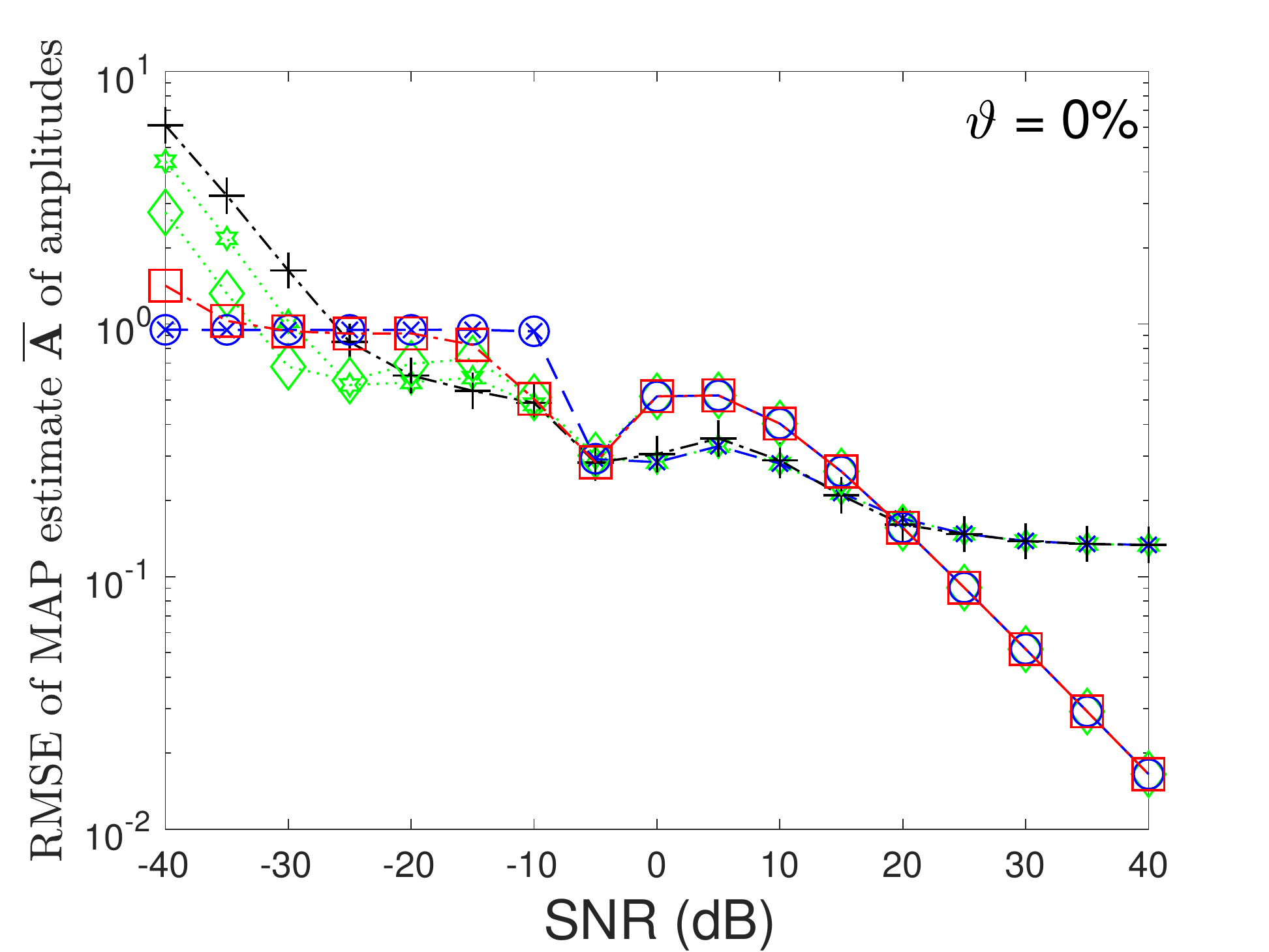}\includegraphics[width=0.5\columnwidth]{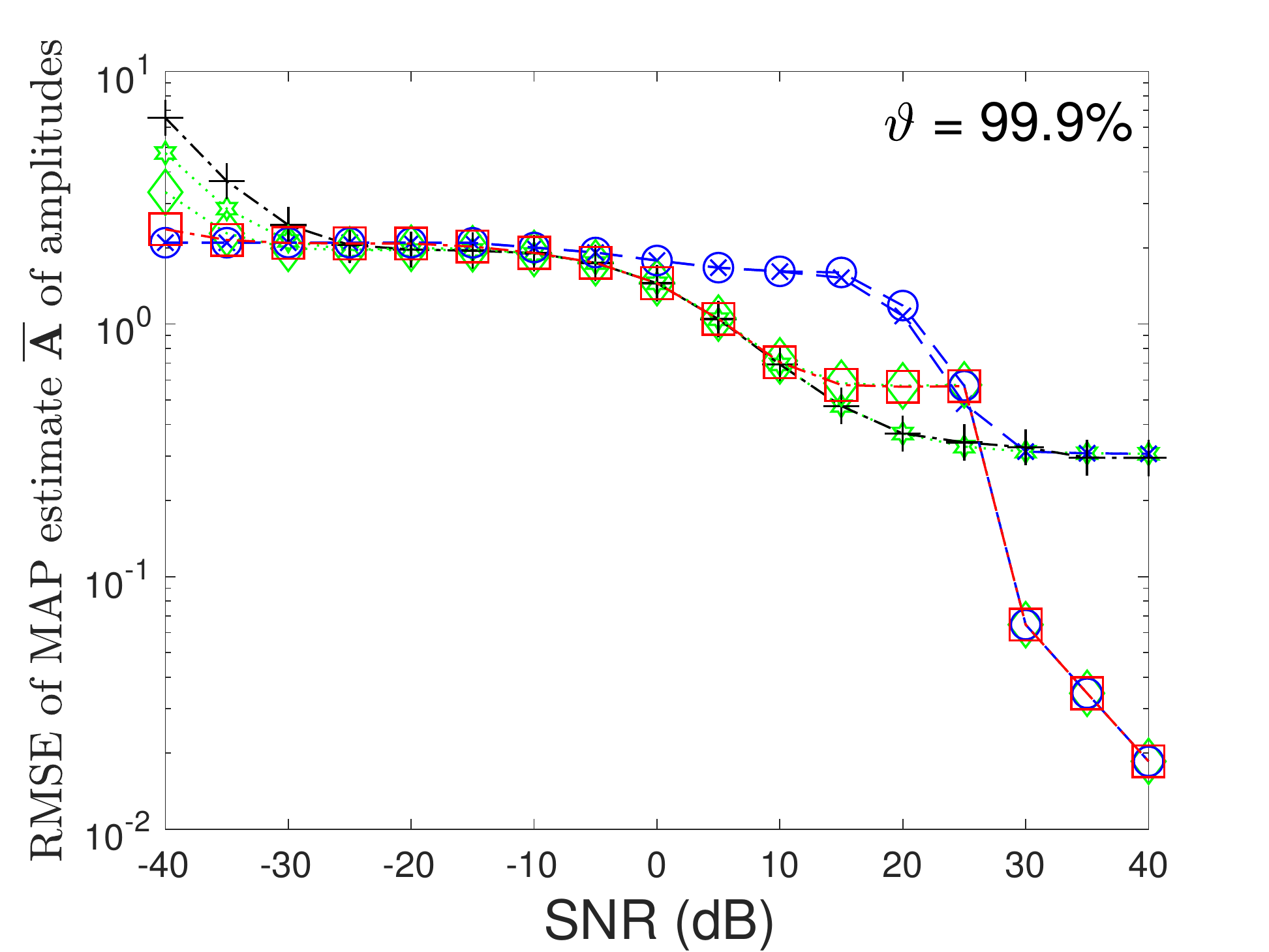}
\par\end{centering}
\begin{centering}
\includegraphics[width=0.5\columnwidth]{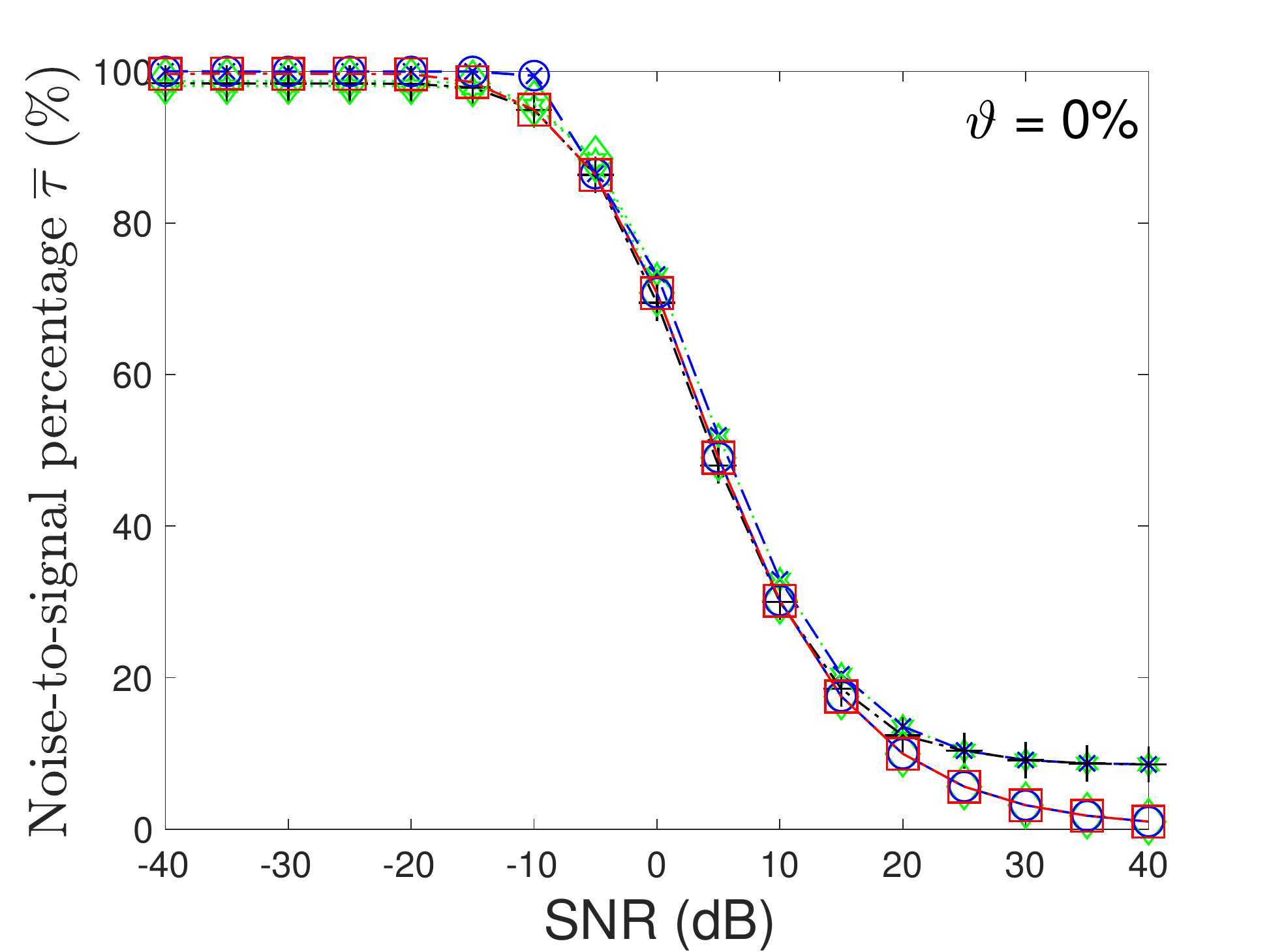}\includegraphics[width=0.5\columnwidth]{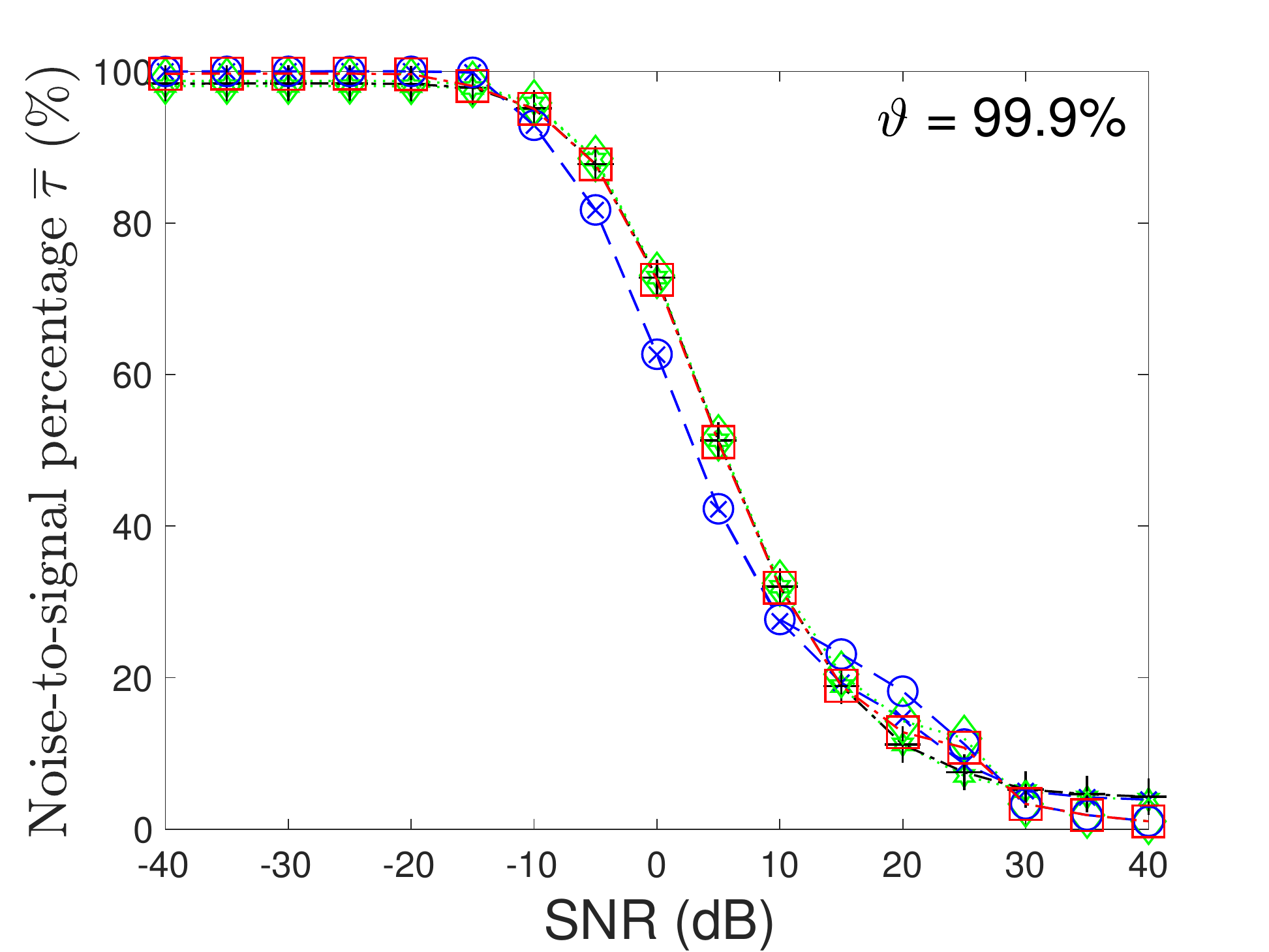}
\par\end{centering}
\caption{\label{fig:correlatedDOA}Simulations for the case of highly correlated
DOAs ($\Delta\protect\Angle=4^{0}$), with the same default setting
in Fig. \ref{fig:Overlapping}. The first row is DTFT and MUSIC spectrums
for the case $\protect\SNR=-10$ (dB) and non-overlapping ($\vartheta=0\%$),
with the same convention in Fig. \ref{fig:OverlapSpectrum}. The legend
is the same for all figures in other rows.}
\end{figure}

\begin{figure}
\begin{centering}
\includegraphics[width=0.25\textwidth]{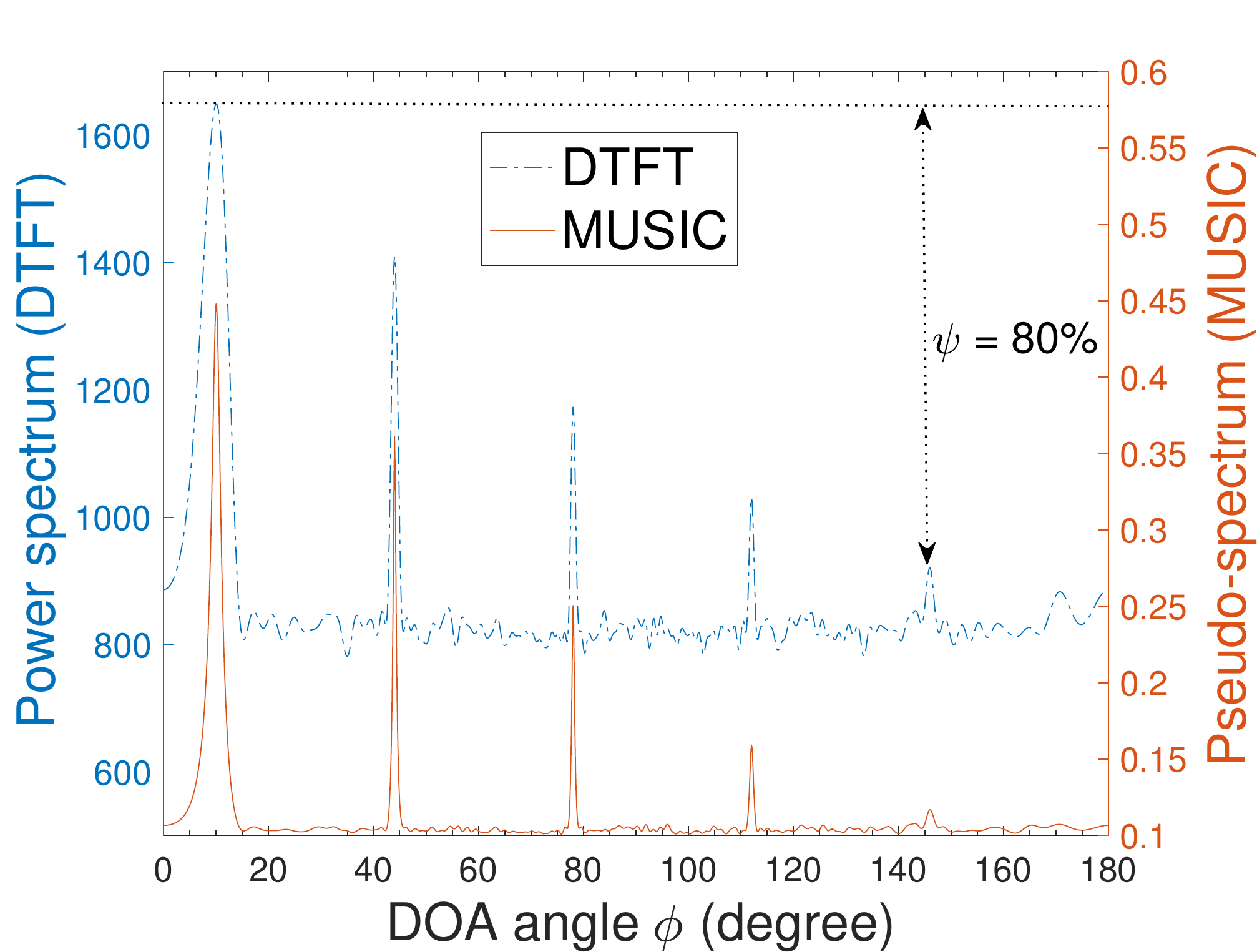}\includegraphics[width=0.25\textwidth]{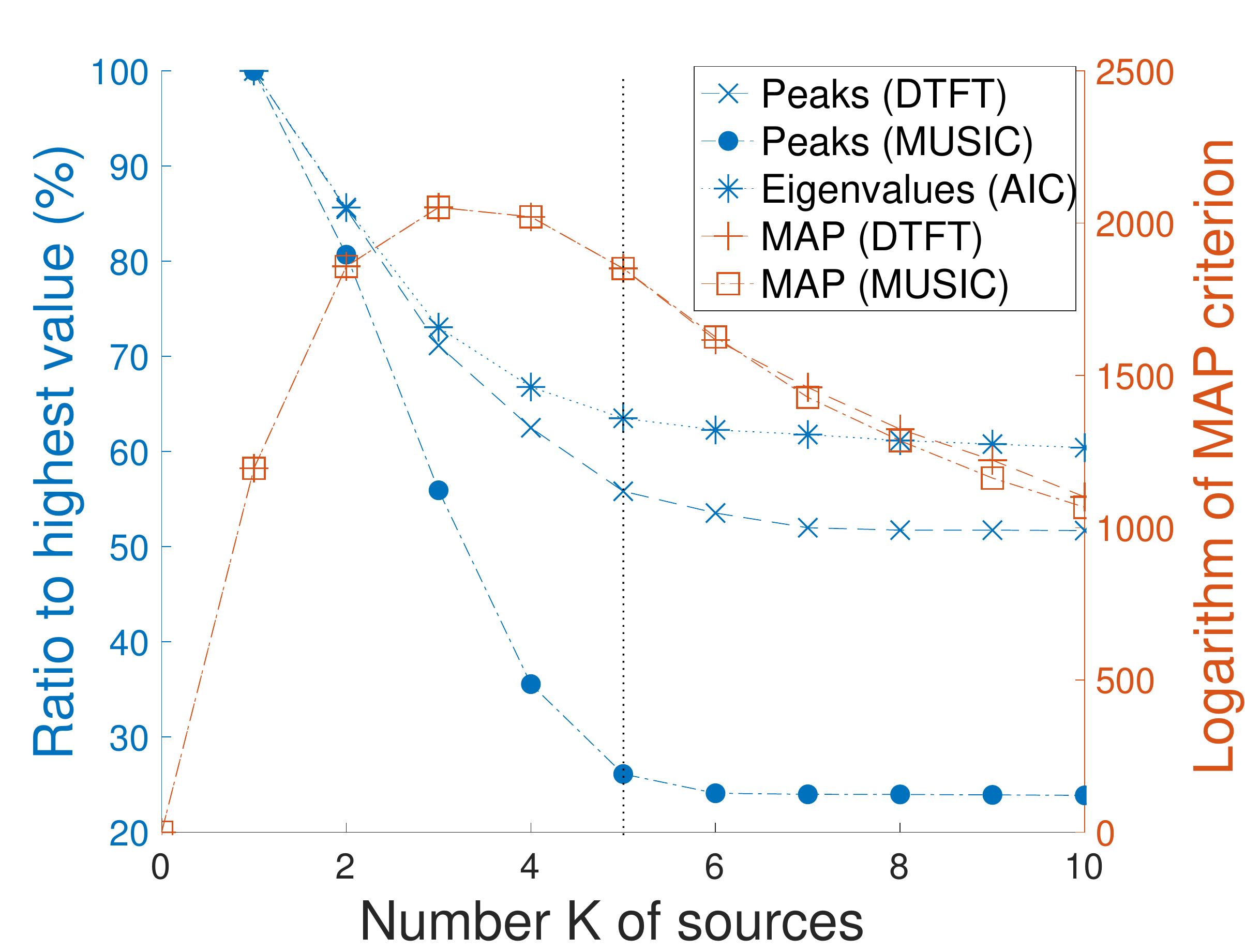}
\par\end{centering}
\begin{centering}
\includegraphics[width=0.25\textwidth]{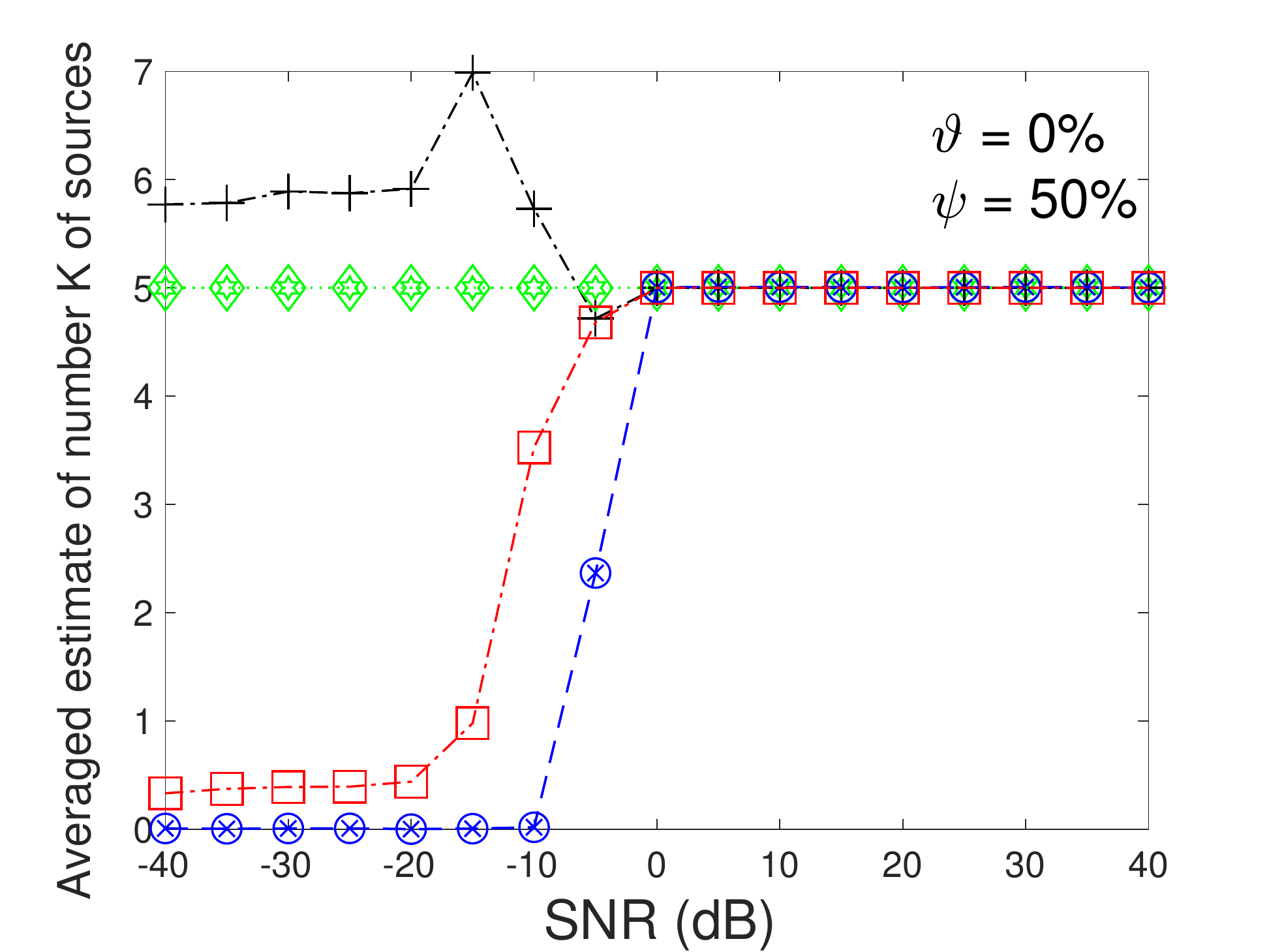}\includegraphics[width=0.25\textwidth]{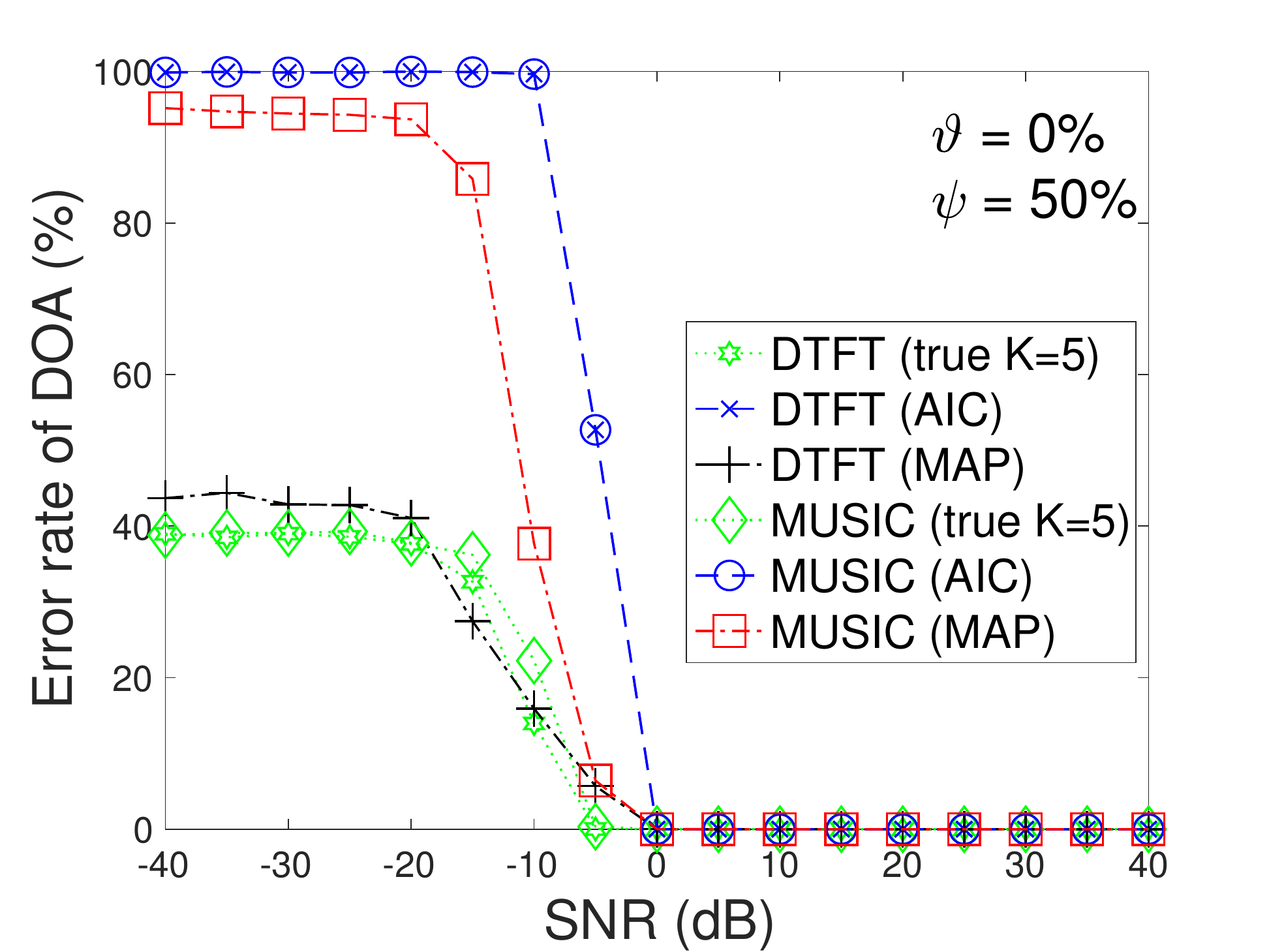}
\par\end{centering}
\begin{centering}
\includegraphics[width=0.25\textwidth]{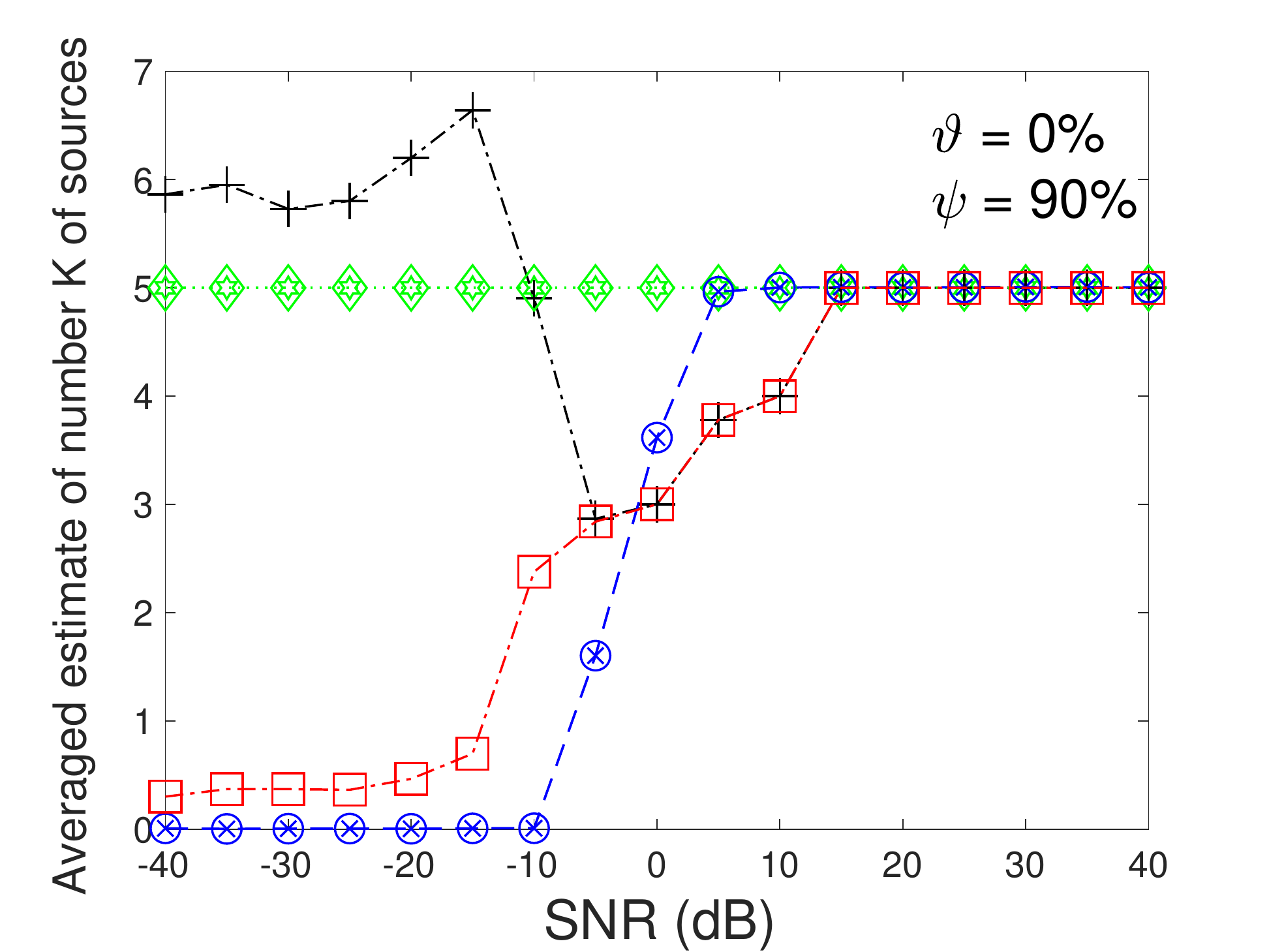}\includegraphics[width=0.25\textwidth]{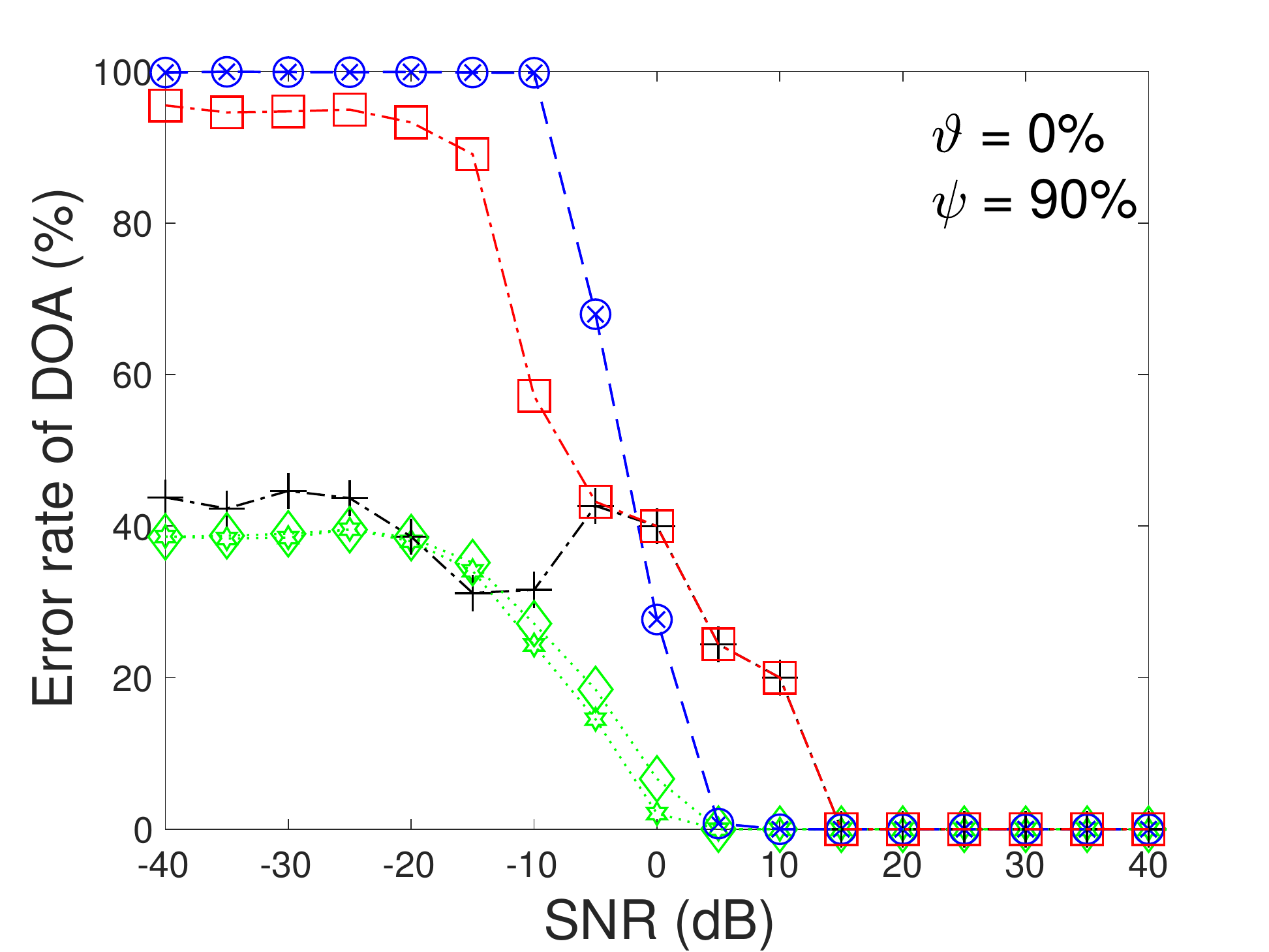}
\par\end{centering}
\begin{centering}
\includegraphics[width=0.25\textwidth]{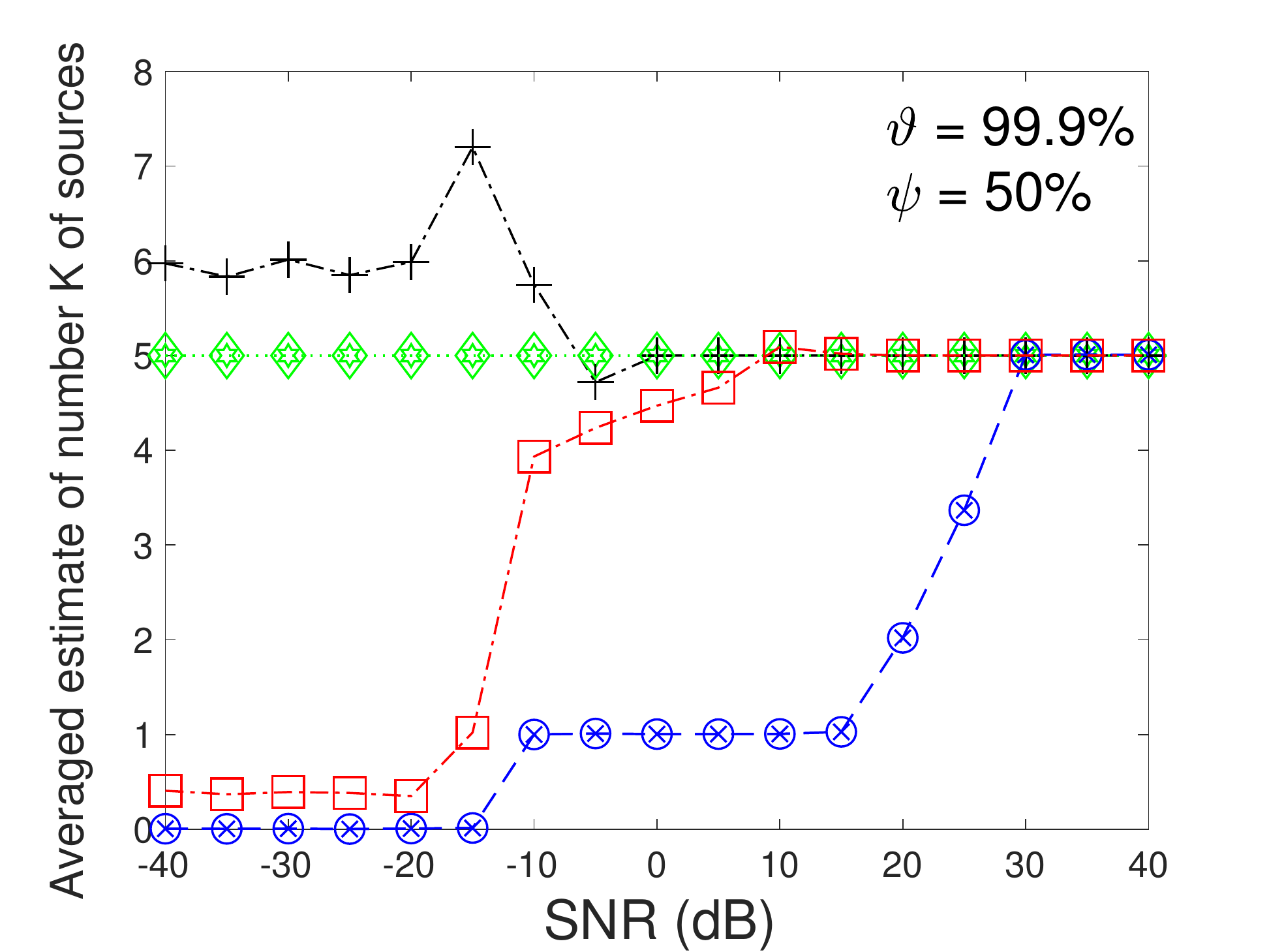}\includegraphics[width=0.25\textwidth]{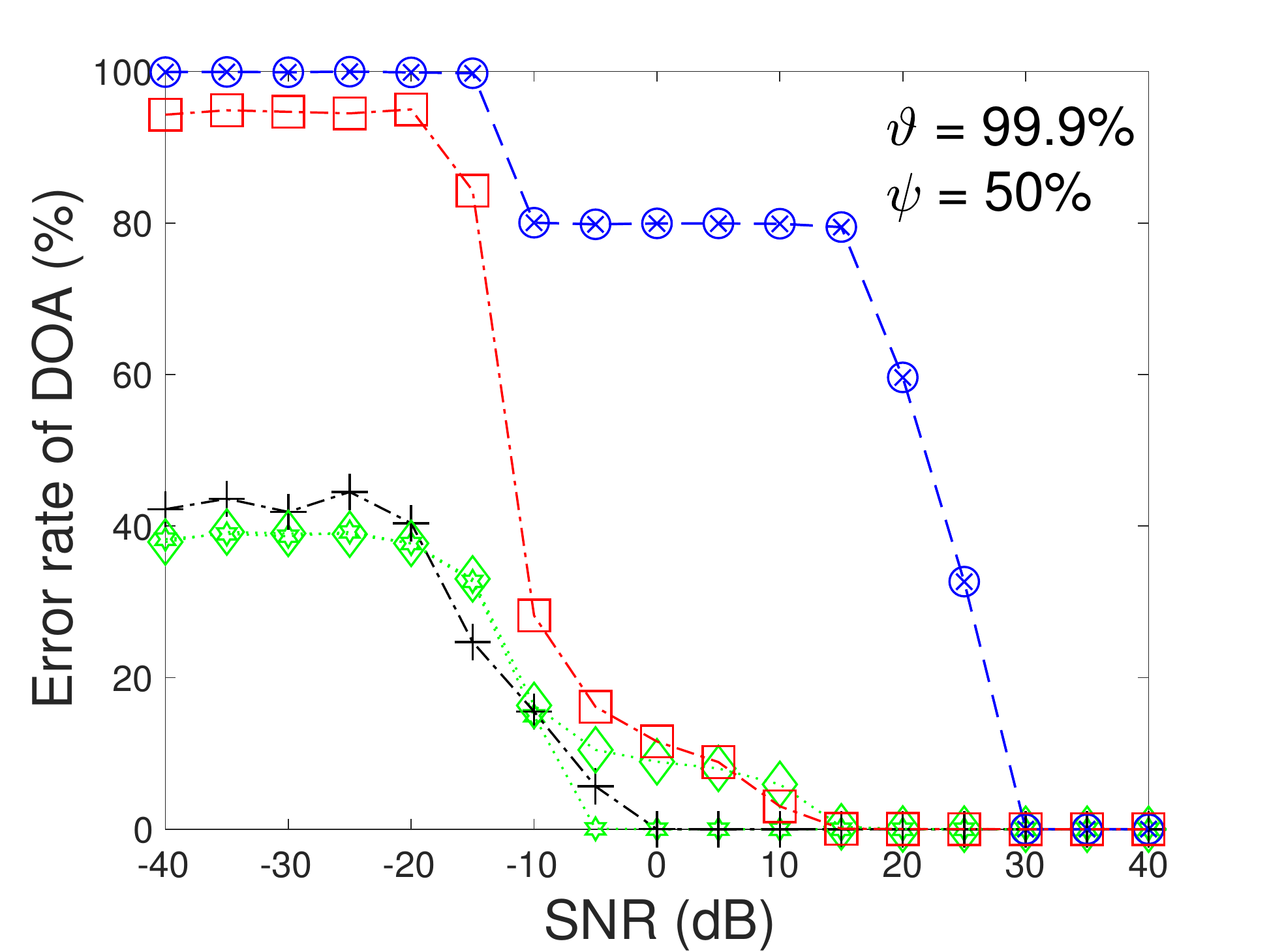}
\par\end{centering}
\begin{centering}
\includegraphics[width=0.25\textwidth]{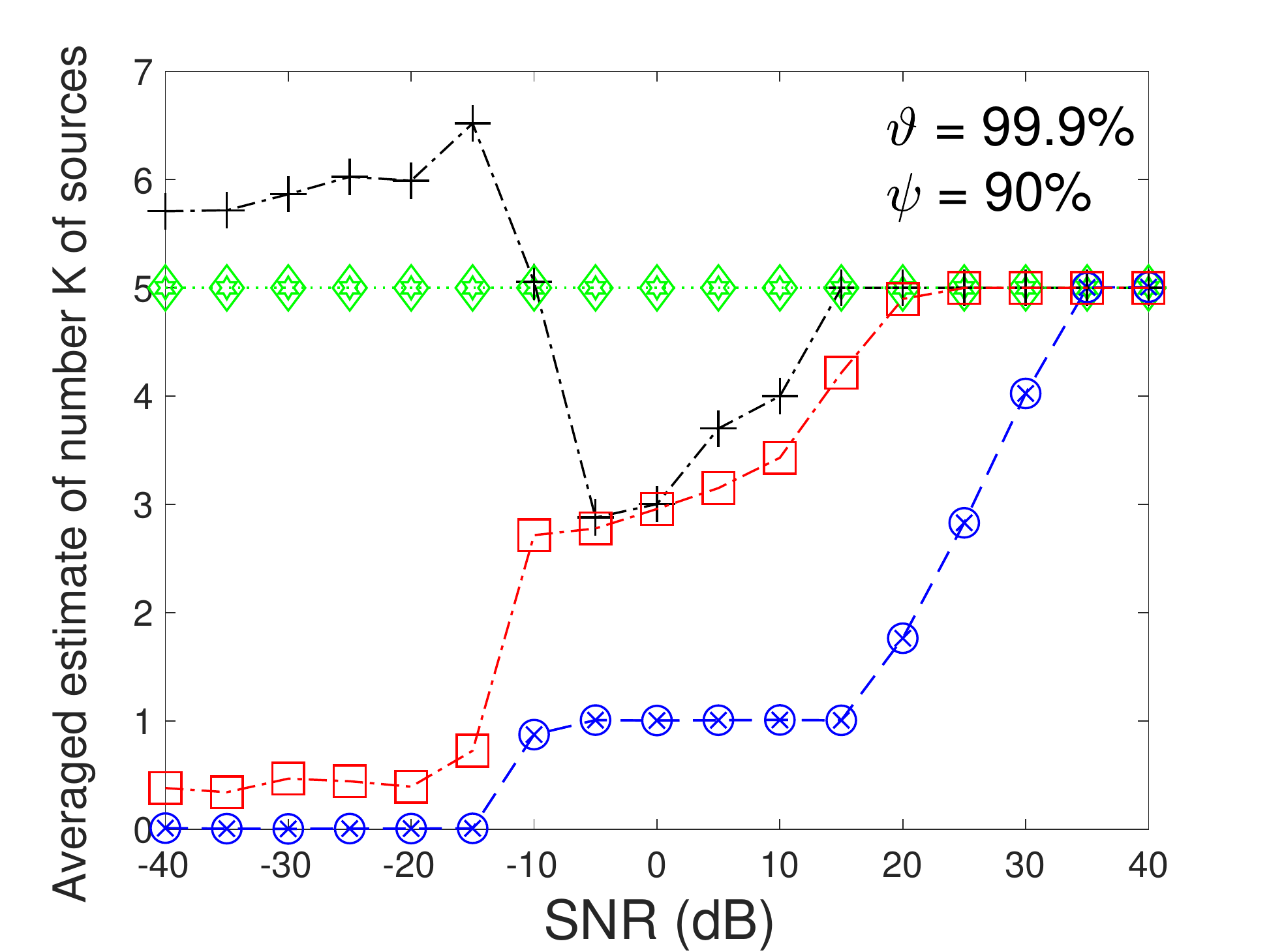}\includegraphics[width=0.25\textwidth]{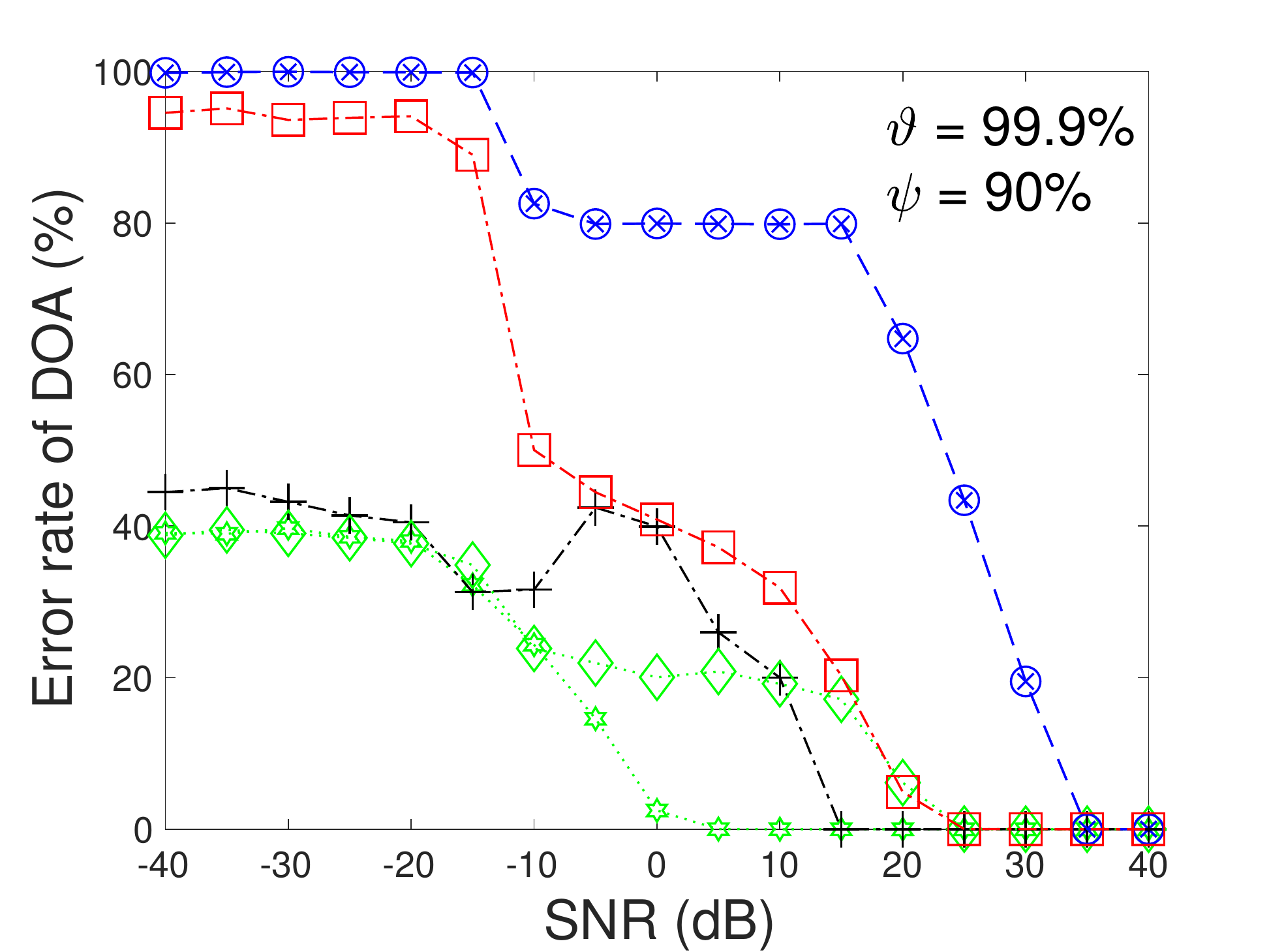}
\par\end{centering}
\begin{centering}
\includegraphics[width=0.25\textwidth]{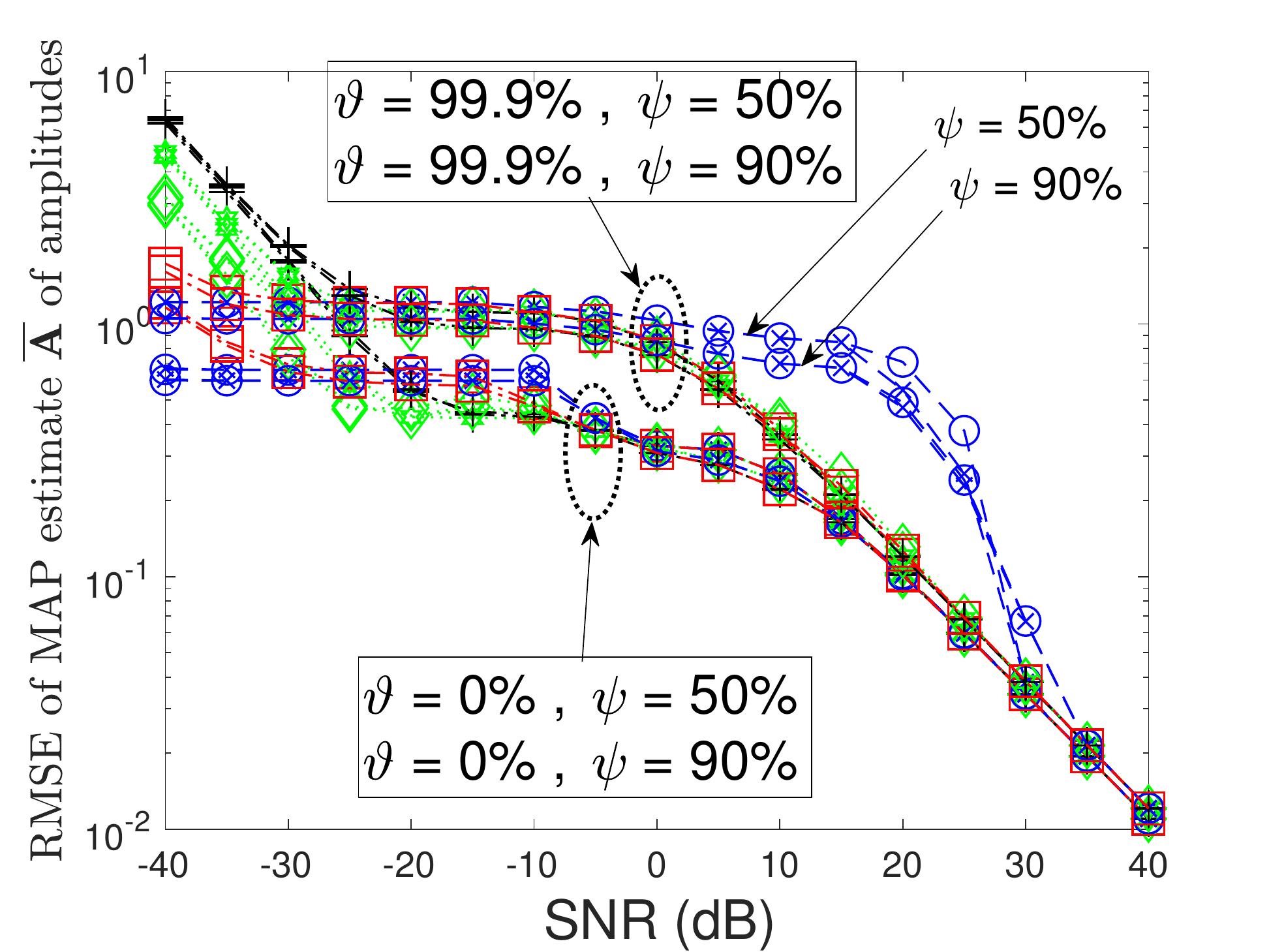}\includegraphics[width=0.25\textwidth]{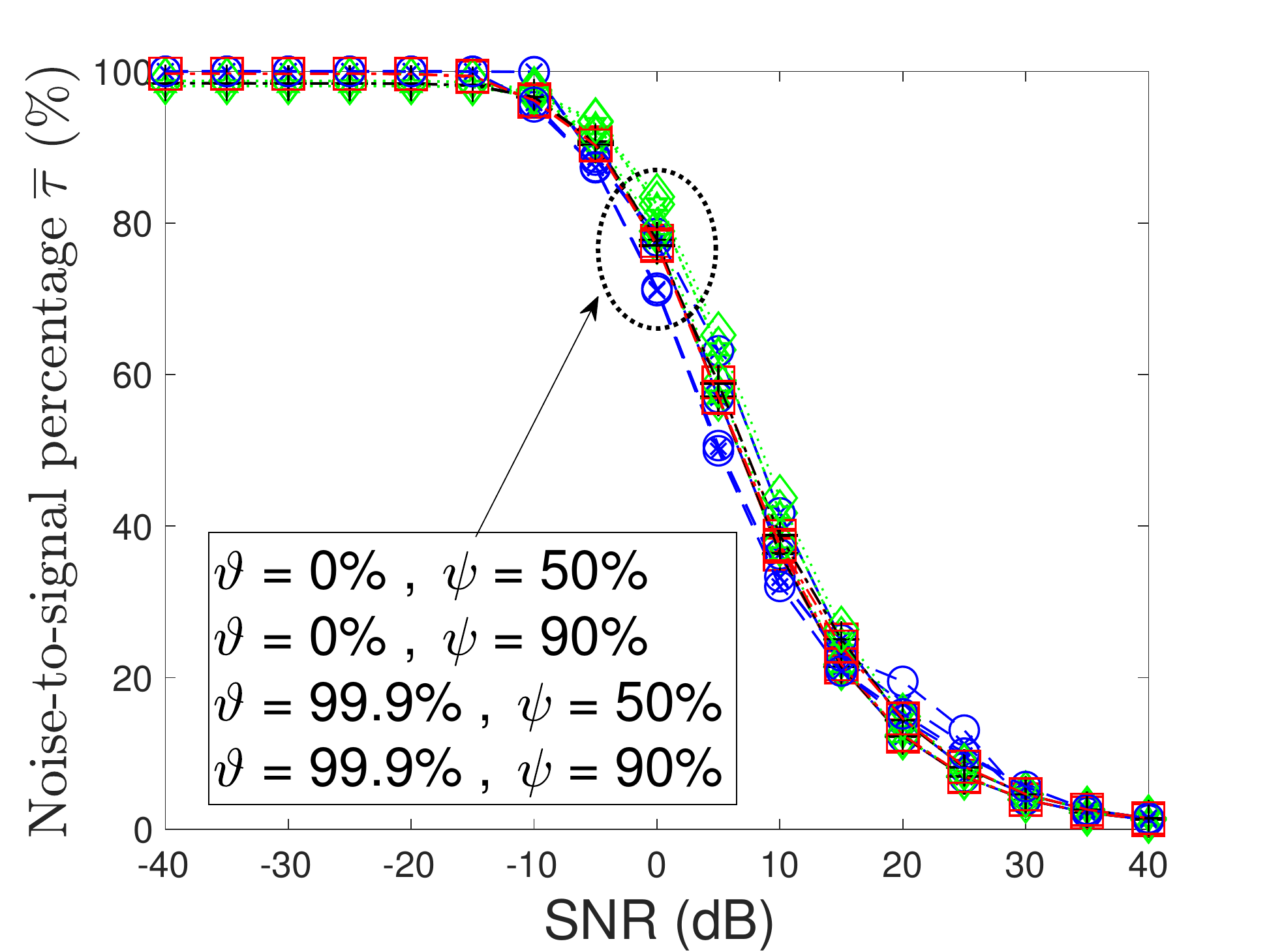}
\par\end{centering}
\caption{\label{fig:Decay}Simulations for the case of decayed amplitudes,
with the same legend and default setting in Fig. \ref{fig:Overlapping}.
The first row is DTFT and MUSIC spectrums for the case $\protect\SNR=0$
(dB), non-overlapping ($\vartheta=0\%$) and decayed ratio $\psi=80\%$,
with the same convention in Fig. \ref{fig:OverlapSpectrum}. The legend
is the same for all figures in other rows. }
\end{figure}

We can also see this phenomenon via eigen-decomposition (\ref{eq:PCA},
\ref{eq:Vmax}) of empirical covariance matrix $\yData\yData^{\Hermitian}\approx\bULA(\Amplitude\Amplitude^{\Hermitian})\bULA^{\Hermitian}$
in high SNR scenario. When $\vartheta\rightarrow0\%$, the matrix
$\Amplitude\Amplitude^{\Hermitian}$ becomes diagonal and, hence,
the $\istate$-th eigenvalue of $\yData\yData^{\Hermitian}$ would
be a good approximation of the $\istate$-th source's power $\NORMTWO{\bamplitude_{\istate}}=\bamplitude_{\istate}^{\Hermitian}\bamplitude_{\istate}$
in (\ref{eq:MATRIX_MODEL}, \ref{eq:SNR}). When $\vartheta\rightarrow100\%,$
however, the matrix $\Amplitude\Amplitude^{\Hermitian}$ is close
to a constant matrix, whose rank is one. The eigenvalues of $\yData\yData^{\Hermitian}$
are not good approximations of source's powers $\NORMTWO{\bamplitude_{\istate}}$
anymore and most of the eigenvalues deteriorate to zero in this case,
as shown in Fig. \ref{fig:OverlapSpectrum}. Hence, hard-threshold
eigen methods do not yield good estimation for the number $\nstate$
of sources in overlapping case $\vartheta\rightarrow100\%$, even
with infinite amount of data. Since AIC for PCA is an eigen-based
method, as shown in \cite{DoA:AIC:VanTrees:02}, its performance decreases
when the overlapping ratio $\vartheta$ increases, as shown in Figs.
\ref{fig:Overlapping}-\ref{fig:Decay}.


\subsubsection{MAP estimate versus AIC}

The simulations with default setting are given in Fig. \ref{fig:Overlapping}.
Since our MAP estimate $\widehat{\nstate}$ (\ref{eq:K_MAP_MUSIC})
is not a hard-threshold eigen method, the performance of DTFT and
MUSIC with $\widehat{\nstate}$ is almost the same for any overlapping
ratios $\vartheta$ and, hence, is far superior to that of the AIC
method in all cases. 

In Fig. \ref{fig:Overlapping}a, it is difficult to estimate $\nstate$
correctly when $\SNR<-10$ dB, i.e. $\max_{\istate}\sigma_{\istate}<\frac{\stdZero}{\sqrt{10}}=\frac{\stdZero}{3.16}$
in (\ref{eq:SNR}). Hence, intuitively, when projected noise's deviation
$\stdZero$ on signal subspace is higher than three empirical deviation
of any source's amplitudes, the noise would completely dominate the
signal and it is very hard to extract the signal from noisy data. 

Given the estimate of $\nstate$ in Fig. \ref{fig:Overlapping}a,
the performance of DOA's estimation via (\ref{eq:MAP_V_Stiefel})
is shown in Fig. \ref{fig:Overlapping}b. Here we can see that the
MAP estimate $\widehat{\nstate}$ yields significant improvement for
DOA's estimation accuracy in DTFT and MUSIC algorithms. Although DTFT
yields overfitting for the MAP estimate $\widehat{\nstate}$ in Fig.
\ref{fig:Overlapping}a, its performance is closer to the case of
known $\nstate$ than the MUSIC and AIC methods. Nonetheless, this
is mainly owing to the imperfection of our clustering-based error-rate
of DOAs in (\ref{eq:ERR}), which becomes lower when there are more
estimated sources close to true source's DOA. Although this error
rate is good enough for high SNR, care should be taken for the case
of low SNR. Hence, it is safe to say that the credibility of estimated
DOAs is very low if $\SNR<-10$ dB.

Given estimates of DOAs and $\nstate$, the RMSE for posterior mean
(\ref{eq:noise_MSE}) of noise's deviation $\std$ is given in Fig.
\ref{fig:Overlapping}c. In non-overlapping case, all methods yield
good estimates for~$\std$. In overlapping cases, our MAP estimate
$\widehat{\nstate}$ helps DTFT maintain the same performance. In
contrast, the eigen-based AIC method yields poor estimates for $\std$
in overlapping cases, particularly in high SNR. Likewise, since MUSIC
is an eigen-based method for the MAP estimation $\widehat{\bULA}$
in (\ref{eq:MUSIC}, \ref{eq:MAP_V_Stiefel}), it yields worse estimates
for $\std$ in overlapping cases, even with known $\nstate$. Nonetheless,
our MAP estimate $\widehat{\nstate}$ is still much better than AIC
in middle SNR with $\vartheta\leq99.9\%$. 

From estimate of noise's deviation in Fig. \ref{fig:Overlapping}c,
the estimate $\overline{\rationoise}$ in (\ref{eq:ratio_noise_MSE})
is then illustrated in Fig. \ref{fig:Overlapping}d. Since SNR is
usually unknown in practice, this estimate $\overline{\rationoise}$
is a good indicator of credibility for estimates of $\nstate$ and
DOAs. Indeed, the estimate $\overline{\rationoise}$ is consistently
around $90\%$, i.e. $\stdZero\approx3\stdamplitude$ in (\ref{eq:ra_PCA}),
when SNR is around $-10$ (dB). In completely overlapping case, however,
the AIC method yields bad estimate for $\overline{\rationoise}$  with
high SNR.

Given estimates of $\nstate$ and DOAs in Fig. \ref{fig:Overlapping}a-b,
the RMSE (\ref{eq:MSE_A}) of ML estimate $\AmpZero$ is plotted in
Fig. \ref{fig:Overlapping}e. As expected, the eigen-based AIC method
is the worst method in overlapping cases, while MAP estimate $\widehat{\nstate}$
maintains the good performance for DTFT and MUSIC in all cases of
$\vartheta$. The DTFT spectrum, when combined with MAP estimate $\widehat{\nstate}$,
is better than eigen-based MUSIC in the case $\vartheta=100\%$ with
very high SNR. 

When $\SNR\leq-10$ dB, the AIC method cannot detect any source in
Fig. \ref{fig:Overlapping}a and, hence, returns zero values for estimated
amplitudes in (\ref{eq:MSE_A}). This explains the low RMSE line for
amplitude's estimate of the AIC method in Fig. \ref{fig:Overlapping}e.
Despite being artificial, this zero value of amplitude's estimate
in low SNR is actually a better estimate of amplitudes in terms of
RMSE. Indeed, as shown in Fig. \ref{fig:Overlapping}f, the MAP estimate
$\overline{\Amplitude}=(1-\overline{\rationoise})\AmpZero$ in (\ref{eq:A_tau})
yields lower RMSE than ML estimate $\AmpZero$ since $\overline{\Amplitude}$
can automatically switch to zero value if SNR is too low, which is
indicated by the estimate $\overline{\rationoise}$ in Fig. \ref{fig:Overlapping}d. 

For illustration, the critical case of $\SNR=-10$ dB in Fig.~\ref{fig:Overlapping}
is shown in Fig.~\ref{fig:OverlapSpectrum}. We can see that the
peaks in DTFT and MUSIC spectrums linearly decrease with $\nstate$
and, hence, there is no clear difference between noise's peaks and
signal's peaks around $\nstate=5$ in this low SNR regime. It is then
difficult to extract the correct number $\nstate=5$ of sources from
these spectrum's peaks. In contrast, the MAP criterion (\ref{eq:K_MAP_MUSIC})
reaches the peak at $\widehat{\nstate}=5$, since it can return the
maximum difference among all possible binomial combinations of signal
and noise subspaces via (\ref{eq:Likeli=00005BV,K=00005Da}-\ref{eq:Likeli=00005BV,K=00005Db}).
The MAP estimate $\widehat{\nstate}$ is therefore the best estimate
in this case.


\subsection{Correlated multi-tone sources}

As explained above, both DTFT spectrum and MUSIC algorithm can yield
good estimates for all cases of uncorrelated DOAs, when combined with
MAP estimate $\widehat{\nstate}$. The eigen-based AIC method, however,
yields the worst estimates for overlapping multi-tone sources.

In this subsection, let us consider the case of correlated DOAs. Since
the uncorrelated condition (\ref{eq:uncorrApprox}, \ref{eq:PCA})
for PCA and MUSIC models is violated in this case, their estimates
$\widehat{\bULA}$ and $\widehat{\nstate}$ in (\ref{eq:MAP_V_Stiefel},
\ref{eq:K_MAP_MUSIC}) are not exact MAP estimates anymore, but merely
approximations. Hence, the optimal performance of estimates $\widehat{\bULA}$
and $\widehat{\nstate}$ is not guaranteed in this case.

From default setting, the number of sensors is reduced from $\ndim=100$
to  $\ndim=15$ in Fig. \ref{fig:correlatedD}. In this case of moderate
correlation, the performance is still similar to the uncorrelated
case of default setting. The estimate of $\nstate$ in DTFT, however,
switches from overfitting to underfitting in Fig. \ref{fig:correlatedD}.
The performance of non eigen-based DTFT is similar to that of eigen-based
MUSIC algorithm in non-overlapping case $\vartheta=0\%$, but becomes
better in nearly overlapping case $\vartheta=99.9\%$. The approximated
MAP estimate $\widehat{\nstate}$ (\ref{eq:K_MAP_MUSIC}) in this
case is still much better than the AIC method, although their performance
in this correlated case is worse than default setting. The estimate
$\overline{\rationoise}$ is still a useful indicator for detecting
the limit $\SNR=-10$ (dB), although it is not suitable for the AIC
method in nearly overlapping case $\vartheta=99.9\%$.

In Fig. \ref{fig:correlatedDOA}, the DOA's difference is reduced
gradually from $\Delta\Angle=34^{0}$ in default setting to $\Delta\Angle=4^{0}$.
For $\Delta\Angle\leq4^{0}$, we found that the approximated MAP $\widehat{\nstate}$
in the DTFT method began yielding inaccurate estimates in middle SNR.
Intuitively, when DOA's difference $\Delta\Angle$ is too small, the
superposition of peaks of power leakage in Fig. \ref{fig:DOA} will
become comparable with the spectrum's peaks of sources, as illustrated
in first row of Fig. \ref{fig:correlatedDOA}. Hence, it is harder
for the DTFT method to extract the correct peaks of sources and to
return the correct MAP estimate $\widehat{\nstate}$. Also, since
this setting of low $\Delta\Angle$ yields higher correlation than
setting of low $\ndim=15$, the MAP estimate $\widehat{\nstate}$
in MUSIC is slightly worse than the case of $\ndim=15$ in Fig.~\ref{fig:correlatedD}.
Our estimate $\widehat{\nstate}$ is, nonetheless, still much better
than the AIC method overall. The estimate $\overline{\rationoise}$
is still a useful indicator for detecting the limit $\SNR=-10$ (dB)
in this case.

\subsection{Decayed multi-tone sources}

In this subsection, we will study the case of uncorrelated DOAs with
different amplitudes. From default setting, we now set $\amplitude_{\istate,\isource}=(1-\psi\frac{\istate-1}{\nstate})\UNIT_{\isource\in[\isource_{\istate},\isource_{\istate}+\text{BW}]}$,
$\forall\seti{\istate}{\nstate}$, in which the decayed ratio is $\psi\in[0,1]$.
The simulations with different values $\psi$ are given in Fig. \ref{fig:Decay}.

Since there is a power leakage, even for uncorrelated DOAs, as illustrated
in Fig.  \ref{fig:DOA}, the strongly decayed amplitudes will be confused
with this power leakage. Indeed, there are six DTFT peaks instead
of the ground-truth $\nstate=5$ peaks in first row of Fig. \ref{fig:Decay}.
Hence, for middle SNR regime, the estimate's accuracy of this case
is worse than that of the default setting in Fig. \ref{fig:Overlapping}.
Nonetheless, the estimate's accuracy of amplitudes and $\overline{\rationoise}$
for all methods in this case is not much different from default setting,
for all cases of SNR.

In the case of non-overlapping $\vartheta=0\%$ in Fig. \ref{fig:Decay},
the MAP estimate $\widehat{\nstate}$ is superior to that by the AIC
method in moderately decayed setting $\psi=50\%$, although it is
worse than the AIC method in middle SNR regime of strongly decayed
setting $\psi=90\%$. The reason for this is likely owing to our amplitude's
prior in (\ref{eq:ra_PCA}, \ref{eq:postAmatrix}), which takes into
account the average of all amplitude's variances. The decision of
our MAP estimate $\widehat{\nstate}$ is, hence, influenced by the
average value of all decayed amplitudes, instead of each decayed amplitude
separately. This decayed setting suggests that we may have to consider
individual decayed amplitudes for the MAP estimate $\widehat{\nstate}$
in future works.

In the case of almost overlapping $\vartheta=99.9\%$ in Fig. \ref{fig:Decay},
nonetheless, the MAP estimate $\widehat{\nstate}$ is still much superior
to AIC method, since our MAP criterion (\ref{eq:K_MAP_Stiefel}, \ref{eq:K_MAP_MUSIC})
is not an eigen-based method, as explained in default setting above. 

\section{Conclusion\label{sec:Conclusion}}

In this paper, we have derived a closed-form solution for MAP estimate
of the number $\nstate$ of sources in PCA, MUSIC and DTFT spectrum
methods. For this purpose, we have also derived two novel probability
distributions, namely double gamma and double inverse-gamma distributions.
Owing to these distributions, we recognized that the posterior probability
distribution of $\nstate$ takes into account all possible binomial
combinations of signal and noise subspaces in noisy data space. The
MAP estimate of $\nstate$ then corresponds to the dimension of signal
subspace with highest probability of domination of signal-plus-noise's
variance over noise's variance. 

In simulations of linear sensor array, we also recognized that, for
accurate estimation, the $\SNR$ of maximum signal's power should
be higher than $-10$dB, which means the estimated noise-to-signal
percentage $\overline{\ratioNoise}$ should be less than $90\%$ (i.e.
the projected noise's deviation on signal space should be less than
three deviation of source's amplitudes). 

For overlapping multi-tone sources, our MAP estimate method was shown
to be far superior to eigen-based methods like Akaike information
criterion (AIC) for PCA. Our MAP estimate method is, however, only
based on averaged value of amplitude's variances and uncorrelated
principal/steering vectors. The MAP estimates for individual amplitudes
and correlated principal vectors are, hence, interesting cases for
future works. 

\appendices{}

\section{Negative binomial and double gamma distributions\label{sec:Double-Gamma}}

In this Appendix, let us derive two novel distributions, namely double
gamma and double inverse-gamma distributions, which were used for
estimating the signal's and noise's variance in section \ref{subsec:Posterior-variance}.
For this purpose, let us firstly show the relationship between negative
binomial distribution and order statistics of two independent gamma
distributions, as follows:
\begin{thm}
(Negative binomial distribution)\label{thm:(Negative-binomial)} Let
$\GGx\sim\calG_{\GGx}(\GGnx,\GGsx)$ and $\GGy\sim\calG_{\GGy}(\GGny,\GGsy)$
be random variables (r.v.) of two independent gamma distributions,
with positive integers $\GGnx$, $\GGny$ being the degree of freedom.
The probability of the event $\GGx\leq\GGy$ is: 
\begin{align}
\Pr[\UNIT_{\GGx\leq\GGy}] & =\Regular_{p}(\GGnx,\GGny)=1-\Regular_{1-p}(\GGny,\GGnx),\ \text{with}\ p\TRIANGLEQ\frac{\GGsx}{\GGsx+\GGsy},\label{eq:xi_beta}
\end{align}
where $\UNIT_{\GGx\leq\GGy}$ is the boolean indicator function, $\Regular_{p}(\GGnx,\GGny)=\frac{\int_{0}^{\GGp}\GGt^{\GGnx-1}(1-\GGt)^{\GGny-1}d\GGt}{B(\GGnx,\GGny)}$
is the regularized incomplete beta function and: 
\begin{align}
\Regular_{p}(\GGnx,\GGny) & =1-\underset{\calNB_{\GGi}(\GGny,p)}{\sum_{\GGi=0}^{\GGnx-1}\underbrace{\left(\begin{array}{c}
\GGny+\GGi-1\\
\GGi
\end{array}\right)(1-\GGp)^{\GGny}\GGp^{\GGi}}}\label{eq:Ip}\\
 & =1-\Regular_{1-p}(\GGny,\GGnx)=\sum_{\GGi=0}^{\GGny-1}\calNB_{\GGi}(\GGnx,1-p),\nonumber 
\end{align}
with $\left(\begin{array}{c}
\GGn\\
\GGi
\end{array}\right)\TRIANGLEQ\frac{\GGn!}{(\GGn-\GGi)!\GGi!}$ denoting binomial coefficient. 

Note that, $\Regular_{p}(\GGnx,\GGny)$ is actually the cumulative
mass function (c.m.f) of a negative binomial distribution $\calNB_{\GGi}(\GGnx,1-p)$,
$\forall\GGi\in\{0,1\ldots,\infty\}$. Likewise, the reverse form
$\Regular_{1-p}(\GGny,\GGnx)=1-\Regular_{p}(\GGnx,\GGny)$ is the
c.m.f of $\calNB_{\GGi}(\GGny,p)$ in (\ref{eq:Ip}). Hence, we also
have $\Regular_{p}(\GGnx,\GGny)=\sum_{\GGi=\GGnx}^{+\infty}\calNB_{\GGi}(\GGny,p)$,
i.e. the reverse c.m.f. of $\calNB_{\GGi}(\GGny,p)$. 
\end{thm}
Let us prove Theorem \ref{thm:(Negative-binomial)} together with
Corollary \ref{cor:(Double-gamma)} below.
\begin{cor}
(Double gamma distributions)\label{cor:(Double-gamma)} In Theorem
\ref{thm:(Negative-binomial)}, the conditional probability distribution
function (p.d.f) of $\GGx$ given $\GGy$ is the right-truncated gamma
distribution, while the marginal p.d.f of $\GGx$ and $\GGy$ are
called the lower- and upper-double gamma distributions, respectively,
as follows:
\begin{align}
f(\GGx|\GGy,\UNIT_{\GGx\leq\GGy}) & =\calrG_{\GGx\leq\GGy}(\GGnx,\GGsx)\TRIANGLEQ\frac{\calG_{\GGx}(\GGnx,\GGsx)}{\frac{\gamma(\GGnx,\GGsx\GGy)}{\Gamma(\GGnx)}},\label{eq:Double-gamma}\\
f(\GGx|\UNIT_{\GGx\leq\GGy}) & =\calGGL_{\GGx}(\GGnx,\GGny,\GGsx,\GGsy)\TRIANGLEQ\frac{\Gamma\left(\GGny,\GGsy\GGx\right)}{\Gamma(\GGny)}\frac{\calG_{\GGx}(\GGnx,\GGsx)}{\Regular_{p}(\GGnx,\GGny)},\nonumber \\
f(\GGy|\UNIT_{\GGx\leq\GGy}) & =\calGGU_{\GGy}(\GGnx,\GGny,\GGsx,\GGsy)\TRIANGLEQ\frac{\gamma(\GGnx,\GGsx\GGy)}{\Gamma(\GGnx)}\frac{\calG_{\GGy}(\GGny,\GGsy)}{\Regular_{p}(\GGnx,\GGny)},\nonumber 
\end{align}
where $\gamma(\GGn,\GGs)=\Gamma(n)-\Gamma(\GGn,\GGs)$ and $\Gamma(\GGn,\GGs)$
denote the lower and upper incomplete gamma functions, respectively,
with $\Gamma(n)=(n-1)!$ denoting the gamma function. Then, their
$\istate$-th moments are:
\begin{align}
\overline{\GGx^{k}} & =\EXPECTATION_{f(\GGx|\UNIT_{\GGx\leq\GGy})}\GGx^{k}=\frac{\Gamma(\GGnx+k)}{\GGsx^{k}\Gamma(\GGnx)}\frac{\Regular_{p}(\GGnx+\GGi,\GGny)}{\Regular_{p}(\GGnx,\GGny)},\nonumber \\
\overline{\GGy^{k}} & =\EXPECTATION_{f(\GGy|\UNIT_{\GGx\leq\GGy})}\GGy^{k}=\frac{\Gamma(\GGny+k)}{\GGsy^{k}\Gamma(\GGny)}\frac{\Regular_{p}(\GGnx,\GGny+\GGi)}{\Regular_{p}(\GGnx,\GGny)}.\label{eq:k-moment}
\end{align}
\end{cor}
\begin{IEEEproof}
Firstly, the conditional probability mass function (p.m.f) of $\UNIT_{\GGx\leq\GGy}$
is $\Prob[\UNIT_{\GGx\leq\GGy}|\GGx,\GGy]=\UNIT_{\GGx\leq\GGy}$.
The joint distribution is then: 
\begin{align*}
f(\GGx,\GGy,\UNIT_{\GGx\leq\GGy}) & =\Prob[\UNIT_{\GGx\leq\GGy}|\GGx,\GGy]f(\GGx)f(\GGy)\\
 & =f(\GGx|\GGy,\UNIT_{\GGx\leq\GGy})\Prob[\UNIT_{\GGx\leq\GGy}|\GGy]f(\GGy)\\
 & =f(\GGx|\GGy,\UNIT_{\GGx\leq\GGy})f(\GGy|\UNIT_{\GGx\leq\GGy})\Prob[\UNIT_{\GGx\leq\GGy}],
\end{align*}
in which, by Bayes' rule, the posterior $f(\GGx|\GGy,\UNIT_{\GGx\leq\GGy})=\frac{\Prob[\UNIT_{\GGx\leq\GGy}|\GGx,\GGy]f(\GGx)}{\Prob[\UNIT_{\GGx\leq\GGy}|\GGy]}$
is right-truncated inverse-gamma distribution in (\ref{eq:Double-gamma}),
since $\Prob[\UNIT_{\GGx\leq\GGy}|\GGy]=\int_{0}^{\GGy}\Prob[\UNIT_{\GGx\leq\GGy}|\GGx,\GGy]f(\GGx)d\GGx=\UNIT_{\GGx\leq\GGy}\int_{0}^{\GGx}\calG_{\GGx}(s,t)d\GGy=\UNIT_{\GGx\leq\GGy}\frac{\gamma(s,t\GGy)}{\Gamma(s)}$.
Likewise, the Bayes' rule yields $f(\GGy|\UNIT_{\GGx\leq\GGy})=\frac{\Prob[\UNIT_{\GGx\leq\GGy}|\GGy]f(\GGy)}{\Prob[\UNIT_{\GGx\leq\GGy}]}$
in (\ref{eq:Double-gamma}), as follows: 
\begin{align}
\Pr[\UNIT_{\GGx\leq\GGy}] & =\int_{0}^{\infty}\Prob[\UNIT_{\GGx\leq\GGy}|\GGy]f(\GGy)d\GGy\label{eq:proof_Pr}\\
 & =\UNIT_{\GGx\leq\GGy}\int_{0}^{\infty}\frac{\gamma(\GGnx,\GGsx\GGy)}{\Gamma(\GGnx)}\calG_{\GGy}\left(\GGny,\GGsy\right)d\GGy,\nonumber 
\end{align}
Solving (\ref{eq:proof_Pr}) via series form $\frac{\gamma(\GGn,\GGs)}{\Gamma(\GGn)}=1-\sum_{\GGi=0}^{\GGn-1}\frac{\GGs^{\GGi}e^{-\GGs}}{\GGi!}$
and expectation of gamma distribution $\calG_{x}\left(\GGu,\GGv\right)=\frac{\GGv^{\GGu}}{\Gamma(\GGu)}x^{\GGu-1}e^{-\GGv x}$,
we obtain: 
\begin{align*}
\int_{0}^{\infty}\frac{\gamma(\GGnx,\GGsx\GGy)}{\Gamma(\GGnx)}\calG_{\GGy}\left(\GGny,\GGsy\right)d\GGy & =1-\sum_{\GGi=0}^{\GGnx-1}\frac{\Gamma(\GGny+\GGi)}{\Gamma(\GGny)\GGi!}\frac{\GGsy^{\GGny}\GGsx^{\GGi}}{(\GGsy+\GGsx)^{\GGny+\GGi}}\\
 & =\Regular_{\frac{\GGsx}{\GGsx+\GGsy}}(\GGnx,\GGny),
\end{align*}
which yields (\ref{eq:xi_beta}). Similarly, we can compute $f(X|\UNIT_{X\leq Y})=\int_{X}^{\infty}f(X|Y,\UNIT_{X\leq Y})f(Y|\UNIT_{X\leq Y})dY$
in (\ref{eq:Double-gamma}). Also, we have: $\GGx^{\GGi}\calG_{\GGx}(\GGnx,\GGsx)=\frac{\Gamma(\GGnx+k)}{\GGsx^{k}\Gamma(\GGnx)}\calG_{X}(\GGnx+\GGi,t)$,
hence the moments (\ref{eq:k-moment}).
\end{IEEEproof}
By simply changing the gamma distribution to inverse-gamma distribution,
we can extend the above results to inverse-gamma distributions feasibly,
as follows:
\begin{cor}
(Double inverse-gamma distributions)\label{cor:(inverse-gamma)} Similar
to Theorem \ref{thm:(Negative-binomial)} and  Corollary \ref{cor:(Double-gamma)},
let $\GGx\sim\caliG_{\GGx}(\GGnx,\GGsx)$ and $\GGy\sim\calG_{\GGy}(\GGny,\GGsy)$
be r.v. of independent inverse-gamma distributions, with positive
integers $\GGnx$, $\GGny$. If $\GGx\geq\GGy,$ the conditional p.d.f
of $\GGx$ given $\GGy$ is the left-truncated inverse-gamma distribution,
while the marginal p.d.f of $\GGx$ and $\GGy$ are called the upper-
and lower-double inverse-gamma distributions, respectively, as follows:
\begin{align}
f(\GGx|\GGy,\UNIT_{\GGx\geq\GGy}) & =\caliG_{\GGx\geq\GGy}(\GGnx,\GGsx)\TRIANGLEQ\frac{\caliG_{\GGx}(\GGnx,\GGsx)}{\frac{\gamma(\GGnx,\frac{\GGsx}{\GGy})}{\Gamma(\GGnx)}},\label{eq:inverse-gamma}\\
f(\GGx|\UNIT_{\GGx\geq\GGy}) & =\caliGGU_{\GGx}(\GGnx,\GGny,\GGsx,\GGsy)\TRIANGLEQ\frac{\Gamma\left(\GGny,\frac{\GGsy}{\GGx}\right)}{\Gamma(\GGny)}\frac{\caliG_{\GGx}(\GGnx,\GGsx)}{\Regular_{p}(\GGnx,\GGny)},\nonumber \\
f(\GGy|\UNIT_{\GGx\geq\GGy}) & =\caliGGL_{\GGy}(\GGnx,\GGny,\GGsx,\GGsy)\TRIANGLEQ\frac{\gamma(\GGnx,\frac{\GGsx}{\GGy})}{\Gamma(\GGnx)}\frac{\caliG_{\GGy}(\GGny,\GGsy)}{\Regular_{p}(\GGnx,\GGny)},\nonumber 
\end{align}
with $\Pr[\UNIT_{\GGx\geq\GGy}]=\Regular_{p}(\GGnx,\GGny),$ $p\TRIANGLEQ\frac{\GGsx}{\GGsx+\GGsy}$
and $\Regular_{p}(\GGnx,\GGny)$ given in (\ref{eq:xi_beta}, \ref{eq:Ip}).
Then, similar to (\ref{eq:k-moment}), the moments for (\ref{eq:inverse-gamma})
are: 
\begin{align}
\overline{\GGx^{k}} & =\EXPECTATION_{f(\GGx|\UNIT_{\GGx\geq\GGy})}\GGx^{k}=\frac{\Gamma(\GGnx-k)}{\GGsx^{-k}\Gamma(\GGnx)}\frac{\Regular_{p}(\GGnx-\GGi,\GGny)}{\Regular_{p}(\GGnx,\GGny)},\nonumber \\
\overline{\GGy^{k}} & =\EXPECTATION_{f(\GGy|\UNIT_{\GGx\geq\GGy})}\GGy^{k}=\frac{\Gamma(\GGny-k)}{\GGsy^{-k}\Gamma(\GGny)}\frac{\Regular_{p}(\GGnx,\GGny-\GGi)}{\Regular_{p}(\GGnx,\GGny)}.\label{eq:inverse-moment}
\end{align}
\end{cor}

\section{Bayesian minimum-risk estimation \label{sec:Bayesian-minimum-risk}}

Let us briefly review the importance of posterior distributions in
practice, via minimum-risk property of Bayesian estimation method.
Without loss of generalization, let us assume that the unknown parameter
$\para$ in our model is continuous. In practice, the aim is often
to return estimated value $\hat{\para}\TRIANGLEQ\hat{\para}(\Data)$,
as a function of noisy data $\Data$, with minimum mean squared error
$\text{MSE}(\hat{\para},\para)\TRIANGLEQ\EXPECTATION_{f(\Data,\para)}||\hat{\para}(\Data)-\para||_{2}^{2}$,
where $||\cdot||_{2}$ is $\Lp_{2}$-normed operator. Then, by basic
chain rule of probability $f(\Data,\para)=f(\para|\Data)f(\Data)$,
we have \cite{VH:PhDthesis:14,Bayes:BOOK:Bernado}: 
\begin{align}
\hat{\para} & \TRIANGLEQ\argmin_{\tilde{\para}}\text{MSE}(\tilde{\para},\para)\nonumber \\
 & =\argmin_{\tilde{\para}}\EXPECTATION_{f(\para|\Data)}||\tilde{\para}(\Data)-\para||_{2}^{2}\label{eq:MSE}\\
 & =\EXPECTATION_{f(\para|\Data)}(\para),\nonumber 
\end{align}
which shows that the posterior mean $\hat{\para}=\EXPECTATION_{f(\para|\Data)}(\para)$
is the minimum MSE (MMSE) estimate. In general, we may replace the
$\Ltwo$-norm in (\ref{eq:MSE}) by other normed functions. For example,
it is well-known that the best estimators for averaged $\Lp_{1}$
and $\Lp_{\infty}$-normed error are posterior median and mode of
$f(\para|\Data)$, respectively \cite{VH:PhDthesis:14,Bayes:BOOK:Bernado}. 

\bibliographystyle{IEEEtran}
\bibliography{PCA_MUSIC}

\end{document}